\documentclass{article}

\usepackage[nonatbib,preprint]{neurips_2021}
\usepackage{graphicx}
\usepackage{subfig}
\usepackage{floatrow}
\usepackage{wrapfig}

\usepackage{amssymb,bm,amsthm}
\usepackage{enumerate}
\usepackage{color, colortbl}
\usepackage{booktabs, multirow}
\definecolor{darkgreen}{rgb}{0, 0.5, 0}
\definecolor{red}{rgb}{1, 0, 0}
\definecolor{purple}{rgb}{0.5, 0, 0.5}
\usepackage{chngpage}
\usepackage{comment}
\usepackage{mfirstuc}
\usepackage{mathtools}
\usepackage{lettrine}

\newcommand\ie{\textit{i.e.,}}
\newcommand\eg{\textit{e.g.,}}
\newcommand\st{\textit{s.t.}}
\newcommand\wrt{\textit{w.r.t.}}
\newcommand\etc{\textit{etc.}}
\newcommand\iid{\textit{i.i.d.}}

\newcommand{\norm}[1]{\left\lVert#1\right\rVert}

\newcommand{\beq}{\begin{equation}}
\newcommand{\eeq}{\end{equation}}
\newcommand{\beqnn}{\begin{equation*}}
\newcommand{\eeqnn}{\end{equation*}}
\newcommand{\beqy}{\begin{eqnarray}}
\newcommand{\eeqy}{\end{eqnarray}}
\newcommand{\beqynn}{\begin{eqnarray*}}
\newcommand{\eeqynn}{\end{eqnarray*}}
\newcommand{\bit}{\begin{itemize}}
\newcommand{\eit}{\end{itemize}}
\newcommand{\ben}{\begin{enumerate}}
\newcommand{\een}{\end{enumerate}}
\newcommand{\bex}{\begin{example}}
\newcommand{\eex}{\end{example}}
\newcommand{\trace}{\mathrm{trace}}

\newcommand{\balg}[1]{\begin{algorithm} \caption{#1}}
\newcommand{\ealg}{\end{algorithm}}

\newcommand{\balgc}{\begin{algorithmic}[1]}
\newcommand{\ealgc}{\end{algorithmic}}

\newcommand{\bary}{\begin{array}}
\newcommand{\eary}{\end{array}}
\newcommand{\bmx}{\begin{bmatrix}}
\newcommand{\emx}{\end{bmatrix}}
\newcommand{\bsmx}{\left[\begin{smallmatrix}}
\newcommand{\esmx}{\end{smallmatrix}\right]}
\newcommand{\bmxc}[1]{\left[\begin{array}{@{}#1@{}}}
\newcommand{\emxc}{\end{array}\right]}
\newcommand{\bcn}{\begin{center}}
\newcommand{\ecn}{\end{center}}

\newcommand{\diag}{\mathrm{diag}}

\newcommand{\Rbb}{{\mathbb{R}}}

\newcommand{\e}{\boldsymbol{e}}

\renewcommand{\u}{\boldsymbol{u}}

\newcommand{\x}{{\boldsymbol{x}}}

\providecommand{\norm}[1]{\lVert#1\rVert}

\providecommand{\abs}[1]{\left| #1 \right|}

\newenvironment{theorem}[2][Theorem]{\begin{trivlist}
		\item[\hskip \labelsep {\bfseries #1}\hskip \labelsep {\bfseries #2.}]}{\end{trivlist}}

\newenvironment{corollary}[2][Corollary]{\begin{trivlist}
		\item[\hskip \labelsep {\bfseries #1}\hskip \labelsep {\bfseries #2.}]}{\end{trivlist}}

\usepackage[utf8]{inputenc} 
\usepackage[T1]{fontenc}    
\usepackage{hyperref}       
\usepackage{url}            
\usepackage{booktabs}       
\usepackage{amsfonts}       
\usepackage{nicefrac}       
\usepackage{microtype}      
\usepackage{xcolor}         

\newtheorem{definition}{Definition}
\newfloatcommand{capbtabbox}{table}[][\FBwidth]

\usepackage{amsmath}

\allowdisplaybreaks

\title{Is Heterophily A Real Nightmare For Graph Neural Networks To Do Node Classification?}

\author{
Sitao Luan$^{1,2}$, Chenqing Hua$^{1}$, Qincheng Lu$^{1}$, Jiaqi Zhu$^{1}$, Mingde Zhao$^{1,2}$, Shuyuan Zhang$^{1,2}$,\\
\textbf{ Xiao-Wen Chang$^{1}$, Doina Precup$^{1,2,3}$}\\
\{sitao.luan@mail, chenqing.hua@mail, qincheng.lu@mail, jiaqi.zhu@mail, mingde.zhao@mail,\\ shuyuan.zhang@mail, chang@cs, dprecup@cs\}.mcgill.ca\\
$^1$McGill University; $^2$Mila; $^3$DeepMind\\
}
 
\begin{document}

\maketitle

\begin{abstract}
  Graph Neural Networks (GNNs) extend basic Neural Networks (NNs) by using the graph structures based on the relational inductive bias (homophily assumption). Though GNNs are believed to outperform NNs in real-world tasks, performance advantages of GNNs over graph-agnostic NNs seem not generally satisfactory. Heterophily has been considered as a main cause and numerous works have been put forward to address it. In this paper, we first show that not all cases of heterophily are harmful \footnote{In general, harmful heterophily means the heterophilous structure that will make a graph-aware model underperform its corresponding graph-agnostic model.} for GNNs with aggregation operation. Then, we propose new metrics based on a similarity matrix which considers the influence of both graph structure and input features on GNNs. The metrics demonstrate advantages over the commonly used homophily metrics by tests on synthetic graphs. From the metrics and the observations, we find some cases of harmful heterophily can be addressed by diversification operation. With this fact and knowledge of filterbanks, we propose the Adaptive Channel Mixing (ACM) framework to adaptively exploit aggregation, diversification and identity channels in each GNN layer to address harmful heterophily. We validate the ACM-augmented baselines with $10$ real-world node classification tasks. They consistently achieve significant performance gain and exceed the state-of-the-art GNNs on most of the tasks without incurring significant computational burden. 
\end{abstract}

\section{Introduction}
\label{sec:introduction}
Deep Neural Networks (NNs) \cite{lecun2015deep} have revolutionized many machine learning areas, including image recognition \cite{krizhevsky2012imagenet}, speech recognition \cite{graves2013speech} and natural language processing \cite{bahdanau2014neural}, \etc  One major strength is their capacity and effectiveness of learning latent representation from Euclidean data. Recently, the focus has been put on its applications on non-Euclidean data \cite{bronstein2016geometric}, \eg{} relational data or graphs. Combining graph signal processing and convolutional neural networks \cite{lecun1998gradient}, numerous Graph Neural Networks (GNNs) \cite{scarselli2008graph,defferrard2016fast,hamilton2017inductive,velivckovic2017attention,kipf2016classification,luan2019break} have been proposed which empirically outperform traditional neural networks on graph-based machine learning tasks, \eg{} node classification, graph classification, link prediction and graph generation, \etc
GNNs are built on the homophily assumption\cite{mcpherson2001birds}, \ie{} connected nodes tend to share similar attributes with each other \cite{hamilton2020graph}, which offers additional information besides node features. Such relational inductive bias \cite{battaglia2018relational} is believed to be a key factor leading to GNNs' superior performance over NNs' in many tasks. 

Nevertheless, growing evidence shows that GNNs do not always gain advantages over traditional NNs when dealing with relational data. In some cases, even simple Multi-Layer Perceptrons (MLPs) can outperform GNNs by a large margin \cite{zhu2020beyond,liu2020non,chien2021adaptive}. An important reason for the performance degradation is believed to be the heterophily problem, \ie{} connected nodes tend to have different labels which makes the homophily assumption fail. Heterophily challenge has received attention recently and there are increasing number of models being put forward to address this problem  \cite{zhu2020beyond,liu2020non,chien2021adaptive,zhu2020graph,yan2021two}.  

\paragraph{Contributions} In this paper, we first demonstrate that not all heterophilous graphs are harmful for aggregation-based GNNs and the existing metrics of homophily are insufficient to decide whether the aggregation operation will make nodes less distinguishable or not. By constructing a similarity matrix from backpropagation analysis, we derive new metrics to depict how much GNNs are influenced by the graph structure and node features. We show the advantage of our metrics over the existing metrics by comparing the ability of characterizing the performance of two baseline GNNs on synthetic graphs of different levels of homophily. From the similarity matrix, we find that diversification operation is able to address some harmful heterophily cases, and based on which we propose Adaptive Channel Mixing (ACM) GNN framework. The experiments on the synthetic datasets, ablation studies and real-world datasets consistently show that the baseline GNNs augmented by ACM framework are able to obtain significant performance boost on node classification tasks on heterophilous graphs.

The rest of this paper is mainly organized as follows: In section \ref{sec:prelimiary_notation}, we introduce the notation and the background knowledge. In section \ref{sec:heterophily_analysis}, we conduct node-wise analysis on heterophily, derive new homophily metrics based on a similarity matrix and conduct experiments to show their advantages over the existing homophily metrics. In section \ref{sec:acm_framework}, we demonstrate the capability of diversification operation on addressing some cases of harmful heterophily and propose the ACM-GNN framework to adaptively utilize the information from different filterbank channels to address heterophily problem. In section \ref{sec:related_works}, we discuss the related works and clarify the differences to our method. In section \ref{sec:experiments}, we provide empirical evaluations on ACM framework, including ablation study and tests on $10$ real-world node classification tasks.

\section{Preliminaries}
\label{sec:prelimiary_notation}
We will introduce the related notation and background knowledge in this section. We use \textbf{bold} fonts for vectors (\eg{} $\bm{v}$). Suppose we have an undirected connected graph $\mathcal{G}=(\mathcal{V},\mathcal{E}, A)$, where $\mathcal{V}$ is the node set with $\abs{\mathcal{V}}=N$; $\mathcal{E}$ is the edge set without self-loop; $A \in \mathbb{R}^{N\times N}$ is the symmetric adjacency matrix with $A_{i,j}=1$ \textit{iff} $e_{ij} \in \mathcal{E}$, otherwise $A_{i,j}=0$.
We use $D$ to denote the diagonal degree matrix of ${\cal G}$, \ie{} $D_{i,i} = d_i = \sum_j A_{i,j}$ and 
use $\mathcal{N}_i$ to denote the neighborhood set of node $i$, \ie{}  $\mathcal{N}_i=\{j: e_{ij} \in \mathcal{E}\}$. A graph signal is a vector $\bm{x} \in \mathbb{R}^N$ defined on $\mathcal{V}$, where $\bm{x}_i$ is defined on the node $i$. We also have a feature matrix ${X} \in \mathbb{R}^{N\times F}$, whose columns are graph signals and whose $i$-th row  ${X_{i,:}}$ is a feature vector of node $i$. We use $Z\in \mathbb{R}^{N\times C}$ to denote the label encoding matrix, whose $i$-th row  $Z_{i,:}$ is the one-hot encoding of the label of node $i$. 

\subsection{Graph Laplacian, Affinity Matrix and Their Variants} 
\label{sec:laplacian_affinity_matrix}
The (combinatorial) graph Laplacian is defined as $L = D - A$, which is Symmetric Positive Semi-Definite (SPSD)  \cite{chung1997spectral}. Its eigendecomposition gives $L=U\Lambda U^T$, where the columns $\u_i$ of $U\in \Rbb^{N\times N}$ are orthonormal eigenvectors, namely the \textit{graph Fourier basis}, $\Lambda = \diag(\lambda_1, \ldots, \lambda_N)$ with $\lambda_1 \leq \cdots \leq \lambda_N$, and these eigenvalues are also called \textit{frequencies}. The graph Fourier transform of the graph signal $\x$ is defined as $\bm{x}_\mathcal{F} = U^{-1} \bm{x} = U^{T} \bm{x} = [\u_1^T\x, \ldots, \u_N^T\x]^T$, where $\bm{u}_i^T \bm{x}$ is the component of $\bm{x}$ in the direction of $\bm{u_i}$. 

In additional to $L$, some variants are also commonly used, \eg{} the symmetric normalized Laplacian $L_{\text{sym}} = D^{-1/2} L D^{-1/2} = I-D^{-1/2} A D^{-1/2}$ and the random walk normalized Laplacian $L_{\text{rw}} = D^{-1} L = I - D^{-1} A$. 
The affinity (transition) matrices can be derived from the Laplacians, \eg{} $A_\text{rw} = I - L_\text{rw} = D^{-1} A$, $A_\text{sym} = I-L_\text{sym} = D^{-1/2} A D^{-1/2}$ and are considered to be low-pass filters \cite{maehara2019revisiting}.
Their eigenvalues satisfy $\lambda_i(A_\text{rw}) = \lambda_i(A_\text{sym}) = 1- \lambda_i(L_\text{sym}) = 1- \lambda_i(L_\text{rw}) \in (-1,1]$. 
Applying the renormalization trick \cite{kipf2016classification} to affinity and Laplacian matrices respectively leads to $\hat{A}_\text{sym} = \tilde{D}^{-1/2} \tilde{A} \tilde{D}^{-1/2}$ and $\hat{L}_{\text{sym}} = I - \hat{A}_\text{sym}$, 
where $\tilde{A} \equiv A+I$ and $\tilde{D} \equiv D+I$. The renormalized affinity matrix essentially adds a self-loop to each node in the graph, and is widely used in Graph Convolutional Network (GCN) \cite{kipf2016classification} as follows,
\begin{equation}
    \label{eq:gcn_original}
   Y = \text{softmax} (\hat{A}_\text{sym} \; \text{ReLU} (\hat{A}_\text{sym} {X} W_0 ) \; W_1 )
\end{equation}
where $W_0 \in \Rbb^{F\times F_1}$ and $W_1 \in \Rbb^{F_1\times O}$ are learnable parameter matrices. GCN can be trained by minimizing the following cross entropy loss
\begin{equation}
\label{eq:cross_entropy_loss}
     \mathcal{L}  = -\trace(Z^T \log Y)
\end{equation}
where $\log(\cdot)$ is a component-wise logarithm operation. The random walk renormalized matrix $\hat{A}_{\text{rw}} = \tilde{D}^{-1} \tilde{A}$,
which shares the same eigenvalues as $\hat{A}_{\text{sym}}$, can also be applied in GCN. The corresponding Laplacian is defined as $\hat{L}_{\text{rw}} = I - \hat{A}_\text{rw}$. 
$\hat{A}_{\text{rw}}$ is essentially a random walk matrix and
behaves as a mean aggregator that is applied in spatial-based GNNs \cite{hamilton2017inductive,hamilton2020graph}. To bridge the spectral and spatial methods, we use $\hat{A}_{rw}$ in the paper.

\subsection{Metrics of Homophily}
\label{sec:homophily_metrics}
The metrics of homophily are defined by considering different relations between node labels and graph structures defined by adjacency matrix. There are three commonly used homophily metrics: edge homophily \cite{abu2019mixhop,zhu2020beyond}, node homophily \cite{pei2020geom}, and class homophily \cite{lim2021new} \footnote{The authors in \cite{lim2021new} did not name this homophily metric. We name it class homophily based on its definition.} defined as follows:
\begin{equation}
\begin{aligned}
\label{eq:definition_homophily_metrics}
    &H_\text{edge}(\mathcal{G}) = \frac{\big|\{e_{uv} \mid e_{uv}\in \mathcal{E}, Z_{u,:}=Z_{v,:}\}\big|}{|\mathcal{E}|}, \ \ 
    H_\text{node}(\mathcal{G}) = \frac{1}{|\mathcal{V}|} \sum_{v \in \mathcal{V}} 
    \frac{\big|\{u \mid u \in \mathcal{N}_v, Z_{u,:}=Z_{v,:}\}\big|}{d_v}, \\
    &H_\text{class}(\mathcal{G}) = \frac{1}{C-1} \sum_{k=1}^{C}\left[h_{k}
    -\frac{\big|\{v \mid Z_{v,k} = 1 \}\big|}{N}\right]_{+}, \ \ 
    h_{k}=\frac{\sum_{v \in \mathcal{V}} \big|\{u \mid Z_{v,k} = 1, u \in \mathcal{N}_v,   Z_{u,:}=Z_{v,:}\}\big| }{\sum_{v \in \{v|Z_{v,k}=1\}} d_{v}}
\end{aligned}
\end{equation}
where $[a]_{+}=\max (a, 0)$; $h_{k}$ is the class-wise homophily metric \cite{lim2021new}. They are all in the range of $[0,1]$ and a value close to $1$ corresponds to strong homophily while a value close to $0$ indicates strong heterophily. $H_\text{edge}(\mathcal{G})$ measures the proportion of edges that connect two nodes in the same class; $H_\text{node}(\mathcal{G})$ evaluates the average proportion of edge-label consistency of all nodes; $H_\text{class}(\mathcal{G})$ tries to avoid the sensitivity to imbalanced class, which can cause $H_\text{edge}$ misleadingly large. The above definitions are all based on the graph-label consistency and imply that the inconsistency will cause harmful effect to the performance of GNNs. With this in mind, we will show a counter example to illustrate the insufficiency of the above metrics and propose new metrics in the following section.

\section{Analysis of Heterophily}
\label{sec:heterophily_analysis}
\subsection{Motivation and Aggregation Homophily}
\begin{wrapfigure}{R}{0.5\textwidth}
  \begin{center}
    \includegraphics[width=1.0\textwidth]{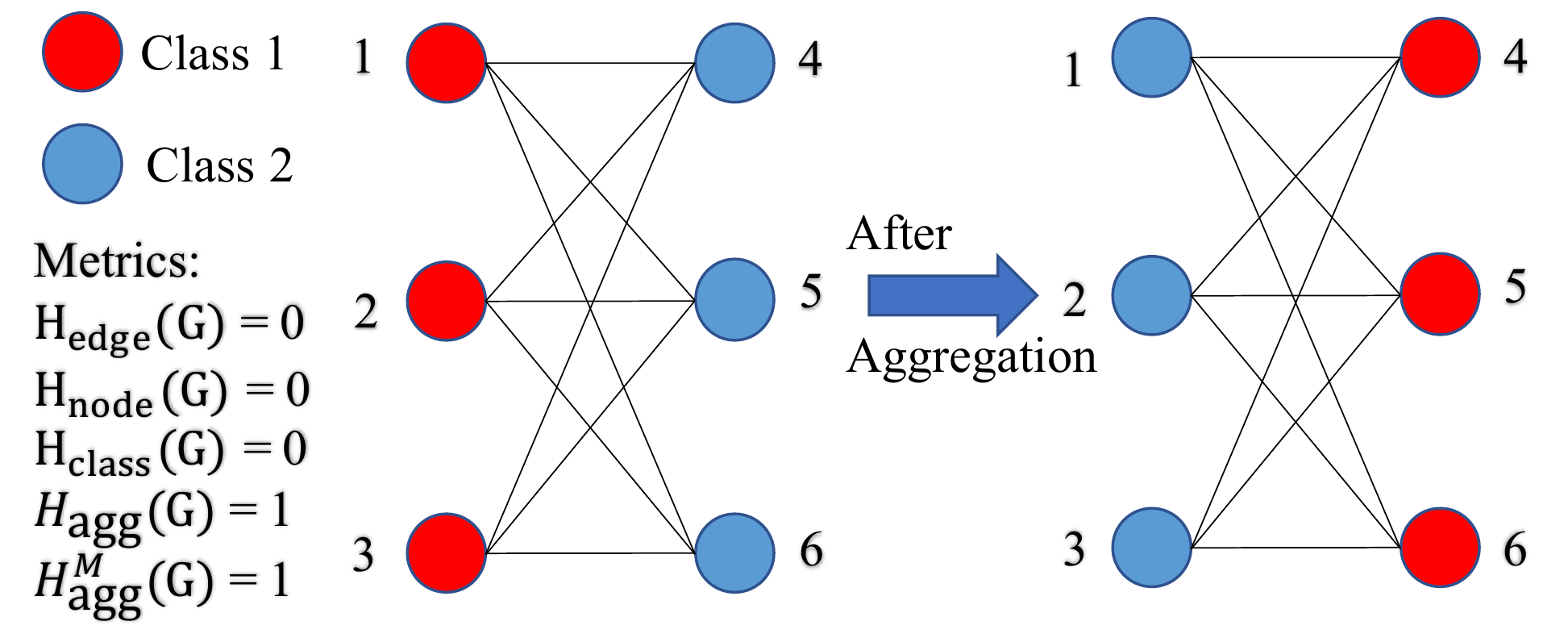}
  \end{center}
  \caption{Example of harmless heterophily}
  \label{fig:example_harmless_heterophily}
\end{wrapfigure}

Heterophily is believed to be harmful for message-passing based GNNs \cite{zhu2020beyond,pei2020geom,chien2021adaptive} because intuitively features of nodes in different classes will be falsely mixed and this will lead nodes indistinguishable \cite{zhu2020beyond}.
Nevertheless, it is not always the case, \eg{} the bipartite graph shown in Figure \ref{fig:example_harmless_heterophily} is highly heterophilous according to the homophily metrics in \eqref{eq:definition_homophily_metrics}, but after mean aggregation, the nodes in classes 1 and 2 only exchange colors and are still distinguishable. Authors in \cite{chien2021adaptive} also point out the insufficiency of $H_\text{node}$ by examples to show that different graph typologies with the same $H_\text{node}$ can carry different label information. 

To analyze to what extent the graph structure can affect the output of a GNN, we first simplify the GCN by removing its nonlinearity as \cite{wu2019simplifying}. Let $\hat{A} \in \mathbb{R}^{N\times N}$ denote a general aggregation operator. Then, equation \eqref{eq:gcn_original} can be simplified as,
\begin{equation}
    \begin{aligned}
    Y  = \text{softmax} (\hat{A}  X W ) = \text{softmax} (Y')
    \end{aligned}
\end{equation}
After each gradient decent step $\Delta W = \gamma \frac{d \mathcal{L}}{d W}$, where $\gamma$ is the learning rate, the update of $Y'$ will be (see Appendix \ref{appendix:details_of_nll_loss_explanation} for  derivation),
\begin{equation}
    \begin{aligned}
    \label{eq:gradient_descent_update}
    \Delta Y' = \hat{A} X \Delta W = \gamma \hat{A}X \frac{d \mathcal{L}}{d W} \propto \hat{A}X \frac{d \mathcal{L}}{d W} = \hat{A}X X^T\hat{A}^T (Z-Y) = S(\hat{A},X) (Z-Y)
    \end{aligned}
\end{equation}
where $S(\hat{A},X) \equiv \hat{A}X (\hat{A}X)^T$ is a post-aggregation node similarity matrix, $Z-Y$ is the prediction error matrix. The update direction of node $i$ is essentially a weighted sum of the prediction error, \ie{} $\Delta (Y')_{i,:} = \sum_{j\in \mathcal{V}} \left[S(\hat{A},X)\right]_{i,j} (Z-Y)_{j,:}$. 

To study the effect of heterophily, we first define the {\em aggregation similarity score} as follows.

\begin{definition} Aggregation similarity score 
\begin{equation}
\label{eq:aggregation_similarity}
    S_\text{agg}\left(S(\hat{A},X)\right) = \frac{\left| \left\{v   \,\big| \,
    \mathrm{Mean}_u\big( \{S(\hat{A},X)_{v,u} | Z_{u,:}=Z_{v,:} \}\big) 
    \geq \mathrm{Mean}_u\big(\{S(\hat{A},X)_{v,u} | Z_{u,:} \neq Z_{v,:} \} \big) \right\} \right|}{\left| \mathcal{V} \right|}
\end{equation}
where $\mathrm{Mean}_u\left(\{\cdot\}\right)$ takes the average over $u$ of a given multiset of values or variables.
\end{definition}
$S_\text{agg}(S(\hat{A},X))$ measures the proportion of nodes $v\in\mathcal{V}$ that will put relatively larger similarity weights on nodes in the same class than in other classes after aggregation. It is easy to see that $S_\text{agg}(S(\hat{A},X)) \in [0,1]$. But in practice, we observe that in most datasets, we will have $S_\text{agg}(S(\hat{A},X)) \geq 0.5$. Based on this observation, we rescale \eqref{eq:aggregation_similarity} to the following modified aggregation similarity for practical usage,
\begin{equation}
  S^M_\text{agg}\left(S(\hat{A},X)\right) = \left[2 S_\text{agg}\left(S(\hat{A},X)\right)-1\right]_{+}
\end{equation}
In order to measure the consistency between labels and graph structures without considering node features and make a fair comparison with the existing homophily metrics in \eqref{eq:definition_homophily_metrics}, we define the graph ($\mathcal{G}$) aggregation ($\hat{A}$) homophily and its modified version as
\begin{equation}
    \label{eq:aggregation_homophily_metrics}
    H_{\text{agg}}(\mathcal{G}) = S_\text{agg}\left(S(\hat{A},Z)\right), \; H_{\text{agg}}^M(\mathcal{G}) = S_\text{agg}^M\left(S(\hat{A},Z)\right)
\end{equation}
In practice, we will only check $H_{\text{agg}}(\mathcal{G})$ when $H_{\text{agg}}^M(\mathcal{G})=0$. As Figure \ref{fig:example_harmless_heterophily} shows, when $\hat{A} = \hat{A}_\text{rw}$, $H_{\text{agg}}(\mathcal{G}) = H_{\text{agg}}^M(\mathcal{G}) = 1$. Thus, this new metric reflects the fact that nodes in classes 1 and 2 are still highly distinguishable after aggregation,
while other metrics mentioned before fail to capture the information and misleadingly give value 0. This shows the advantage of $H_{\text{agg}}(\mathcal{G})$ and $H_{\text{agg}}^M(\mathcal{G})$ by additionally considering information from aggregation operator $\hat{A}$ and the similarity matrix.

To comprehensively compare $H_{\text{agg}}^M(\mathcal{G})$ with the metrics in \eqref{eq:definition_homophily_metrics} in terms of how they reveal the influence of graph structure on the GNN performance, we generate synthetic graphs and evaluate SGC \cite{wu2019simplifying} and GCN \cite{kipf2016classification} on them in the next subsection.


\subsection{Evaluation and Comparison on Synthetic Graphs}
\begin{figure}[h]
     {
     \subfloat[$H_\text{edge}(\mathcal{G})$]{
     \captionsetup{justification = centering}
     \includegraphics[height=0.11\textheight]{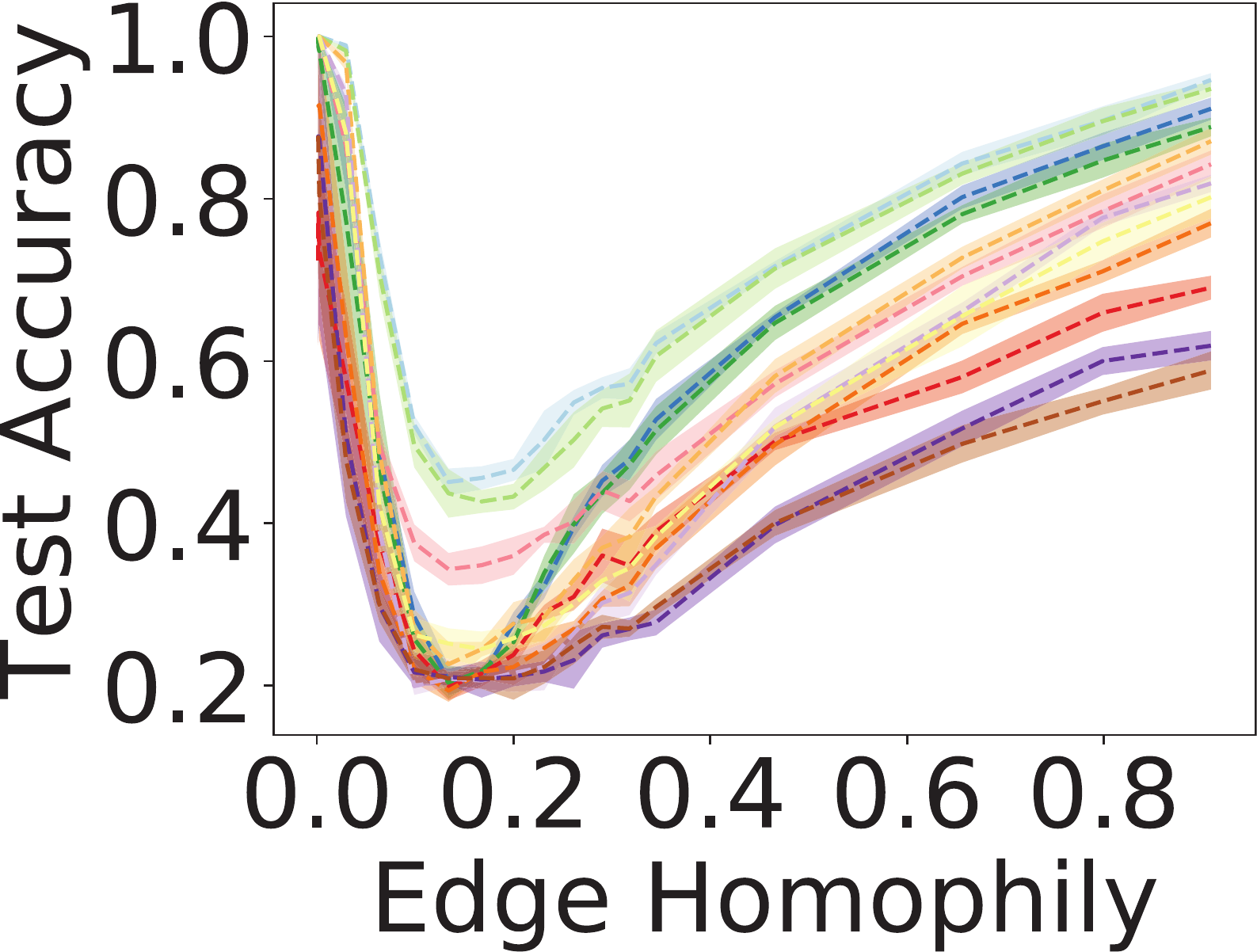}
     } 
     \subfloat[$H_\text{node}(\mathcal{G})$]{
     \captionsetup{justification = centering}
     \includegraphics[height=0.11\textheight]{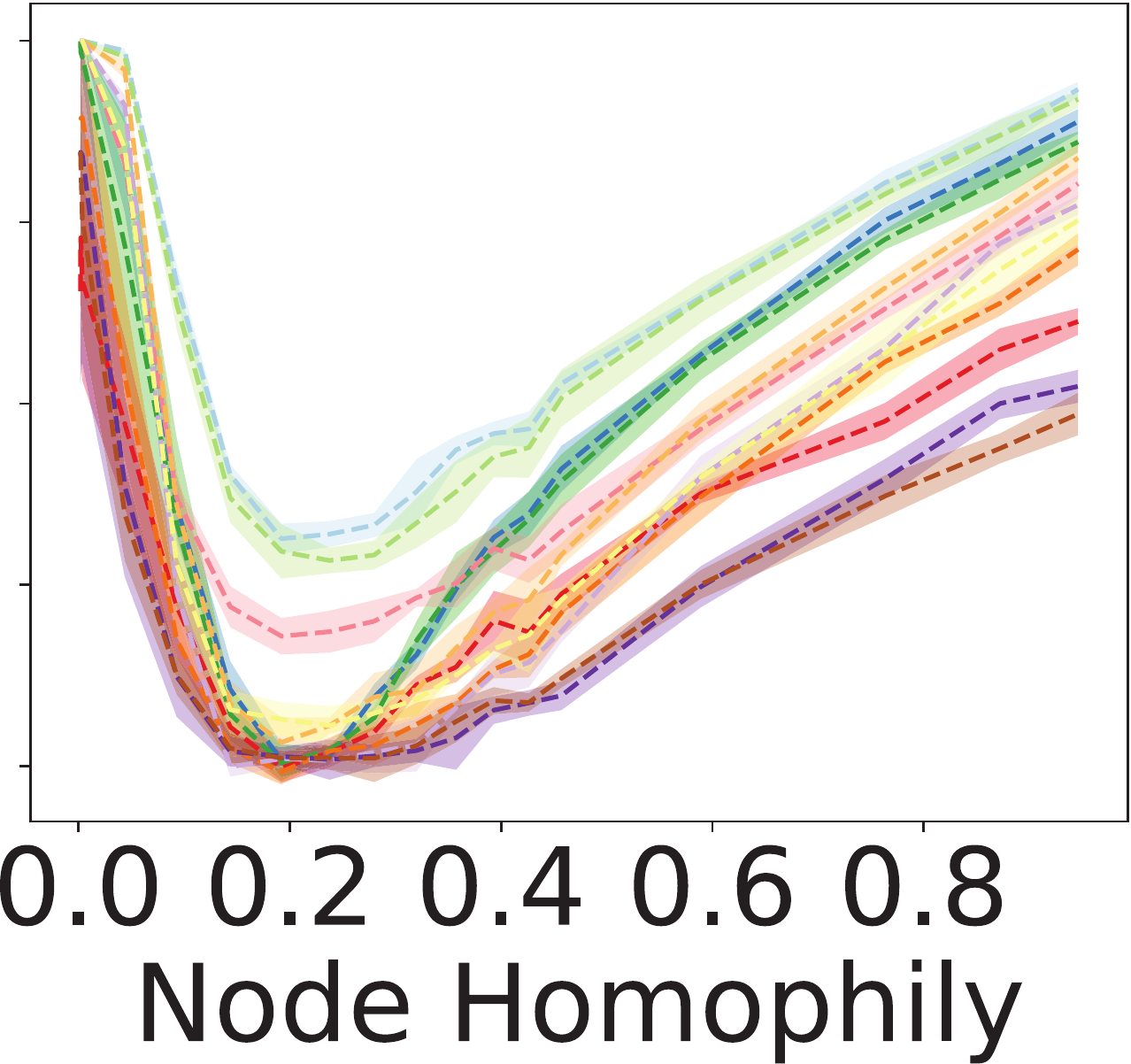}
     } 
     \subfloat[$H_\text{class}(\mathcal{G})$]{
     \captionsetup{justification=centering}
     \includegraphics[height=0.11\textheight]{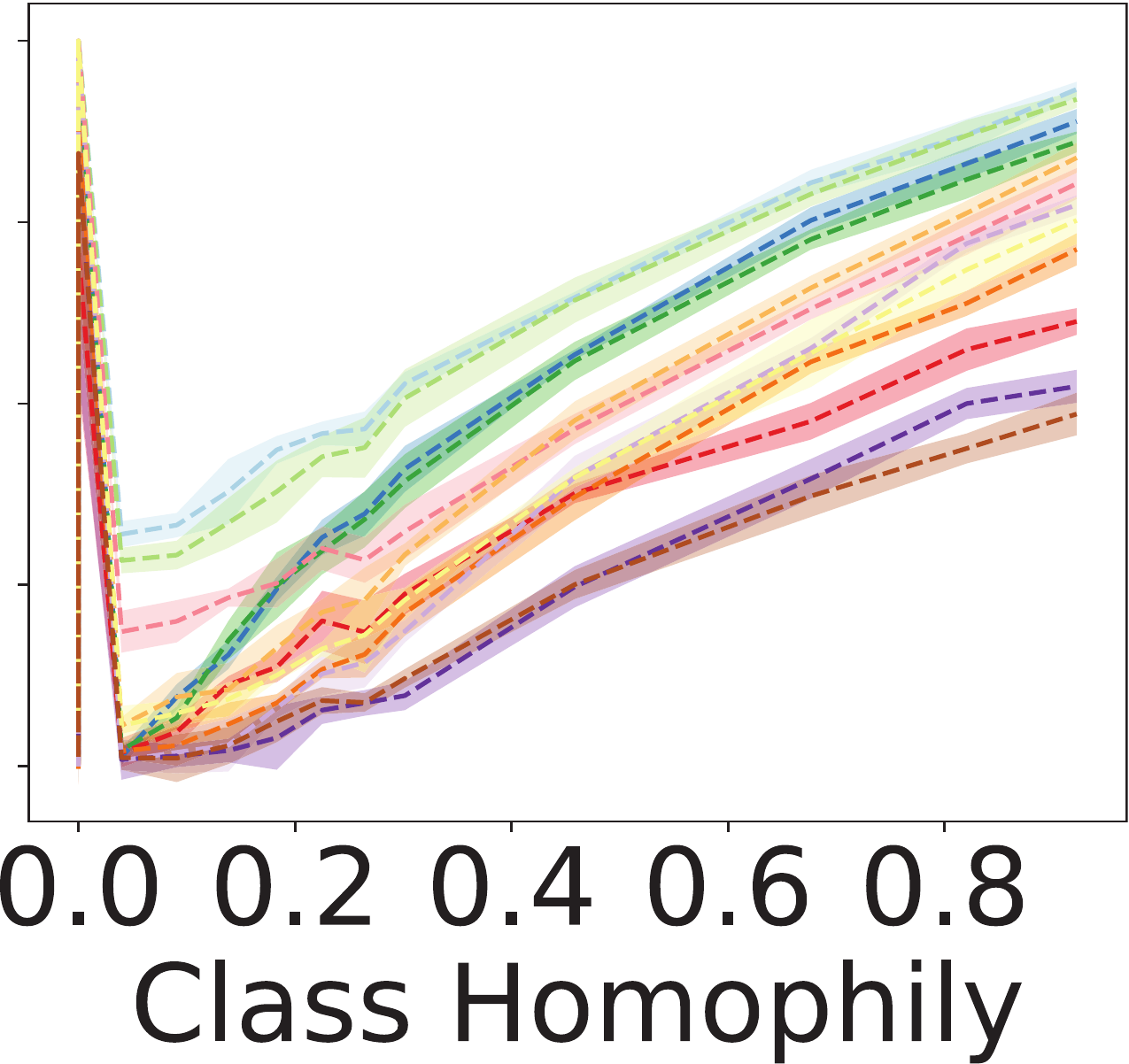}
     } 
     \subfloat[$H_{\text{agg}}^M(\mathcal{G})$]{
     \captionsetup{labelsep=newline,format=plain,indention=100pt}
     \includegraphics[height=0.11\textheight]{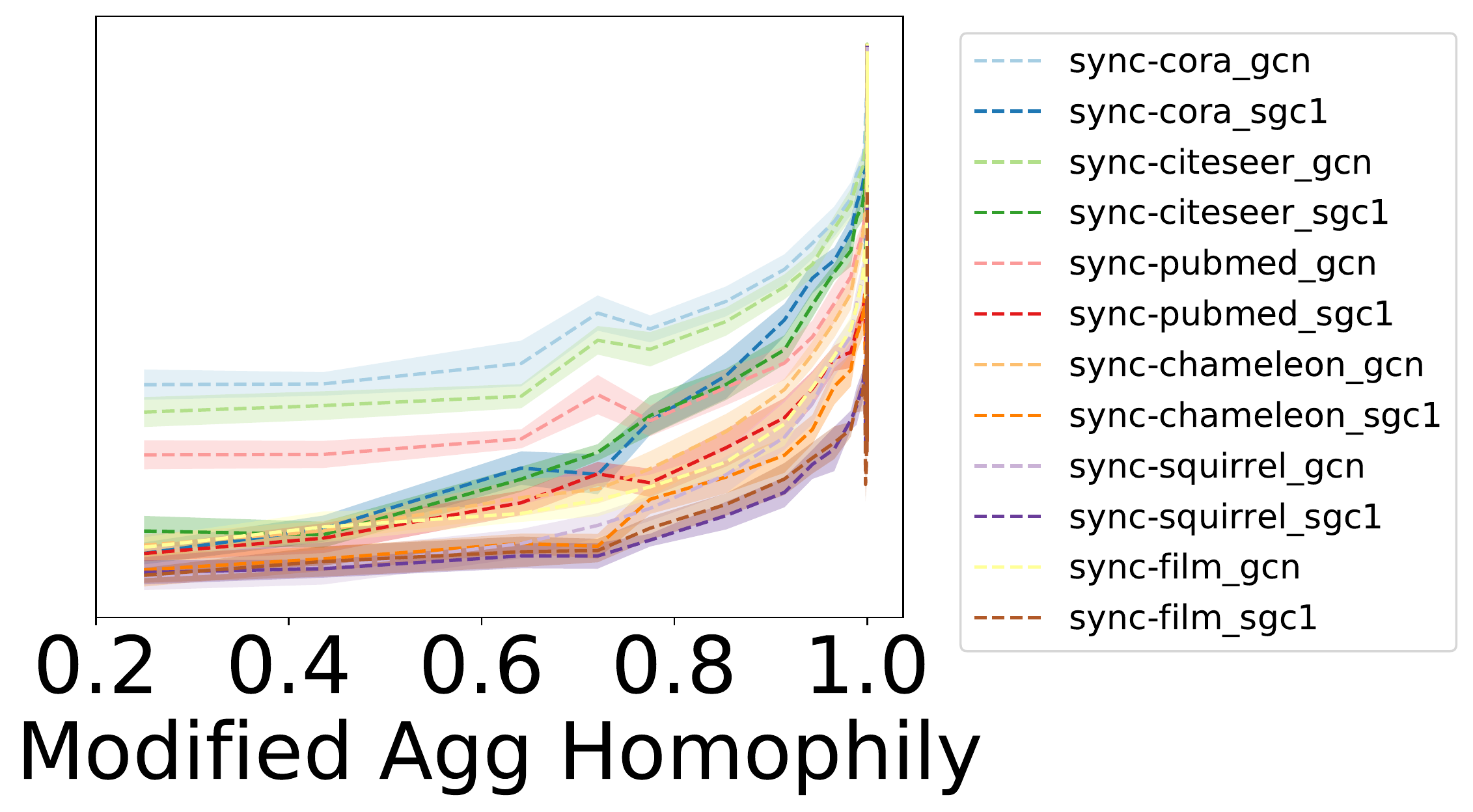}
     } 
     }
     \caption{Comparison of baseline performance under different homophily metrics.}
     \label{fig:comparison_homophily_metrics}
\end{figure}

\paragraph{Data Generation \& Experimental Setup}
We first generate 280 graphs in total with 28 edge homophily levels varied from 0.005 to 0.95, each corresponding to $10$ graphs. For every generated graph, we have 5 classes with 400 nodes in each class. For each node, we randomly generate 2 intra-class edges and [$\frac{2}{h} -2$] inter-class edges (see the details about the data generation process in appendix \ref{appendix:details_syn_exps}). The features of nodes in each class are sampled from node features in the corresponding class of the base dataset. Nodes are randomly split into 60\%/20\%/20\% for train/validation/test. We train 1-hop SGC (\textit{sgc-1}) \cite{wu2019simplifying} and GCN \cite{kipf2016classification} on synthetic data  
(see appendix \ref{appendix:hyperparameter_space_synthetic_graphs} for hyperparameter searching range). For each value of $H_\text{edge}(\mathcal{G})$, we take the average test accuracy and standard deviation of runs over 10 generated graphs. For each generated graph, we also calculate its $H_{\text{node}}(\mathcal{G}), H_{\text{class}}(\mathcal{G})$ and $H_{\text{agg}}^M(\mathcal{G})$. Model performance with respect to different homophily values are shown in Figure \ref{fig:comparison_homophily_metrics}.
\paragraph{Comparison of Homophily Metrics}
The performance of SGC-1 and GCN are expected to be monotonically increasing with a proper and informative homophily metric. However, Figure \ref{fig:comparison_homophily_metrics}(a)(b)(c) show that the performance curves under $H_{\text{edge}}(\mathcal{G}), H_{\text{node}}(\mathcal{G})$ and $H_{\text{class}}(\mathcal{G})$ are $U$-shaped \footnote{A similar J-shaped curve is found in \cite{zhu2020beyond}, though using different data generation processes. It does not mention the insufficiency of edge homophily.}, while Figure \ref{fig:comparison_homophily_metrics}(d) 
reveals a nearly monotonic curve with a little perturbation around 1. This indicates that $H_{\text{agg}}^M(\mathcal{G})$ can describe how the graph structure affects the performance of SGC-1 and GCN more appropriately and adequately than the existing metrics.

In addition, we notice that in Figure \ref{fig:comparison_homophily_metrics}(a), both SGC-1 and GCN get the worst performance on all datasets when $H_{\text{edge}}(\mathcal{G})$ is around somewhere between 0.1 and 0.2. This interesting phenomenon can be explained by the following theorem based on the similarity matrix which can verify the usefulness of $H_{\text{agg}}^M(\mathcal{G})$.
\begin{theorem} 1
(See Appendix \ref{appendix:proof_theorem1} for proof). Suppose there are $C$ classes in the graph $\cal G$ and $\cal G$ is a $d$-regular graph (each node has $d$ neighbors). Given $d$, edges for each node are \iid generated, such that each edge of any node has probability $h$ to connect with nodes in the same class and probability $1-h$ to connect with nodes in different classes. Let the aggregation operator $\hat{A} = \hat{A}_\text{rw}$. Then, for nodes $v$, $u_1$ and $u_2$, where $Z_{u_1,:}=Z_{v,:}$ and $Z_{u_2,:} \neq Z_{v,:}$, we have
\begin{equation}
\label{eq:theorem1_expectation_differences}
    g(h)\equiv 
    \mathbb{E}\left(S(\hat{A},Z)_{v,u_1}  \right) 
    - \mathbb{E}\left(S(\hat{A},Z)_{v,u_2}  \right)
   = \left(\frac{(C-1)(hd+1)-(1-h)d}{(C-1)(d+1)} \right)^2
\end{equation}
and the minimum of $g(h)$ is reached at 
$$
h=\frac{d+1-C}{Cd} = \frac{d_{\text{intra}}/h + 1 -C}{C (d_{\text{intra}}/h)}  \Rightarrow h=\frac{d_{\text{intra}}}{C d_{\text{intra}} +C-1}
$$
where $d_{\text{intra}}=dh$, which is the expected number of neighbors of a node that have the same label as the node.
\end{theorem}
The value of $g(h)$ in \eqref{eq:theorem1_expectation_differences} is the expected differences of the similarity values between nodes in the same class as $v$ and nodes in other classes. $g(h)$ is strongly related to the definition of aggregation homophily and its minimum potentially implies the worst value of $H_{\text{agg}}(\mathcal{G})$. In the synthetic experiments, we have $d_{\text{intra}} = 2, C=5$ and the minimum of $g(h)$ is reached at $h = 1/7 \approx 0.14$, which corresponds to the lowest point in the performance curve in Figure \ref{fig:comparison_homophily_metrics}(a). In other words, the $h$ where SGC-1 and GCN perform worst is where $g(h)$ gets the smallest value, instead of the point with the smallest edge homophily value $h=0$. This again reveals the advantage of $H_{\text{agg}}(\mathcal{G})$ over $H_\text{edge}(\mathcal{G})$ by taking use of the similarity matrix.

\begin{figure}[htbp]
\centering
{
\captionsetup{justification = centering}
\includegraphics[width=1\textwidth]{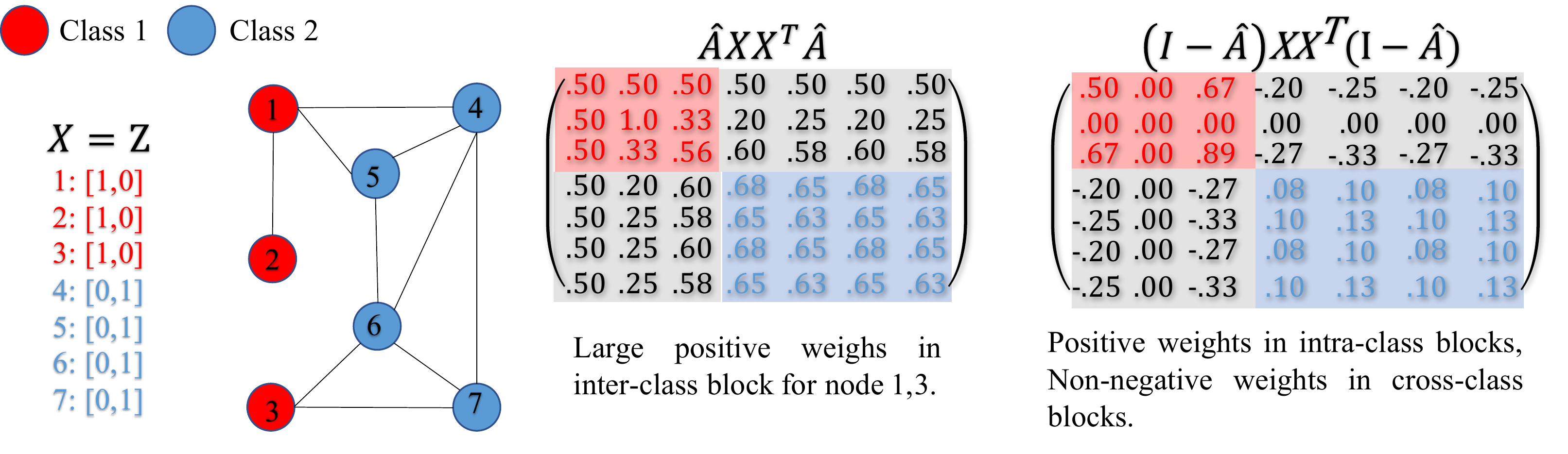}}
{%
  \caption{Example of how HP filter addresses harmful heterophily}%
  \label{fig:successful_example_hp_filter}
}
\end{figure}
\section{Adaptive Channel Mixing (ACM) Framework}
\label{sec:acm_framework}
Besides the new homophily metrics, in this section, we will also figure out how diversification operation (high-pass filter) is potentially capable to address some cases of harmful heterophily based on the similarity matrix proposed in equation \eqref{eq:gradient_descent_update}. From the analysis, we argue that low-pass filter and high-pass filter should be combined together for feature extraction, which lead us to the filterbank method in subsection \ref{sec:filterbank_spectral_spatial}. We generalize filterbank method and propose ACM framework in subsection \ref{sec:acm_gnn_architecture}.
 
\subsection{How Diversification Operation Helps with Harmful Heterophily}

We first consider the example shown in Figure \ref{fig:successful_example_hp_filter}. From $S(\hat{A},X)$, nodes 1,3 assign relatively large positive weights to nodes in class 2 after aggregation, which will 
make node 1,3 hard to be distinguished from nodes in class 2. Despite the fact, we can still 
distinguish between nodes 1,3 and 4,5,6,7 by considering 
their neighborhood difference: nodes 1,3 are different from most of their neighbors while nodes 4,5,6,7 are similar to most of their neighbors. This indicates, in some cases, although some nodes become similar after aggregation, they are still distinguishable via their surrounding dissimilarities. This leads us to use \textit{diversification operation}, \ie{} high-pass (HP) filter $I-\hat{A}$ \cite{ekambaram2014graph} (will be introduced in the next subsection) to extract the information of neighborhood differences and address harmful heterophily. As $S(I-\hat{A},X)$ in Figure \ref{fig:successful_example_hp_filter} shows, nodes 1,3 will assign negative weights to nodes 4,5,6,7 after diversification operation, 
\ie{} nodes 1,3 treat nodes 4,5,6,7 as negative samples and will move away from them during backpropagation. Base on this example, we first propose diversification distinguishability as follows to measures the proportion of nodes that diversification operation is potentially helpful for, 
\begin{definition} Diversification Distinguishability (DD) based on $S(I-\hat{A},X)$.

Given $S(I-\hat{A},X)$, a node $v$ is diversification distinguishable if the following two conditions are satisfied at the same time,
\vspace*{-1mm}
\begin{equation}
\label{eq:diversification_distinguishability}
\begin{split}
    \textbf{1.}\ \mathrm{Mean}_u \left(\{S(I-\hat{A},X)_{v,u}|u \in \mathcal{V} \land Z_{u,:}=Z_{v,:}\} \right) \geq 0; \\
    \textbf{2.}\ \mathrm{Mean}_u \left(\{S(I-\hat{A},X)_{v,u}|u \in \mathcal{V} \land Z_{u,:} \neq Z_{v,:}\} \right) \leq 0
    \end{split}
\end{equation}
Then, graph diversification distinguishability value is defined as
\begin{equation}
    \mathrm{DD}_{\hat{A},X}(\mathcal{G}) = \frac{1}{\left|\mathcal{V}\right|} \Big|\{v|v \mbox{ is diversification distinguishable}\}\Big|
\end{equation}
\end{definition}
\vspace*{-1mm}
We can see that $\mathrm{DD}_{\hat{A},X}(\mathcal{G}) \in [0,1]$ . The effectiveness of diversification operation can be proved for binary classification problems under certain conditions based on definition $2$, leading us to:
\begin{theorem} 2
(See Appendix \ref{appendix:proof_theorem2} for proof). 
Suppose $X=Z, \hat{A}=\hat{A}_{\text{rw}}$.
Then, for a binary classification problem, \ie{} $C=2$, all nodes are diversification distinguishable, \ie{} $\mathrm{DD}_{\hat{A},Z}(\mathcal{G})=1$.
\end{theorem}
Theorem 2 theoretically demonstrates the importance of diversification operation to extract high-frequency information of graph signal \cite{ekambaram2014graph}. Combined with aggregation operation, which is a low-pass filter \cite{ekambaram2014graph,maehara2019revisiting}, we can get a filterbank which uses both aggregation and diversification operations to distinctively extract the low- and high-frequency information from graph signals. We will introduce filterbank in the next subsection.

\subsection{Filterbank in Spectral and Spatial Forms}
\label{sec:filterbank_spectral_spatial}
\paragraph{Filterbank} For the graph signal $\bm{x}$ defined on $\mathcal{G}$, a 2-channel linear (analysis) filterbank \cite{ekambaram2014graph} \footnote{In graph signal processing, an additional synthesis filter \cite{ekambaram2014graph} is required to form the 2-channel filterbank. But synthesis filter is not needed in our framework, so we do not introduce it in our paper.} includes a pair of low-pass(LP) and high-pass(HP) filters $H_\text{LP}, H_\text{HP}$, where $H_\text{LP}$ and $H_\text{HP}$ retain the low-frequency and high-frequency content of $\bm{x}$, respectively.

Most existing GNNs are under uni-channel filtering architecture \cite{kipf2016classification,velivckovic2017attention,hamilton2017inductive} with either $H_\text{LP}$ or $H_\text{HP}$ channel that only partially preserves the input information. Unlike the uni-channel architecture, filterbanks with $H_\text{LP} + H_\text{HP} = I$ will not lose any information of the input signal, \ie{} perfect reconstruction property \cite{ekambaram2014graph}.

Generally, the Laplacian matrices ($L_\text{sym}$, $L_\text{rw}$, $\hat{L}_\text{sym}$, $\hat{L}_\text{rw}$) can be regarded as HP filters \cite{ekambaram2014graph} and affinity matrices ($A_\text{sym}$, $A_\text{rw}$, $\hat{A}_\text{sym}$, $\hat{A}_\text{rw}$) can be treated as LP filters \cite{maehara2019revisiting, hamilton2020graph}. Moreover, MLPs can be considered as owing a special identity filterbank with matrix $I$ that satisfies $H_\text{LP} + H_\text{HP} = I+0 = I$.
\paragraph{Filterbank in Spatial Form}
Filterbank methods can also be extended to spatial GNNs. Formally, on the node level, left multiplying $H_\text{LP}$ and $H_\text{HP}$ on $\bm{x}$ performs as aggregation and diversification operations, respectively. For example, suppose $H_\text{LP} = \hat{A}$ and $H_\text{HP}=I-\hat{A}$, then for node $i$ we have
\begin{equation}
\label{eq:spatial_form_filterbank}
    (H_\text{LP} \bm{x})_i =  \sum_{j \in \{\mathcal{N}_i \cup i\} } \hat{A}_{i,j} \bm{x_j}, \ (H_\text{HP} \bm{x})_i = \bm{x_i} - \sum_{j \in \{\mathcal{N}_i \cup i\} } \hat{A}_{i,j} \bm{x_j}
\end{equation} 
where $\hat{A}_{i,j}$ is the connection weight between two nodes. To leverage HP and identity channels in GNNs, we propose the Adaptive Channel Mixing (ACM) architecture in the following subsection.
\subsection{Adaptive Channel Mixing(ACM) GNN  Framework}
\label{sec:acm_gnn_architecture}
ACM framework can be applied to lots of baseline GNNs and in this subsection, we use GCN as an example and introduce ACM framework in matrix form. We use $H_\text{LP}$ and $H_\text{HP}$ to represent general LP and HP filters. The ACM framework includes $3$ steps as follows,
\vspace*{-1.5mm}
\begin{equation}
\begin{aligned}
\label{eq:acm_gnn_spectral}
& \textbf{{Step 1. Feature Extraction for Each Channel:}} \\
& \text{Option 1: } {H}^{l}_L = \text{ReLU}\left(H_\text{LP} {H^{l-1}} W^{l-1}_L\right), \ {{H}^{l}_H} =  \text{ReLU}\left(H_\text{HP} {H^{l-1}} W^{l-1}_H\right), {H}^{l}_I  = \ \text{ReLU}\left(I {H^{l-1}} W^{l-1}_I\right); \\
& \text{Option 2: } {H}^{l}_L = H_\text{LP} \text{ReLU}\left({H^{l-1}} W^{l-1}_L\right), \ {{H}^{l}_H} = H_\text{HP} \text{ReLU}\left({H^{l-1}} W^{l-1}_H\right), {H}^{l}_I  = I\ \text{ReLU}\left({H^{l-1}} W^{l-1}_I\right); \\
& W_L^{l-1},\ W_H^{l-1}, \ W_I^{l-1} \in \mathbb{R}^{F_{l-1} \times F_l}; \\
& \textbf{Step 2. Feature-based Weight Learning} \\
&\tilde{\alpha}_L^l = \sigma\left({H}^{l}_L \tilde{W}^{l}_L\right),\ \tilde{\alpha}_H^l = \sigma \left({H}^{l}_H \tilde{W}^{l}_H\right),\ \tilde{\alpha}_I^l = \sigma \left({H}^{l}_I \tilde{W}^{l}_I\right),\ \tilde{W}_L^{l-1},\ \tilde{W}_H^{l-1},\ \tilde{W}_I^{l-1} \in \mathbb{R}^{F_l \times 1}\\ 
& \left[{\alpha}_L^l, {\alpha}_H^l, {\alpha}_I^l \right] = \text{Softmax}\left(\left[\tilde{\alpha}_L^l,\tilde{\alpha}_H^l,\tilde{\alpha}_I^l\right]W_\text{Mix}^l/T, \right),\ W_\text{Mix}^l \in \mathbb{R}^{3\times 3}, T \in \mathbb{R} \text{ is the temperature} ; \\
&\textbf{Step 3. Node-wise Channel Mixing:}\\
& {H^{l}}  = \left( \text{diag}(\alpha_L^l){H}^{l}_L + \text{diag}(\alpha_H^l){H}^{l}_H + \text{diag}(\alpha_I^l){H}^{l}_I \right).
\end{aligned}
\end{equation}
The framework with option 1 in step 1 is ACM framework and with option 2 is ACMII framework. ACM(II)-GCN first implement distinct feature extractions for $3$ channels, respectively. After processed by a set of filterbanks, $3$ filtered components $H_L^l,H_H^l, H_I^l$ are obtained. Different nodes may have different needs for the information in the 3 channels, \eg{} in Figure \ref{fig:successful_example_hp_filter}, nodes 1,3 demand high-frequency information while node 2 only needs low-frequency information. To adaptively exploit information from different channels, ACM(II)-GCN learns row-wise (node-wise) feature-conditioned weights to combine the $3$ channels. ACM(II) can be easily plugged into spatial GNNs by replacing $H_\text{LP}$ and $H_\text{HP}$ by aggregation and diversification operations as shown in \eqref{eq:spatial_form_filterbank}.  See Appendix \ref{appendix:model_comparison_synthetic_datasets} for a detailed discussion of model comparison on synthetic datasets.
\vspace*{-2.5mm}
\paragraph{Complexity} Number of learnable parameters in layer $l$ of ACM(II)-GCN is $3F_{l-1}(F_l +1)+9$, while it is $F_{l-1}F_l$ in GCN. The computation of step 1-3 takes $NF_l(8+6F_{l-1}) + 2F_l(\text{nnz}(H_\text{LP}) + \text{nnz}(H_\text{HP}))+ 18N$ flops, while GCN layer takes $2NF_{l-1}F_l + 2F_l(\text{nnz}(H_\text{LP}))$ flops, where $\text{nnz}(\cdot)$ is the number of non-zero elements. A detailed comparison on running time is conducted in section \ref{sec:ablation_tests_running_time}.
\vspace*{-2.5mm}
\paragraph{Limitations} Diversification operation does not work well in all harmful heterophily cases. For example, consider an imbalanced dataset where several small clusters with distinctive labels are densely connected to a large cluster. In this case, the surrounding differences of nodes in small clusters are similar, \ie{} the neighborhood differences are mainly from their connection to the same large cluster, and this possibly makes diversification operation fail to discriminate them. See a more detailed demonstration and discussion in Appendix \ref{appendix:limitation_diversification}.

\vspace*{-2.5mm}
\section{Prior Work}
\label{sec:related_works}
\paragraph{GNNs on Addressing Heterophily}
We discuss relevant work of GNNs on addressing heterophily challenge in this part. \cite{abu2019mixhop} acknowledges the difficulty of learning on graphs with weak homophily and propose MixHop to extract features from multi-hop neighborhood to get more information. Geom-GCN \cite{pei2020geom} precomputes unsupervised node embeddings and uses graph structure defined by geometric relationships in the embedding space to define the bi-level aggregation process. \cite{hou2019measuring} proposes measurements based on feature smoothness and label smoothness that are potentially helpful to guide GNNs on dealing with heterophilous graphs. H$_2$GCN \cite{zhu2020beyond} combines 3 key designs to address heterophily: (1) ego- and neighbor-embedding separation; (2) higher-order neighborhoods; (3) combination of intermediate representations. CPGNN \cite{zhu2020graph} models label correlations by the compatibility matrix, which is beneficial for heterophily settings, and propagates a prior belief estimation into GNNs by the compatibility matrix. FBGNN \cite{luan2020complete} first proposes to use filterbank to address heterophily problem, but it does not fully explain the insights behind HP filters and does not contain identity channel and node-wise channel mixing mechanism. FAGCN \cite{bo2021beyond} learns edge-level aggregation weights as GAT \cite{velivckovic2017attention} but allows the weights to be negative which enables the network to capture the high-frequency components in graph signals. GPRGNN \cite{chien2021adaptive} uses learnable weights that can be both positive and negative for feature propagation, it allows GRPGNN to adapt heterophily structure of graph and is able to handle both high- and low-frequency parts of the graph signals.
\vspace*{-2.5mm}
\paragraph{GNNs with Filterbanks}
Previously, there are geometric scattering networks \cite{gao2019geometric, min2020scattering} that apply filterbanks to address over-smoothing \cite{li2018deeper} problem. The scattering construction captures different channels of variation from node features or labels. In geometric learning and graph signal processing, the band-pass filtering operations extract geometric information beyond smooth signals, thus it is believed that filterbanks can alleviate over-smoothing in GNNs. 
In ACM framework, we aim to design a framework with the help of filterbanks to adaptively utilize different channels to address the challenge of learning on heterophilous graph. We deal with different problem from \cite{gao2019geometric, min2020scattering}. 
\vspace*{-2.5mm}
\section{Experiments on Real-World Datasets}
\label{sec:experiments}
In this section, we evaluate ACM(II) framework on real-world datasets. We first conduct ablation studies in subsection \ref{sec:ablation_tests_running_time} to validate different components. Then, we compare with the state-of-the-arts (SOTA) models in subsection \ref{sec:comparison_with_sota}.

\subsection{Ablation Study \& Efficiency}
\label{sec:ablation_tests_running_time}
\begin{table}[htbp]
  \centering
  \tiny
  \caption{Ablation study on 9 real-world datasets \cite{pei2020geom}. Cell with \checkmark means the component is applied to the baseline model. The best test results are highlighted.}
  \label{tab:ablation_study}
  \setlength{\tabcolsep}{1pt}
    \begin{tabular}{c|cccc|ccccccccc|r}
    \toprule
    \toprule
    \multicolumn{14}{c}{Ablation Study on Different Components in ACM-SGC and ACM-GCN (\%)}                       &  \\
    \midrule
    \multicolumn{1}{p{4.125em}|}{Baseline    } & \multicolumn{4}{c|}{Model Components} & Cornell & Wisconsin & Texas & Film  & Chameleon & Squirrel & Cora  & CiteSeer & PubMed & \multicolumn{1}{c}{\multirow{1}[4]{*}{Rank}} \\
\cmidrule{2-14}    \multicolumn{1}{p{4.125em}|}{Models} & LP    & HP    & Identity & Mixing & Acc $\pm$ Std & Acc $\pm$ Std & Acc $\pm$ Std & Acc $\pm$ Std & Acc $\pm$ Std & Acc $\pm$ Std & Acc $\pm$ Std & Acc $\pm$ Std & Acc $\pm$ Std &  \\
    \midrule
    \multicolumn{1}{c|}{\multirow{5}[1]{*}{SGC-1 w/}} & $\checkmark$ &       &       &       & 70.98 $\pm$ 8.39 & 70.38 $\pm$ 2.85 & 83.28 $\pm$ 5.43 & 25.26 $\pm$ 1.18 & 64.86 $\pm$ 1.81 & 47.62 $\pm$ 1.27 & 85.12 $\pm$ 1.64 & 79.66 $\pm$ 0.75 & 85.5 $\pm$ 0.76 & \multicolumn{1}{c}{12.89} \\
          & $\checkmark$ & $\checkmark$ &       & $\checkmark$ & 83.28 $\pm$ 5.81 & 91.88 $\pm$ 1.61 & 90.98 $\pm$ 2.46 & 36.76 $\pm$ 1.01 & 65.27 $\pm$ 1.9 & 47.27 $\pm$ 1.37 & 86.8 $\pm$ 1.08 & 80.98 $\pm$ 1.68 & 87.21 $\pm$ 0.42 & \multicolumn{1}{c}{10.44} \\
          & $\checkmark$ &       & $\checkmark$ & $\checkmark$ & 93.93 $\pm$ 3.6 & 95.25 $\pm$ 1.84 & 93.93 $\pm$ 2.54 & 38.38 $\pm$ 1.13 & 63.83 $\pm$ 2.07 & 46.79 $\pm$ 0.75 & 86.73 $\pm$ 1.28 & 80.57 $\pm$ 0.99 & 87.8 $\pm$ 0.58 & \multicolumn{1}{c}{9.44} \\
          & $\checkmark$ & $\checkmark$ & $\checkmark$ &       & 88.2 $\pm$ 4.39 & 93.5 $\pm$ 2.95 & 92.95 $\pm$ 2.94 & 37.19 $\pm$ 0.87 & 62.82 $\pm$ 1.84 & 44.94 $\pm$ 0.93 & 85.22 $\pm$ 1.35 & 80.75 $\pm$ 1.68 & 88.11 $\pm$ 0.21 & \multicolumn{1}{c}{11.00} \\
          & $\checkmark$ & $\checkmark$ & $\checkmark$ & $\checkmark$ & 93.77 $\pm$ 1.91 & 93.25 $\pm$ 2.92 & 93.61 $\pm$ 1.55 & 39.33 $\pm$ 1.25 & 63.68 $\pm$ 1.62 & 46.4 $\pm$ 1.13 & 86.63 $\pm$ 1.13 & 80.96 $\pm$ 0.93 & 87.75 $\pm$ 0.88 & \multicolumn{1}{c}{10.00} \\
          \midrule
    \multicolumn{1}{c|}{\multirow{5}[0]{*}{ACM-GCN w/}} & $\checkmark$ &       &       &       & 82.46 $\pm$ 3.11 & 75.5 $\pm$ 2.92 & 83.11 $\pm$ 3.2 & 35.51 $\pm$ 0.99 & 64.18 $\pm$ 2.62 & 44.76 $\pm$ 1.39 & 87.78 $\pm$ 0.96 & 81.39 $\pm$ 1.23 & 88.9 $\pm$ 0.32 & \multicolumn{1}{c}{11.44} \\
          & $\checkmark$ & $\checkmark$ &       & $\checkmark$ & 82.13 $\pm$ 2.59 & 86.62 $\pm$ 4.61 & 89.19 $\pm$ 3.04 & 38.06 $\pm$ 1.35 & \cellcolor[rgb]{ .816,  .808,  .808}\textbf{69.21 $\pm$ 1.68} & 57.2 $\pm$ 1.01 & 88.93 $\pm$ 1.55 & \cellcolor[rgb]{ .816,  .808,  .808}\textbf{81.96 $\pm$ 0.91} & 90.01 $\pm$ 0.8  & \multicolumn{1}{c}{7.22} \\
          & $\checkmark$ &       & $\checkmark$ & $\checkmark$ & 94.26 $\pm$ 2.23 & 96.13 $\pm$ 2.2 & 94.1 $\pm$ 2.95 & 41.51 $\pm$ 0.99 & 67.44 $\pm$ 2.14 & 53.97 $\pm$ 1.39 & 88.95 $\pm$ 0.9 & 81.72 $\pm$ 1.22 & 90.88 $\pm$ 0.55 & \multicolumn{1}{c}{4.44} \\
          & $\checkmark$ & $\checkmark$ & $\checkmark$ &       & 91.64 $\pm$ 2 & 95.37 $\pm$ 3.31 & \cellcolor[rgb]{ .816,  .808,  .808}\textbf{95.25 $\pm$ 2.37} & 40.47 $\pm$ 1.49 & 68.93 $\pm$ 2.04 & 54.78 $\pm$ 1.27 & \cellcolor[rgb]{ .816,  .808,  .808}\textbf{89.13 $\pm$ 1.77} & \cellcolor[rgb]{ .816,  .808,  .808}\textbf{81.96 $\pm$ 2.03} & \cellcolor[rgb]{ .816,  .808,  .808}\textbf{91.01 $\pm$ 0.7} & \multicolumn{1}{c}{3.11} \\
          & $\checkmark$ & $\checkmark$ & $\checkmark$ & $\checkmark$ & 94.75 $\pm$ 2.62 & \cellcolor[rgb]{ .816,  .808,  .808}\textbf{96.75 $\pm$ 1.6} & 95.08 $\pm$ 3.2 & 41.62 $\pm$ 1.15 & 69.04 $\pm$ 1.74 & \cellcolor[rgb]{ .816,  .808,  .808}\textbf{58.02 $\pm$ 1.86} & 88.95 $\pm$ 1.3 & 81.80 $\pm$ 1.26 & 90.69 $\pm$ 0.53 & \multicolumn{1}{c}{\cellcolor[rgb]{ .816,  .808,  .808}\textbf{2.78}} \\
          \midrule
    \multicolumn{1}{c|}{\multirow{4}[1]{*}{ACMII-GCN w/}} & $\checkmark$ & $\checkmark$ &       & $\checkmark$ & 82.46 $\pm$ 3.03 & 91.00 $\pm$ 1.75 & 90.33 $\pm$ 2.69 & 38.39 $\pm$ 0.75 & 67.59 $\pm$ 2.14 & 53.67 $\pm$ 1.71 & \cellcolor[rgb]{ .816,  .808,  .808}\textbf{89.13 $\pm$ 1.14} & 81.75 $\pm$ 0.85 & 89.87 $\pm$ 0.39 & \multicolumn{1}{c}{7.44} \\
          & $\checkmark$ &       & $\checkmark$ & $\checkmark$ & 94.26 $\pm$ 2.57 & 96.00 $\pm$ 2.15 & 94.26 $\pm$ 2.96 & 40.96 $\pm$ 1.2 & 66.35 $\pm$ 1.76 & 50.78 $\pm$ 2.07 & 89.06 $\pm$ 1.07 & 81.86 $\pm$ 1.22 & 90.71 $\pm$ 0.67 & \multicolumn{1}{c}{4.67} \\
          & $\checkmark$ & $\checkmark$ & $\checkmark$ &       & 91.48 $\pm$ 1.43 & 96.25 $\pm$ 2.09 & 93.77 $\pm$ 2.91 & 40.27 $\pm$ 1.07 & 66.52 $\pm$ 2.65 & 52.9 $\pm$ 1.64 & 88.83 $\pm$ 1.16 & 81.54 $\pm$ 0.95 & 90.6 $\pm$ 0.47 & \multicolumn{1}{c}{6.67} \\
          & $\checkmark$ & $\checkmark$ & $\checkmark$ & $\checkmark$ & \cellcolor[rgb]{ .816,  .808,  .808}\textbf{95.9 $\pm$ 1.83} & 96.62 $\pm$ 2.44 & 95.25 $\pm$ 3.15 & \cellcolor[rgb]{ .816,  .808,  .808}\textbf{41.84 $\pm$ 1.15} & 68.38 $\pm$ 1.36 & 54.53 $\pm$ 2.09 & 89.00 $\pm$ 0.72 & 81.79 $\pm$ 0.95 & 90.74 $\pm$ 0.5 & \multicolumn{1}{c}{\cellcolor[rgb]{ .816,  .808,  .808}\textbf{2.78}} \\
\cmidrule{1-14}    \multicolumn{14}{c|}{Comparison of Average Running Time Per Epoch(ms)}                                        &  \\
\cmidrule{1-14}    \multicolumn{1}{c|}{\multirow{5}[2]{*}{SGC-1 w/}} & $\checkmark$ &       &       &       & 2.53  & 2.83  & 2.5   & 3.18  & 3.48  & 4.65  & 3.47  & 3.43  & 4.04  &  \\
          & $\checkmark$ & $\checkmark$ &       & $\checkmark$ & 4.01  & 4.57  & 4.24  & 4.55  & 4.76  & 5.09  & 5.39  & 4.69  & 4.75  &  \\
          & $\checkmark$ &       & $\checkmark$ & $\checkmark$ & 3.88  & 4.01  & 4.04  & 4.43  & 4.06  & 4.5   & 4.38  & 3.82  & 4.16  &  \\
          & $\checkmark$ & $\checkmark$ & $\checkmark$ &       & 3.31  & 3.49  & 3.18  & 3.7   & 3.53  & 4.83  & 3.92  & 3.87  & 4.24  &  \\
          & $\checkmark$ & $\checkmark$ & $\checkmark$ & $\checkmark$ & 5.53  & 5.96  & 5.43  & 5.21  & 5.41  & 6.96  & 6     & 5.9   & 6.04  &  \\
\cmidrule{1-14}    \multicolumn{1}{c|}{\multirow{5}[2]{*}{ACM-GCN w/}} & $\checkmark$ &       &       &       & 3.67  & 3.74  & 3.59  & 4.86  & 4.96  & 6.41  & 4.24  & 4.18  & 5.08  &  \\
          & $\checkmark$ & $\checkmark$ &       & $\checkmark$ & 6.63  & 8.06  & 7.89  & 8.11  & 7.8   & 9.39  & 7.82  & 7.38  & 8.74  &  \\
          & $\checkmark$ &       & $\checkmark$ & $\checkmark$ & 5.73  & 5.91  & 5.93  & 6.86  & 6.35  & 7.15  & 7.34  & 6.65  & 6.8   &  \\
          & $\checkmark$ & $\checkmark$ & $\checkmark$ &       & 5.16  & 5.25  & 5.2   & 5.93  & 5.64  & 8.02  & 5.73  & 5.65  & 6.16  &  \\
          & $\checkmark$ & $\checkmark$ & $\checkmark$ & $\checkmark$ & 8.25  & 8.11  & 7.89  & 7.97  & 8.41  & 11.9  & 8.84  & 8.38  & 8.63  &  \\
\cmidrule{1-14}    \multicolumn{1}{c|}{\multirow{4}[2]{*}{ACMII-GCN w/}} & $\checkmark$ & $\checkmark$ &       & $\checkmark$ & 6.62  & 7.35  & 7.39  & 7.62  & 7.33  & 9.69  & 7.49  & 7.58  & 7.97  &  \\
          & $\checkmark$ &       & $\checkmark$ & $\checkmark$ & 6.3   & 6.05  & 6.26  & 6.87  & 6.44  & 6.5   & 6.14  & 7.21  & 6.6   &  \\
          & $\checkmark$ & $\checkmark$ & $\checkmark$ &       & 5.24  & 5.27  & 5.46  & 5.72  & 5.65  & 7.87  & 5.48  & 5.65  & 6.33  &  \\
          & $\checkmark$ & $\checkmark$ & $\checkmark$ & $\checkmark$ & 7.59  & 8.28  & 8.06  & 8.85  & 8     & 10    & 8.27  & 8.5   & 8.68  &  \\
    \bottomrule
    \bottomrule
    \end{tabular}%
\end{table}%

We investigate the effectiveness and efficiency of adding HP, identity channels and the adaptive mixing mechanism in ACM(II) framework by ablation study. Specifically, we apply the above components to SGC-1 and GCN separately, run 10 times on each dataset with 60\%/20\%/20\% random splits for train/validation/test used in \cite{chien2021adaptive} and report the average test accuracy as well as the standard deviation. We also record the average running time per epoch (in milliseconds) to compare the efficiency. We set the temperature $T$ in \eqref{eq:acm_gnn_spectral} to be $3$. (See Appendix \ref{appendix:hyperparameter} for hyperparameter searching range.)

From the results we can see that on most datasets, the additional HP and identity channels are helpful, even on strong homophily datasets, such as Cora, CiteSeer and PubMed. The adaptive mixing mechanism also shows its advantage over the method that directly adds the three channels together. This illustrates the necessity of learning to customize the channel usage adaptively for different nodes. As for efficiency, we can see that the running time is approximately doubled under ACM(II) framework than the original model.
\subsection{Comparison with State-of-the-art Models}
\label{sec:comparison_with_sota}
\begin{table}[htbp]
  \centering
  \tiny
  \setlength{\tabcolsep}{0.5pt}
  \caption{Experimental results: average test accuracy $\pm$ standard deviation on $10$ real-world benchmark datasets. The best results are highlighted. Results "*" are reported from \cite{chien2021adaptive,lim2021new} and results "$^\dagger$" are from \cite{pei2020geom}. NA means the reported results are not available and OOM means out of memory. }
    \begin{tabular}{c|cccccccccc|cc}
    \toprule
    \toprule
         & Cornell & Wisconsin & Texas & Film  & Chameleon & Squirrel & Deezer-Europe & Cora  & CiteSeer & PubMed &  \\
    \midrule
    \#nodes & 183   & 251   & 183   & 7,600 & 2,277 & 5,201 & 28,281 & 2,708 & 3,327 & 19,717 &  \\
    \#edges & 295   & 499   & 309   & 33,544 & 36,101 & 217,073 & 92,752 & 5,429 & 4,732 & 44,338 &  \\
    \#features & 1,703 & 1,703 & 1,703 & 931   & 2,325 & 2,089 & 31,241 & 1,433 & 3,703 & 500   &  \\
    \#classes & 5     & 5     & 5     & 5     & 5     & 5     & 2     & 7     & 6     & 3     &  \\
    H\_edge & 0.5669 & 0.4480 & 0.4106 & 0.3750 & 0.2795 & 0.2416 & 0.5251 & 0.8100 & 0.7362 & 0.8024 &  \\
    H\_node & 0.3855 & 0.1498 & 0.0968 & 0.2210 & 0.2470 & 0.2156 & 0.5299 & 0.8252 & 0.7175 & 0.7924 &  \\
    H\_class & 0.0468 & 0.0941 & 0.0013 & 0.0110 & 0.0620 & 0.0254 & 0.0304 & 0.7657 & 0.6270 & 0.6641 &  \\
    Data Splits & 60\%/20\%/20\% & 60\%/20\%/20\% & 60\%/20\%/20\% & 60\%/20\%/20\% & 60\%/20\%/20\% & 60\%/20\%/20\% & 50\%/25\%/25\% & 60\%/20\%/20\% & 60\%/20\%/20\% & 60\%/20\%/20\% &  \\
    H\_agg\^M(G) & 0.8032 & 0.7768 & 0.694 & 0.6822 & 0.61  & 0.3566 & 0.5790 & 0.9904 & 0.9826 & 0.9432 &  \\
    \midrule
    \midrule
          & \multicolumn{10}{c|}{Test Accuracy (\%) of State-of-the-art Models, Baseline GNN Models and ACM-GNN models} & Rank \\
    \midrule
    MLP-2* & 91.30 $\pm$ 0.70 & 93.87 $\pm$ 3.33 & 92.26 $\pm$ 0.71 & 38.58 $\pm$ 0.25 & 46.72 $\pm$ 0.46 & 31.28 $\pm$ 0.27 & 66.55 $\pm$ 0.72 & 76.44 $\pm$ 0.30 & 76.25 $\pm$ 0.28 & 86.43 $\pm$ 0.13 & 18.60 \\
    \midrule
    GAT*   & 76.00 $\pm$ 1.01 & 71.01 $\pm$ 4.66 & 78.87 $\pm$ 0.86 & 35.98 $\pm$ 0.23 & 63.9 $\pm$ 0.46 & 42.72 $\pm$ 0.33 & 61.09 $\pm$ 0.77 & 76.70 $\pm$ 0.42 & 67.20 $\pm$ 0.46 & 83.28 $\pm$ 0.12 & 21.40 \\
    APPNP* & 91.80 $\pm$ 0.63 & 92.00 $\pm$ 3.59 & 91.18 $\pm$ 0.70 & 38.86 $\pm$ 0.24 & 51.91 $\pm$ 0.56 & 34.77 $\pm$ 0.34 & 67.21 $\pm$ 0.56 & 79.41 $\pm$ 0.38 & 68.59 $\pm$ 0.30 & 85.02 $\pm$ 0.09 & 18.00 \\
    GPRGNN* & 91.36 $\pm$ 0.70 & 93.75 $\pm$ 2.37 & 92.92 $\pm$ 0.61 & 39.30 $\pm$ 0.27 & 67.48 $\pm$ 0.40 & 49.93 $\pm$ 0.53 & 66.90 $\pm$ 0.50 & 79.51pm 0.36 & 67.63 $\pm$ 0.38 & 85.07 $\pm$ 0.09 & 14.40 \\
    H2GCN & 86.23 $\pm$ 4.71 & 87.5 $\pm$ 1.77 & 85.90 $\pm$ 3.53 & 38.85 $\pm$ 1.17 & 52.30 $\pm$ 0.48 & 30.39 $\pm$ 1.22 & \cellcolor[rgb]{ .816,  .808,  .808}\textbf{67.22 $\pm$ 0.90} & 87.52 $\pm$ 0.61 & 79.97 $\pm$ 0.69 & 87.78 $\pm$ 0.28 & 17.00 \\
    MixHop & 60.33 $\pm$ 28.53 & 77.25 $\pm$ 7.80 & 76.39 $\pm$ 7.66 & 33.13 $\pm$ 2.40 & 36.28 $\pm$ 10.22 & 24.55 $\pm$ 2.60 & 66.80 $\pm$ 0.58 & 65.65 $\pm$ 11.31 & 49.52 $\pm$ 13.35 & 87.04 $\pm$ 4.10 & 23.50 \\
    GCN+JK & 66.56 $\pm$ 13.82 & 62.50 $\pm$ 15.75 & 80.66 $\pm$ 1.91 & 32.72 $\pm$ 2.62 & 64.68 $\pm$ 2.85 & 53.40 $\pm$ 1.90 & 60.99 $\pm$ 0.14 & 86.90 $\pm$ 1.51 & 73.77 $\pm$ 1.85 & 90.09 $\pm$ 0.68 & 18.80 \\
     GAT+JK & 74.43 $\pm$ 10.24 & 69.50 $\pm$ 3.12 & 75.41 $\pm$ 7.18 & 35.41 $\pm$ 0.97 & 68.14 $\pm$ 1.18 & 52.28 $\pm$ 3.61 & 59.66 $\pm$ 0.92 & 89.52 $\pm$ 0.43 & 74.49 $\pm$ 2.76 & 89.15 $\pm$ 0.87 & 16.70 \\
    FAGCN & 88.03 $\pm$ 5.6 & 89.75 $\pm$ 6.37 & 88.85 $\pm$ 4.39 & 31.59 $\pm$ 1.37 & 49.47 $\pm$ 2.84 & 42.24 $\pm$ 1.2 & 66.86 p, 0.53 & 88.85 $\pm$ 1.36 & \cellcolor[rgb]{ .816,  .808,  .808}\textbf{82.37 $\pm$ 1.46} & 89.98 $\pm$ 0.54 & 14.10 \\
    GraphSAGE & 71.41 $\pm$ 1.24 & 64.85 $\pm$ 5.14 & 79.03 $\pm$ 1.20 & 36.37 $\pm$ 0.21 & 62.15 $\pm$ 0.42 & 41.26 $\pm$ 0.26 & OOM   & 86.58 $\pm$ 0.26 & 78.24 $\pm$ 0.30 & 86.85 $\pm$ 0.11 & 20.89 \\
    Geom-GCN$^\dagger$ & 60.81 & 64.12 & 67.57 & 31.63 & 60.9  & 38.14 & NA    & 85.27 & 77.99 & 90.05 & 22.67 \\
    \midrule
    SGC-1 & 70.98 $\pm$ 8.39 & 70.38 $\pm$ 2.85 & 83.28 $\pm$ 5.43 & 25.26 $\pm$ 1.18 & 64.86 $\pm$ 1.81 & 47.62 $\pm$ 1.27 & 59.73 $\pm$ 0.12 & 85.12 $\pm$ 1.64 & 79.66 $\pm$ 0.75 & 85.5 $\pm$ 0.76 & 20.10 \\
    SGC-2 & 72.62 $\pm$ 9.92 & 74.75 $\pm$ 2.89 & 81.31 $\pm$ 3.3 & 28.81 $\pm$ 1.11 & 62.67 $\pm$ 2.41 & 41.25 $\pm$ 1.4 & 61.56 $\pm$ 0.51 & 85.48 $\pm$ 1.48 & 80.75 $\pm$ 1.15 & 85.36 $\pm$ 0.52 & 20.70 \\
    GCNII & 89.18 $\pm$ 3.96 & 83.25 $\pm$ 2.69 & 82.46 $\pm$ 4.58 & 40.82 $\pm$ 1.79 & 60.35 $\pm$ 2.7 & 38.81 $\pm$ 1.97 & 66.38 $\pm$ 0.45 & 88.98 $\pm$ 1.33 & 81.58 $\pm$ 1.3 & 89.8 $\pm$ 0.3 & 14.80 \\
    GCNII* & 90.49 $\pm$ 4.45 & 89.12 $\pm$ 3.06 & 88.52 $\pm$ 3.02 & 41.54 $\pm$ 0.99 & 62.8 $\pm$ 2.87 & 38.31 $\pm$ 1.3 & 66.42 $\pm$ 0.56 & 88.93 $\pm$ 1.37 & 81.83 $\pm$ 1.78 & 89.98 $\pm$ 0.52 & 12.30 \\
    GCN   & 82.46 $\pm$ 3.11 & 75.5 $\pm$ 2.92 & 83.11 $\pm$ 3.2 & 35.51 $\pm$ 0.99 & 64.18 $\pm$ 2.62 & 44.76 $\pm$ 1.39 & 62.23 $\pm$ 0.53 & 87.78 $\pm$ 0.96 & 81.39 $\pm$ 1.23 & 88.9 $\pm$ 0.32 & 16.30 \\
    Snowball-2 & 82.62 $\pm$ 2.34 & 74.88 $\pm$ 3.42 & 83.11 $\pm$ 3.2 & 35.97 $\pm$ 0.66 & 64.99 $\pm$ 2.39 & 47.88 $\pm$ 1.23 & OOM   & 88.64 $\pm$ 1.15 & 81.53 $\pm$ 1.71 & 89.04 $\pm$ 0.49 & 15.22 \\
    Snowball-3 & 82.95 $\pm$ 2.1 & 69.5 $\pm$ 5.01 & 83.11 $\pm$ 3.2 & 36.00 $\pm$ 1.36 & 65.49 $\pm$ 1.64 & 48.25 $\pm$ 0.94 & OOM   & 89.33 $\pm$ 1.3 & 80.93 $\pm$ 1.32 & 88.8 $\pm$ 0.82 & 14.78 \\
    \midrule
    ACM-SGC-1 & 93.77 $\pm$ 1.91 & 93.25 $\pm$ 2.92 & 93.61 $\pm$ 1.55 & 39.33 $\pm$ 1.25 & 63.68 $\pm$ 1.62 & 46.4 $\pm$ 1.13 & 66.67 $\pm$ 0.56 & 86.63 $\pm$ 1.13 & 80.96 $\pm$ 0.93 & 87.75 $\pm$ 0.88 & 12.60 \\
    ACM-SGC-2 & 93.77 $\pm$ 2.17 & 94.00 $\pm$ 2.61 & 93.44 $\pm$ 2.54 & 40.13 $\pm$ 1.21 & 60.48 $\pm$ 1.55 & 40.91 $\pm$ 1.39 & 66.53 $\pm$ 0.57 & 87.64 $\pm$ 0.99 & 80.93 $\pm$ 1.16 & 88.79 $\pm$ 0.5 & 13.40 \\
    ACM-GCNII & 92.62 $\pm$ 3.13 & 94.63 $\pm$ 2.96 & 92.46 $\pm$ 1.97 & 41.37 $\pm$ 1.37 & 58.73 $\pm$ 2.52 & 40.9 $\pm$ 1.58 & 66.39 $\pm$ 0.56 & 89.1 $\pm$ 1.61 & 82.28 $\pm$ 1.12 & 90.12 $\pm$ 0.4 & 10.40 \\
    ACM-GCNII* & 93.44 $\pm$ 2.74 & 94.37 $\pm$ 2.81 & 93.28 $\pm$ 2.79 & 41.27 $\pm$ 1.24 & 61.66 $\pm$ 2.29 & 38.32 $\pm$ 1.5 & 66.6 $\pm$ 0.57 & 89.00 $\pm$ 1.35 & 81.69 $\pm$ 1.25 & 90.18 0.51 & 10.10 \\
    ACM-GCN & 94.75 $\pm$ 3.8 & 95.75 $\pm$ 2.03 & 94.92 $\pm$ 2.88 & 41.62 $\pm$ 1.15 & 69.04 $\pm$ 1.74 & \cellcolor[rgb]{ .816,  .808,  .808}\textbf{58.02 $\pm$ 1.86} & 67.01 $\pm$ 0.38 & 88.62 $\pm$ 1.22 & 81.68 $\pm$ 0.97 & 90.66 $\pm$ 0.47 & 4.80 \\
    ACM-Snowball-2 & 95.08 $\pm$ 3.11 & 96.38 $\pm$ 2.59 & \cellcolor[rgb]{ .816,  .808,  .808}\textbf{95.74 $\pm$ 2.22} & 41.4 $\pm$ 1.23 & \cellcolor[rgb]{ .816,  .808,  .808}\textbf{68.51 $\pm$ 1.7} & 55.97 $\pm$ 2.03 & OOM   & 88.83 $\pm$ 1.49 & 81.58 $\pm$ 1.23 & 90.81 $\pm$ 0.52 & 4.44 \\
    ACM-Snowball-3 & 94.26 $\pm$ 2.57 & 96.62 $\pm$ 1.86 & 94.75 $\pm$ 2.41 & 41.27 $\pm$ 0.8 & 68.4 $\pm$ 2.05 & 55.73 $\pm$ 2.39 & OOM   & \cellcolor[rgb]{ .816,  .808,  .808}\textbf{89.59 $\pm$ 1.58} & 81.32 $\pm$ 0.97 & \cellcolor[rgb]{ .816,  .808,  .808}\textbf{91.44 $\pm$ 0.59} & 4.44 \\
    \midrule
    ACMII-GCN & \cellcolor[rgb]{ .816,  .808,  .808}\textbf{95.9 $\pm$ 1.83} & 96.62 $\pm$ 2.44 & 95.08 $\pm$ 2.07 & \cellcolor[rgb]{ .816,  .808,  .808}\textbf{41.84 $\pm$ 1.15} & 68.38 $\pm$ 1.36 & 54.53 $\pm$ 2.09 & 67.15 $\pm$ 0.41 & 89.00 $\pm$ 0.72 & 81.79 $\pm$ 0.95 & 90.74 $\pm$ 0.5 & \cellcolor[rgb]{ .816,  .808,  .808}\textbf{3.40} \\
    ACMII-Snowball-2 & 95.25 $\pm$ 1.55 & 96.63 $\pm$ 2.24 & 95.25 $\pm$ 1.55 & 41.1 $\pm$ 0.75 & 67.83 $\pm$ 2.63 & 53.48 $\pm$ 0.6 & OOM   & 88.95 $\pm$ 1.04 & 82.07 $\pm$ 1.04 & 90.56 $\pm$ 0.39 & 4.78 \\
    ACMII-Snowball-3 & 93.61 $\pm$ 2.79 & \cellcolor[rgb]{ .816,  .808,  .808}\textbf{97.00 $\pm$ 2.63} & 94.75 $\pm$ 3.09 & 40.31 $\pm$ 1.6 & 67.53 $\pm$ 2.83 & 52.31 $\pm$ 1.57 & OOM   & 89.36 $\pm$ 1.26 & 81.56 $\pm$ 1.15 & 91.31 $\pm$ 0.6 & 5.89 \\
    \bottomrule
    \bottomrule
    \end{tabular}%
  \label{tab:sota}%
\end{table}%

\paragraph{Datasets \& Experimental Setup}
In this section, we implement SGC \cite{wu2019simplifying} with 1 hop and 2 hops (SGC-1, SGC-2), GCNII \cite{chen2020simple}, GCNII* \cite{chen2020simple}, GCN \cite{kipf2016classification} and snowball networks with 2 and 3 layers (snowball-2, snowball-3) and apply them in ACM or ACMII framework (see details in appendix \ref{appendix:details_implementation_acm_acmII}): we use $\hat{A}_\text{rw}$ as LP filter and the corresponding HP filter can be derived from \eqref{eq:spatial_form_filterbank}. We compare them with several baselines and SOTA models: MLP with 2 layers (MLP-2), GAT \cite{velivckovic2017attention},  APPNP \cite{klicpera2018predict}, GPRGNN \cite{chien2021adaptive}, H$_2$GCN \cite{zhu2020beyond}, MixHop \cite{abu2019mixhop}, GCN+JK \cite{kipf2016classification, pmlr-v80-xu18c, lim2021new}, GAT+JK \cite{velivckovic2017attention, pmlr-v80-xu18c, lim2021new}, FAGCN \cite{bo2021beyond} GraphSAGE \cite{hamilton2017inductive} and Geom-GCN \cite{pei2020geom}. Besides the 9 benchmark datasets used in \cite{pei2020geom}, we further test the above models on a new benchmark dataset, \textit{Deezer-Europe}, that is proposed in \cite{lim2021new}.
We test these models 10 times on \textit{Cornell}, \textit{Wisconsin}, \textit{Texas}, \textit{Film}, \textit{Chameleon}, \textit{Squirrel}, \textit{Cora}, \textit{Citeseer} and \textit{Pubmed} following the same early stopping strategy, the same random data splitting method \footnote{See the open source code for splits in \url{https://github.com/jianhao2016/GPRGNN/blob/f4aaad6ca28c83d3121338a4c4fe5d162edfa9a2/src/utils.py\#L16}. See table \ref{tab:performance_comparison_fixed_splits} in appendix \ref{appendix:results_on_fixed_splits_as_geomgcn} for the performance comparison with several SOTA models on the fixed 48\%/32\%/20\% splits provided by \cite{pei2020geom}.} and Adam \cite{kingma2014adam} optimizer as used in GPRGNN \cite{chien2021adaptive}.  
For \textit{Deezer-Europe}, we test the above models 5 times with the same early stopping strategy, the same fixed splits and AdamW \cite{loshchilov2017decoupled} used in \cite{lim2021new}. 
The details of hyperparameter searching range and the optimal hyperparameters are reported in appendix \ref{appendix:hyperparameter_searching_range_real_world_datasets} and \ref{appendix:optimal_hyperparameter}. 

The main results of this set of experiments with statistics of datasets are summarized in Table \ref{tab:sota}, where we report the mean accuracy and standard deviation. We can see that after applied in ACM(II) framework, the performance of baseline models are boosted on almost all tasks. Especially, ACMII-GCN performs the best in terms of average rank (3.40) across all datasets and ACM(II)-GNNs achieve SOTA performance on $8$ out of $10$ datasets. Overall, It suggests that ACM(II) framework can help GNNs to generalize better on node classification tasks on heterophilous graphs.



\section{Future Work}
\label{sec:futurework}
The similarity matrix and the new metrics defined in this paper mainly capture the linear relations of the aggregated node features. But this might be insufficient sometimes when nonlinearity information in feature vectors are important for classification. In the future, similarity matrix that is able to capture nonlinear relations between node features can be proposed to define new homophily metrics.

As the limitation raised in section \ref{sec:acm_gnn_architecture}, filterbank method cannot properly handle all cases of harmful heterophily. In the future, we need to explore and address more challenging heterophilous graph with GNNs.

\clearpage
\bibliography{references}
\bibliographystyle{abbrv}

\clearpage
\section*{Checklist}

\begin{enumerate}

\item For all authors...
\begin{enumerate}
  \item Do the main claims made in the abstract and introduction accurately reflect the paper's contributions and scope?
    \answerYes{}
  \item Did you describe the limitations of your work?
    \answerYes{In the Appendix \ref{appendix:limitation_diversification}, we discussion cases that high-pass filter cannot tackle.}
  \item Did you discuss any potential negative societal impacts of your work?
    \answerNo{}{It is in Section 8, we have not come up with significant social negative impact.}
  \item Have you read the ethics review guidelines and ensured that your paper conforms to them?
    \answerYes{}
\end{enumerate}

\item If you are including theoretical results...
\begin{enumerate}
  \item Did you state the full set of assumptions of all theoretical results?
    \answerYes{In Section 3\&4, we mainly define a new homophily metric and it is followed by two theorems.}
	\item Did you include complete proofs of all theoretical results?
    \answerYes{In Appendix \ref{appendix:details_of_nll_loss_explanation}\&\ref{appendix:proof_theorem1}\& \ref{appendix:proof_theorem2}, we justify the new metric and two theorems.}
\end{enumerate}

\item If you ran experiments...
\begin{enumerate}
  \item Did you include the code, data, and instructions needed to reproduce the main experimental results (either in the supplemental material or as a URL)?
    \answerYes{The settings are provided in details and the source code is submitted in the supplemental material.}
  \item Did you specify all the training details (e.g., data splits, hyperparameters, how they were chosen)?
    \answerYes{In Section 6, we specify model details.}
	\item Did you report error bars (e.g., with respect to the random seed after running experiments multiple times)?
    \answerYes{We include average test accuracy of times of running with standard deviation.}
	\item Did you include the total amount of compute and the type of resources used (e.g., type of GPUs, internal cluster, or cloud provider)?
    \answerYes{We include hardware details in Appendix, which is not computationally expensive.}
\end{enumerate}

\item If you are using existing assets (e.g., code, data, models) or curating/releasing new assets...
\begin{enumerate}
  \item If your work uses existing assets, did you cite the creators?
    \answerYes{In Section 6, we specify the datasets with their data split sources in footnotes.}
  \item Did you mention the license of the assets?
    \answerNo{}
  \item Did you include any new assets either in the supplemental material or as a URL?
    \answerNo{}
  \item Did you discuss whether and how consent was obtained from people whose data you're using/curating?
    \answerNo{}
  \item Did you discuss whether the data you are using/curating contains personally identifiable information or offensive content?
    \answerNo{None included.}
\end{enumerate}

\item If you used crowdsourcing or conducted research with human subjects...
\begin{enumerate}
  \item Did you include the full text of instructions given to participants and screenshots, if applicable?
    \answerNo{None included.}
  \item Did you describe any potential participant risks, with links to Institutional Review Board (IRB) approvals, if applicable?
    \answerNo{None included.}
  \item Did you include the estimated hourly wage paid to participants and the total amount spent on participant compensation?
    \answerNo{None included.}
\end{enumerate}

\end{enumerate}


\clearpage
\appendix
\section{Hyperparameters \& Details of The Experiments}
\label{appendix:hyperparameter}

\subsection{Hyperparameters Searching Range for GNNs on Synthetic Graphs}
\label{appendix:hyperparameter_space_synthetic_graphs}

\begin{table}[htbp]
  \centering
  \caption{Hyperparameter Searching Range for Synthetic Experiments}
    \begin{tabular}{c|cccc}
    \toprule
    \toprule
    \multicolumn{5}{c}{Hyperparameter Searching Range for Synthetic Experiments} \\
    \midrule
    Models\textbackslash{}Hyperparameters & lr    & weight\_decay & dropout & hidden \\
    \midrule
    MLP-1 & 0.05  & \{5e-5, 1e-4, 5e-4\} & -     & - \\
    SGC-1 & 0.05  & \{5e-5, 1e-4, 5e-4\} & -     & - \\
    ACM-SGC-1 & 0.05  & \{5e-5, 1e-4, 5e-4\} &  \{ 0.1, 0.2, 0.3, 0.4, 0.5, 0.6, 0.7,0.8,0.9\} 
    & - \\
    \midrule
    MLP-2 & 0.05  & \{5e-5, 1e-4, 5e-4\} & \{ 0.1, 0.2, 0.3, 0.4, 0.5, 0.6, 0.7,0.8,0.9\} & 64 \\
    GCN   & 0.05  & \{5e-5, 1e-4, 5e-4\} & \{ 0.1, 0.2, 0.3, 0.4, 0.5, 0.6, 0.7,0.8,0.9\} & 64 \\
    ACM-GCN & 0.05  & \{5e-5, 1e-4, 5e-4\} & \{ 0.1, 0.2, 0.3, 0.4, 0.5, 0.6, 0.7,0.8,0.9\} & 64 \\
    \bottomrule
    \bottomrule
    \end{tabular}%
  \label{tab:synthetic_data_hyperparameter_searching_range}%
\end{table}%

\subsection{Hyperparameters Searching Range for GNNs on Ablation Study}

\begin{table}[htbp]
  \centering
  \scalebox{0.75}{
  \caption{Hyperparameter Searching Range for Ablation Study}
  \setlength{\tabcolsep}{2pt}
    \begin{tabular}{c|cccc}
    \toprule
    \toprule
    \multicolumn{5}{c}{Hyperparameter Searching Range for Ablation Study} \\
    \midrule
    Models\textbackslash{}Hyperparameters & lr    & weight\_decay & dropout & hidden \\
    \midrule
    SGC-LP+HP & \{0.01, 0.05, 0.1\} & \{0, 5e-6, 1e-5, 5e-5, 1e-4, 5e-4, 1e-3, 5e-3, 1e-2\} & -
    & - \\
    SGC-LP+Identity & \{0.01, 0.05, 0.1\} & \{0, 5e-6, 1e-5, 5e-5, 1e-4, 5e-4, 1e-3, 5e-3, 1e-2\} & -
    & - \\
    ACM-SGC-no adaptive mixing & \{0.01, 0.05, 0.1\} & \{0, 5e-6, 1e-5, 5e-5, 1e-4, 5e-4, 1e-3, 5e-3, 1e-2\} & \{0, 0.1, 0.2, 0.3, 0.4, 0.5, 0.6, 0.7,0.8,0.9\} 
    & - \\
    GCN-LP+HP & \{0.01, 0.05, 0.1\} & \{0, 5e-6, 1e-5, 5e-5, 1e-4, 5e-4, 1e-3, 5e-3, 1e-2\} & \{0, 0.1, 0.2, 0.3, 0.4, 0.5, 0.6, 0.7,0.8,0.9\} & 64 \\
    GCN-LP+Identity & \{0.01, 0.05, 0.1\} & \{0, 5e-6, 1e-5, 5e-5, 1e-4, 5e-4, 1e-3, 5e-3, 1e-2\} & \{0, 0.1, 0.2, 0.3, 0.4, 0.5, 0.6, 0.7,0.8,0.9\} & 64 \\
    ACM-GCN-no adaptive mixing & \{0.01, 0.05, 0.1\} & \{0, 5e-6, 1e-5, 5e-5, 1e-4, 5e-4, 1e-3, 5e-3, 1e-2\} & \{0, 0.1, 0.2, 0.3, 0.4, 0.5, 0.6, 0.7,0.8,0.9\} & 64 \\
    \bottomrule
    \bottomrule
    \end{tabular}%
    }
  \label{tab:ablation_study_hyperparameters}%
\end{table}%

\subsection{Hyperparameters Searching Range for GNNs on Real-world Datasets}
\label{appendix:hyperparameter_searching_range_real_world_datasets}
See table \ref{tab:real_world_datasets_hyperparameter_searching_range}.

\begin{table}[htbp]
  \centering
  \tiny
  \caption{Hyperparameter Searching Range for Real-world Datasets}
    \begin{tabular}{p{10.875em}|cp{10.125em}p{3.625em}cp{5.065em}p{4.875em}p{3.065em}p{5.5em}p{4.25em}}
    \toprule
    \toprule
    Models\textbackslash{}Hyperparameters & {lr} & weight\_decay & dropout & {hidden} & lambda & alpha\_l & head  & layers & JK type \\
    \midrule
    H2GCN & 0.01  & {0.001} & \{0, 0.5\} & \multicolumn{1}{p{7.125em}}{\{8, 16, 32, 64\}} & -     & -  & -  & \{1, 2\} & - \\
    \midrule
    MixHop & 0.01  & {0.001} & {0.5} & \multicolumn{1}{p{7.125em}}{\{8, 16, 32\}} & -   & -   & -   & \{2, 3\} & - \\
    \midrule
    GCN+JK & \multicolumn{1}{p{4.315em}}{\{0.1, 0.01, 0.001\}} & {0.001} & {0.5} & \multicolumn{1}{p{7.125em}}{\{4, 8, 16, 32, 64\}} & -     & -     & -     & {2} & \multicolumn{1}{p{7.125em}}{\{max, cat\}} \\
    \midrule
     GAT+JK & \multicolumn{1}{p{4.315em}}{\{0.1, 0.01, 0.001\}} & {0.001} & {0.5} & \multicolumn{1}{p{7.125em}}{\{4, 8, 12, 32\}} & -     & -     & \{2,4,8\} & {2} & \multicolumn{1}{p{7.125em}}{\{max, cat\}} \\
    \midrule
    GCNII, GCNII* & 0.01  & \{0, 5e-6, 1e-5, 5e-5, 1e-4, 5e-4, 1e-3\} for Deezer-Europe and \{0, 5e-6, 1e-5, 5e-5, 1e-4, 5e-4, 1e-3, 5e-3, 1e-2\} for others & {0.5} & 64    & \{0.5, 1, 1.5\} & \{0.1,0.2,0.3,0,4,0.5\} & -     & \{4, 8, 16, 32\} for Deezer-Europe and \{4, 8, 16, 32, 64\} for others & - \\
    \midrule
    Baselines: \{SGC-1, SGC-2, GCN, Snowball-2, Snowball-3, FAGCN\}; ACM-\{SGC-1, SGC-2, GCN, Snowball-2, Snowball-3\}; ACMII-\{SGC-1, SGC-2, GCN, Snowball-2, Snowball-3\} & \multicolumn{1}{p{4.315em}}{\{0.002, 0.01, 0.05\} for Deezer-Europe and \{0.01, 0.05, 0.1\} for others} & \{0, 5e-6, 1e-5, 5e-5, 1e-4, 5e-4, 1e-3\} for Deezer-Europe and \{0, 5e-6, 1e-5, 5e-5, 1e-4, 5e-4, 1e-3, 5e-3, 1e-2\} for others & \{0, 0.1, 0.2, 0.3, 0.4, 0.5, 0.6, 0.7, 0.8, 0.9\} & 64    & -     & -     & -     & -     & - \\
    \midrule
    GraphSAGE & \multicolumn{1}{p{4.315em}}{\{0.01,0.05, 0.1\}} & \{0, 5e-6, 1e-5, 5e-5, 1e-4, 5e-4, 1e-3\} for Deezer-Europe and \{0, 5e-6, 1e-5, 5e-5, 1e-4, 5e-4, 1e-3, 5e-3, 1e-2\} for others & \{ 0, 0.1, 0.2, 0.3, 0.4, 0.5, 0.6, 0.7, 0.8, 0.9\} & \multicolumn{1}{p{7.125em}}{8 for Deezer-Europe and 64 for others} & -     & -     & -     & -     & - \\
    \midrule
    ACM-\{GCNII, GCNII*\} & 0.01  & \{0, 5e-6, 1e-5, 5e-5, 1e-4, 5e-4, 1e-3\} for Deezer-Europe and \{0, 5e-6, 1e-5, 5e-5, 1e-4, 5e-4, 1e-3, 5e-3, 1e-2\} for others & \{ 0, 0.1, 0.2, 0.3, 0.4, 0.5, 0.6, 0.7, 0.8, 0.9\} & 64    & -     & -     & -     & \{1,2,3,4\} & - \\
    \bottomrule
    \bottomrule
    \end{tabular}%
  \label{tab:real_world_datasets_hyperparameter_searching_range}%
\end{table}%

\subsection{Optimal Hyperparameters for Baselines and ACM(II)-GNNs on Real-world Tasks}
\label{appendix:optimal_hyperparameter}
See table \ref{tab:hyperparameters_baselines} and \ref{tab:hyperparameters_acmgnns}.
\begin{table}[htbp]
  \centering
   \scalebox{.49}{
  \caption{Hyperparameters for baseline models}
    \begin{tabular}{c|c|cccccccccccc}
    \toprule
    \toprule
    \multicolumn{14}{c}{\textbf{Hyperparameters for Baseline GNNs}} \\
    \midrule
    Datasets & Models\textbackslash{}Hyperparameters & lr    & weight\_decay & dropout & hidden & {\# layers} & {Gat heads} & {JK Type} & lambda & alpha\_l & results & std   & average epoch time/average total time \\
    \midrule
    \multirow{12}[1]{*}{\textbf{Cornell}} & SGC-1 & 0.05  & 1.00E-02 & 0     & 64    & -     & -     & -     & -     & -     & 70.98 & 8.39  & 2.53ms/0.51s \\
          & SGC-2 & 0.05  & 1.00E-03 & 0     & 64    & -     & -     & -     & -     & -     & 72.62 & 9.92  & 2.46ms/0.53s \\
          & GCN   & 0.1   & 5.00E-03 & 0.5   & 64    & 2     & -     & -     & -     & -     & 82.46 & 3.11  & 3.67ms/0.74s \\
          & Snowball-2 & 0.01  & 5.00E-03 & 0.4   & 64    & 2     & -     & -     & -     & -     & 82.62 & 2.34  & 4.24ms/0.87s \\
          & Snowball-3 & 0.01  & 5.00E-03 & 0.4   & 64    & 3     & -     & -     & -     & -     & 82.95 & 2.1   & 6.66ms/1.36s \\
          & GCNII & 0.01  & 1.00E-03 & 0.5   & 64    & 16    & -     & -     & 0.5   & 0.5   & 89.18 & 3.96  & 25.41ms/8.11s \\
          & GCNII* & 0.01  & 1.00E-03 & 0.5   & 64    & 8     & -     & -     & 0.5   & 0.5   & 90.49 & 4.45  & 15.35ms/4.05s \\
          & FAGCN & 0.01  & 1.00E-04 & 0.7   & 32    & 2     & -     & -     & -     & -     & 88.03 & 5.6   & 8.1ms/3.8858s \\
          &  Mixhop & 0.01  & 0.001 & 0.5   & 16    & 2     & - & - & -     & -     & 60.33 & 28.53 & {10.379ms/2.105s} \\
          & H2GCN & 0.01  & 0.001 & 0.5   & 64    & 1     & - & - & -     & -     & 86.23 & 4.71  & {4.381ms/1.123s} \\
          & GCN+JK & 0.1   & 0.001 & 0.5   & 64    & 2     & - & {cat} & -     & -     & 66.56 & 13.82 & {5.589ms/1.227s} \\
          &  GAT+JK & 0.1   & 0.001 & 0.5   & 32    & 2     & 8     & {max} & -     & -     & 74.43 & 10.24 & {10.725ms/2.478s} \\
    \midrule
    \multirow{14}[1]{*}{\textbf{Wisconsin}} & SGC-1 & 0.05  & 5.00E-03 & 0     & 64    & -     & -     & -     & -     & -     & 70.38 & 2.85  & 2.83ms/0.57s \\
          & SGC-2 & 0.1   & 1.00E-03 & 0     & 64    & -     & -     & -     & -     & -     & 74.75 & 2.89  & { 2.14ms/0.43s} \\
          & GCN   & 0.1   & 1.00E-03 & 0.7   & 64    & 2     & -     & -     & -     & -     & 75.5  & 2.92  & {3.74ms/0.76s} \\
          & Snowball-2 & 0.1   & 1.00E-03 & 0.5   & 64    & 2     & -     & -     & -     & -     & 74.88 & 3.42  & 3.73ms/0.76s \\
          & Snowball-3 & 0.05  & 5.00E-04 & 0.8   & 64    & 3     & -     & -     & -     & -     & 69.5  & 5.01  & {5.46ms/1.12s} \\
          & GCNII & 0.01  & 1.00E-03 & 0.5   & 64    & 8     & -     & -     & 0.5   & 0.5   & 83.25 & 2.69  &  \\
          & GCNII* & 0.01  & 1.00E-03 & 0.5   & 64    & 4     & -     & -     & 1.5   & 0.3   & 89.12 & 3.06  & 9.26ms/1.96s \\
          & FAGCN & 0.05  & 1.00E-04 & 0     & 32    & 2     & -     & -     & -     & -     & 89.75 & 6.37  & 12.9ms/4.6359s \\
          &  Mixhop & 0.01  & 0.001 & 0.5   & 16    & 2     & - & - & -     & -     & 77.25 & 7.80  & {10.281ms/2.095s} \\
          & H2GCN & 0.01  & 0.001 & 0.5   & 32    & 1     & - & - & -     & -     & 87.5  & 1.77  & {4.324ms/1.134s} \\
          & GCN+JK & 0.1   & 0.001 & 0.5   & 32    & 2     & - & {cat} & -     & -     & 62.5  & 15.75 & {5.117ms/1.049s} \\
          &  GAT+JK & 0.1   & 0.001 & 0.5   & 4     & 2     & 8     & {max} & -     & -     & 69.5  & 3.12  & {10.762ms/2.25s} \\
          & APPNP & 0.05  & 0.001 & 0.5   & 64    & 2     & - & - & -     & -     & 92    & 3.59  & {10.303ms/2.104s} \\
          & GPRGNN & 0.05  & 0.001 & 0.5   & 256   & 2     & - & - & -     & -     & 93.75 & 2.37  & {11.856ms/2.415s} \\
          \midrule
    \multirow{12}[0]{*}{\textbf{Texas}} & SGC-1 & 0.05  & 1.00E-03 & 0     & 64    & -     & -     & -     & -     & -     & 83.28 & 5.43  &  2.55ms/0.54s \\
          & SGC-2 & 0.01  & 1.00E-03 & 0     & 64    & -     & -     & -     & -     & -     & 81.31 & 3.3   & 2.61ms/2.53s \\
          & GCN   & 0.05  & 1.00E-02 & 0.9   & 64    & 2     & -     & -     & -     & -     & 83.11 & 3.2   & 3.59ms/0.73s \\
          & Snowball-2 & 0.05  & 1.00E-02 & 0.9   & 64    & 2     & -     & -     & -     & -     & 83.11 & 3.2   & 3.98ms/0.82s \\
          & Snowball-3 & 0.05  & 1.00E-02 & 0.9   & 64    & 3     & -     & -     & -     & -     & 83.11 & 3.2   & 5.56ms/1.12s \\
          & GCNII & 0.01  & 1.00E-04 & 0.5   & 64    & 4     & -     & -     & 1.5   & 0.5   & 82.46 & 4.58  &  \\
          & GCNII* & 0.01  & 1.00E-04 & 0.5   & 64    & 8     & -     & -     & 0.5   & 0.5   & 88.52 & 3.02  & 15.64ms/3.47s \\
          & FAGCN & 0.01  & 5.00E-04 & 0     & 32    & 2     & -     & -     & -     & -     & 88.85 & 4.39  & 8.8ms/6.5252s \\
          &  Mixhop & 0.01  & 0.001 & 0.5   & 32    & 2     & - & - & - & - & 76.39 & 7.66  & {11.099ms/2.329s} \\
          & H2GCN & 0.01  & 0.001 & 0.5   & 64    & 1     & - & - & - & - & 85.90 & 3.53  & {4.197ms/0.95s} \\
          & GCN+JK & 0.1   & 0.001 & 0.5   & 32    & 2     & - & {cat} & - & - & 80.66 & 1.91  & {5.28ms/1.085s} \\
          &  GAT+JK & 0.1   & 0.001 & 0.5   & 8     & 2     & 2     & {cat} & - & - & 75.41 & 7.18  & {10.937ms/2.402s} \\
          \midrule
    \multirow{12}[0]{*}{\textbf{Film}} & SGC-1 & 0.01  & 5.00E-06 & 0     & 64    & -     & -     & -     & -     & -     & 25.26 & 1.18  & 3.18ms/0.70s \\
          & SGC-2 & 0.01  & 5.00E-06 & 0     & 64    & -     & -     & -     & -     & -     & 28.81 & 1.11  &  2.13ms/0.43s \\
          & GCN   & 0.1   & 5.00E-04 & 0     & 64    & 2     & -     & -     & -     & -     & 35.51 & 0.99  & 4.86ms/0.99s \\
          & Snowball-2 & 0.1   & 5.00E-04 & 0     & 64    & 2     & -     & -     & -     & -     & 35.97 & 0.66  & 5.59ms/1.14s \\
          & Snowball-3 & 0.1   & 5.00E-04 & 0.2   & 64    & 3     & -     & -     & -     & -     & 36    & 1.36  & 7.89ms/1.60s \\
          & GCNII & 0.01  & 1.00E-04 & 0.5   & 64    & 8     & -     & -     & 1.5   & 0.3   & 40.82 & 1.79  & 15.85ms/3.22s \\
          & GCNII* & 0.01  & 1.00E-06 & 0.5   & 64    & 4     & -     & -     & 1     & 0.1   & 41.54 & 0.99  &  \\
          & FAGCN & 0.01  & 5.00E-05 & 0.6   & 32    & 2     & -     & -     & -     & -     & 31.59 & 1.37  & 45.4ms/11.107s \\
          &  Mixhop & 0.01  & 0.001 & 0.5   & 8     & 3     & 8     & {max} & - & - & 33.13 & 2.40  & {17.651ms/3.566s} \\
          & H2GCN & 0.01  & 0.001 & 0     & 64    & 1     & 8     & {max} & - & - & 38.85 & 1.17  & {8.101ms/1.695s} \\
          & GCN+JK & 0.1   & 0.001 & 0.5   & 64    & 2     & 8     & {cat} & - & - & 32.72 & 2.62  & {8.946ms/1.807s} \\
          &  GAT+JK & 0.001 & 0.001 & 0.5   & 32    & 2     & 4     & {cat} & - & - & 35.41 & 0.97  & {20.726ms/4.187s} \\
          \midrule
    \multirow{12}[0]{*}{\textbf{Chameleon}} & SGC-1 & 0.1   & 5.00E-06 & 0     & 64    & -     & -     & -     & -     & -     & 64.86 & 1.81  & 3.48ms/2.96s \\
          & SGC-2 & 0.1   & 0.00E+00 & 0     & 64    & -     & -     & -     & -     & -     & 62.67 & 2.41  &  4.43ms/1.12s \\
          & GCN   & 0.01  & 1.00E-05 & 0.9   & 64    & 2     & -     & -     & -     & -     & 64.18 & 2.62  & 4.96ms/1.18s \\
          & Snowball-2 & 1.00E-01 & 1.00E-05 & 0.9   & 64    & 2     & -     & -     & -     & -     & 64.99 & 2.39  & 4.96ms/1.00s \\
          & Snowball-3 & 0.1   & 5.00E-06 & 0.9   & 64    & 3     & -     & -     & -     & -     & 65.49 & 1.64  & 7.44ms/1.50s \\
          & GCNII & 0.01  & 5.00E-06 & 0.5   & 64    & 4     & -     & -     & 0.5   & 0.1   & 60.35 & 2.7   & 9.76ms/2.26s \\
          & GCNII* & 0.01  & 5.00E-04 & 0.5   & 64    & 4     & -     & -     & 1.5   & 0.5   & 62.8  & 2.87  & 10.40ms/2.17s \\
          & FAGCN & 0.002 & 1.00E-04 & 0     & 32    & 2     & -     & -     & -     & -     & 49.47 & 2.84  & 8.4ms/13.8696s \\
          &  Mixhop & 0.01  & 0.001 & 0.5   & 16    & 2     & 8     & {max} & - & - & 36.28 & 10.2  & {11.372ms/2.297s} \\
          & H2GCN & 0.01  & 0.001 & 0     & 32    & 1     & 8     & {max} & - & - & 52.3  & 0.48  & {4.059ms/0.82s} \\
          & GCN+JK & 0.001 & 0.001 & 0.5   & 32    & 2     & 8     & {cat} & - & - & 64.68 & 2.85  & {5.211ms/1.053s} \\
          &  GAT+JK & 0.001 & 0.001 & 0.5   & 4     & 2     & 8     & {max} & - & - & 68.14 & 1.18  & {13.772ms/2.788s} \\
          \midrule
    \multirow{12}[0]{*}{\textbf{Squirrel}} & SGC-1 & 0.05  & 0.00E+00 & 0     & 64    & -     & -     & -     & -     & -     & 47.62 & 1.27  & 4.65ms/1.44s \\
          & SGC-2 & 0.1   & 0.00E+00 & 0.9   & 64    & -     & -     & -     & -     & -     & 41.25 & 1.4   & 35.06ms/7.81s \\
          & GCN   & 0.01  & 5.00E-05 & 0.7   & 64    & 2     & -     & -     & -     & -     & 44.76 & 1.39  & 8.41ms/2.50s \\
          & Snowball-2 & 0.1   & 0.00E+00 & 0.9   & 64    & 2     & -     & -     & -     & -     & 47.88 & 1.23  & 8.96ms/1.92s \\
          & Snowball-3 & 0.1   & 0.00E+00 & 0.8   & 64    & 3     & -     & -     & -     & -     & 48.25 & 0.94  & 14.00ms/2.90s \\
          & GCNII & 0.01  & 1.00E-04 & 0.5   & 64    & 4     & -     & -     & 1.5   & 0.2   & 38.81 & 1.97  & 13.35ms/2.70s \\
          & GCNII* & 0.01  & 5.00E-04 & 0.5   & 64    & 4     & -     & -     & 1.5   & 0.3   & 38.31 & 1.3   & 13.81ms/2.78s \\
          & FAGCN & 0.05  & 1.00E-04 & 0     & 32    & 2     & -     & -     & -     & -     & 42.24 & 1.2   & 16ms/6.7961s \\
          &  Mixhop & 0.01  & 0.001 & 0.5   & 32    & 2     & - & - & - & - & 24.55 & 2.6   & {17.634ms/3.562s} \\
          & H2GCN & 0.01  & 0.001 & 0     & 16    & 1     & - & - & - & - & 30.39 & 1.22  & {9.315ms/1.882s} \\
          & GCN+JK & 0.001 & 0.001 & 0.5   & 32    & 2     & - & {max} & - & - & 53.4  & 1.9   & {14.321ms/2.905s} \\
          &  GAT+JK & 0.001 & 0.001 & 0.5   & 8     & 2     & 4     & {max} & - & - & 52.28 & 3.61  & {29.097ms/5.878s} \\
          \midrule
    \multirow{12}[1]{*}{\textbf{Cora}} & SGC-1 & 0.1   & 5.00E-06 & 0     & 64    & -     & -     & -     & -     & -     & 85.12 & 1.64  & 3.47ms/11.55s \\
          & SGC-2 & 0.1   & 1.00E-05 & 0     & 64    & -     & -     & -     & -     & -     & 85.48 & 1.48  & 2.91ms/6.85s \\
          & GCN   & 0.1   & 5.00E-04 & 0.2   & 64    & 2     & -     & -     & -     & -     & 87.78 & 0.96  & 4.24ms/0.86s \\
          & Snowball-2 & 0.1   & 5.00E-04 & 0.1   & 64    & 2     & -     & -     & -     & -     & 88.64 & 1.15  & 4.65ms/0.94s \\
          & Snowball-3 & 0.05  & 1.00E-03 & 0.6   & 64    & 3     & -     & -     & -     & -     & 89.33 & 1.3   & 6.41ms/1.32s \\
          & GCNII & 0.01  & 1.00E-04 & 0.5   & 64    & 16    & -     & -     & 0.5   & 0.2   & 88.98 & 1.33  &  \\
          & GCNII* & 0.01  & 5.00E-04 & 0.5   & 64    & 4     & -     & -     & 0.5   & 0.5   & 88.93 & 1.37  & 10.16ms/2.24s \\
          & FAGCN & 0.05  & 5.00E-04 & 0     & 32    & 2     & -     & -     & -     & -     & 88.85 & 1.36  & 8.4ms/3.3183s \\
          &  Mixhop & 0.01  & 0.001 & 0.5   & 16    & 2     & - & - & - & - & 65.65 & 11.31 & {11.177ms/2.278s} \\
          & H2GCN & 0.01  & 0.001 & 0     & 32    & 1     & - & - & - & - & 87.52 & 0.61  & {4.335ms/1.209s} \\
          & GCN+JK & 0.001 & 0.001 & 0.5   & 64    & 2     & - & {cat} & - & - & 86.90 & 1.51  & {6.656ms/1.346s} \\
          &  GAT+JK & 0.001 & 0.001 & 0.5   & 32    & 2     & 2     & {cat} & - & - & 89.52 & 0.43  & {12.91ms/2.608s} \\
    \midrule
    \multirow{12}[1]{*}{\textbf{CiteSeer}} & SGC-1 & 0.1   & 5.00E-04 & 0     & 64    & -     & -     & -     & -     & -     & 79.66 & 0.75  & 3.43ms/7.30s \\
          & SGC-2 & 0.01  & 5.00E-04 & 0.9   & 64    & -     & -     & -     & -     & -     & 80.75 & 1.15  & 5.33ms/4.40s \\
          & GCN   & 0.1   & 1.00E-03 & 0.9   & 64    & 2     & -     & -     & -     & -     & 81.39 & 1.23  & 4.18ms/0.86s \\
          & Snowball-2 & 0.1   & 1.00E-03 & 0.8   & 64    & 2     & -     & -     & -     & -     & 81.53 & 1.71  & 5.19ms/1.11s \\
          & Snowball-3 & 0.1   & 1.00E-03 & 0.9   & 64    & 3     & -     & -     & -     & -     & 80.93 & 1.32  & 7.64ms/1.69s \\
          & GCNII & 0.01  & 1.00E-03 & 0.5   & 64    & 16    & -     & -     & 0.5   & 0.2   & 81.58 & 1.3   &  \\
          & GCNII* & 0.01  & 1.00E-03 & 0.5   & 64    & 16    & -     & -     & 0.5   & 0.2   & 81.83 & 1.78  & 32.50ms/10.29s \\
          & FAGCN & 0.05  & 5.00E-04 & 0     & 32    & 2     & -     & -     & -     & -     & 82.37 & 1.46  & 9.4ms/4.7648s \\
          &  Mixhop & 0.01  & 0.001 & 0.5   & 16    & 2     & - & - & - & - & 49.52 & 13.35 & {13.793ms/2.786s} \\
          & H2GCN & 0.01  & 0.001 & 0     & 8     & 1     & - & - & - & - & 79.97 & 0.69  & {5.794ms/3.049s} \\
          & GCN+JK & 0.001 & 0.001 & 0.5   & 32    & 2     & - & {max} & - & - & 73.77 & 1.85  & {5.264ms/1.063s} \\
          &  GAT+JK & 0.001 & 0.001 & 0.5   & 8     & 2     & 4     & {max} & - & - & 74.49 & 2.76  & {12.326ms/2.49s} \\
          \midrule
    \multirow{12}[0]{*}{\textbf{PubMed}} & SGC-1 & 0.05  & 5.00E-06 & 0.3   & 64    & -     & -     & -     & -     & -     & 87.75 & 0.88  & 6.04ms/2.61s \\
          & SGC-2 & 0.05  & 5.00E-05 & 0.1   & 64    & -     & -     & -     & -     & -     & 88.79 & 0.5   & 8.62ms/3.18s \\
          & GCN   & 0.1   & 5.00E-05 & 0.6   & 64    & 2     & -     & -     & -     & -     & 88.9  & 0.32  & 5.08ms/1.03s \\
          & Snowball-2 & 0.1   & 5.00E-04 & 0     & 64    & 2     & -     & -     & -     & -     & 89.04 & 0.49  & 5.68ms/1.19s \\
          & Snowball-3 & 0.1   & 5.00E-06 & 0     & 64    & 3     & -     & -     & -     & -     & 88.8  & 0.82  & 8.54ms/1.75s \\
          & GCNII & 0.01  & 1.00E-06 & 0.5   & 64    & 4     & -     & -     & 0.5   & 0.5   & 89.8  & 0.3   & 10.98ms/3.21s \\
          & GCNII* & 0.01  & 1.00E-06 & 0.5   & 64    & 4     & -     & -     & 0.5   & 0.1   & 89.98 & 0.52  & 11.47ms/3.24s \\
          & FAGCN & 0.05  & 5.00E-04 & 0     & 32    & 2     & -     & -     & -     & -     & 89.98 & 0.54  & 14.5ms/6.411s \\
          &  Mixhop & 0.01  & 0.001 & 0.5   & 16    & 2     & - & - & - & - & 87.04 & 4.10  & {17.459ms/3.527s} \\
          & H2GCN & 0.01  & 0.001 & 0     & 64    & 1     & - & - & - & - & 87.78 & 0.28  & {8.039ms/2.28s} \\
          & GCN+JK & 0.01  & 0.001 & 0.5   & 32    & 2     & - & {cat} & - & - & 90.09 & 0.68  & {12.001ms/2.424s} \\
          &  GAT+JK & 0.1   & 0.001 & 0.5   & 8     & 2     & 4     & {max} & - & - & 89.15 & 0.87  & {20.403ms/4.125s} \\
          \midrule
    \multirow{3}[1]{*}{\textbf{Deezer-Europe}} & FAGCN & 0.01  & 0.0001 & 0     & 32    & 2     & -     & -     & -     & -     & 66.86 & 0.53  & {41.7ms/20.8362s} \\
          & GCNII & 0.01  & 5e-6,1e-5 & 0.5   & 64    & 32    & -     & -     & 0.5   & 0.5   & 66.38 & 0.45  & 126.58ms/63.16s \\
          & GCNII* & 0.01  & 1e-4,1e-3 & 0.5   & 64    & 32    & -     & -     & 0.5   & 0.5   & 66.42 & 0.56  & 134.05ms/66.89s \\
    \bottomrule
    \bottomrule
    \end{tabular}%
     \label{tab:hyperparameters_baselines}%
    }
\end{table}%

\begin{table}[htbp]
  \centering
  \scalebox{.57}{
  \caption{Hyperparameters for ACM-GNNs and ACMII-GNNs}
    \begin{tabular}{c|c|cccccccccccc}
    \toprule
    \toprule
    \multicolumn{14}{c}{\textbf{Hyperparameters for ACM-GNNs and ACMII-GNNs}} \\
    \midrule
    Datasets & Models\textbackslash{}Hyperparameters & lr    & weight\_decay & dropout & hidden & {\# layers} & {Gat heads} & {JK Type} & lambda & alpha\_l & results & std   & average epoch time/average total time \\
    \midrule
    \multirow{10}[0]{*}{\textbf{Cornell}} & ACM-SGC-1 & 0.01  & 5.00E-03 & 0.6   & 64    & -     & -     & -     & -     & -     & 93.77 & 1.91  & 5.53ms/2.31s \\
          & ACM-SGC-2 & 0.01  & 5.00E-03 & 0.6   & 64    & -     & -     & -     & -     & -     & 93.77 & 2.17  &  4.73ms/1.87s \\
          & ACM-GCN & 0.05  & 1.00E-02 & 0.2   & 64    & 2     & -     & -     & -     & -     & 94.75 & 3.8   & 8.25ms/1.69s \\
          & ACMII-GCN & 0.1   & 1.00E-02 & 0.5   & 64    & 2     & -     & -     & -     & -     & 95.25 & 2.79  & 8.43ms/1.71s \\
          & ACM-GCNII & 0.01  & 1.00E-03 & 0.5   & 64    & 1     & -     & -     & 0.5   & 0.4   & 92.62 & 3.13  & 6.81ms/1.43s \\
          & ACM-GCNII* & 0.01  & 5.00E-04 & 0.5   & 64    & 1     & -     & -     & 0.5   & 0.1   & 93.44 & 2.74  & 6.76ms/1.39s \\
          & ACM-Snowball-2 & 0.05  & 1.00E-02 & 0.2   & 64    & 2     & -     & -     & -     & -     & 95.08 & 3.11  & 9.15ms/1.86s \\
          & ACM-Snowball-3 & 0.1   & 1.00E-02 & 0.4   & 64    & 3     & -     & -     & -     & -     & 94.26 & 2.57  & 13.20ms/2.68s \\
          & ACMII-Snowball-2 & 0.05  & 1.00E-02 & 0.6   & 64    & 2     & -     & -     & -     & -     & 95.25 & 1.55  & 8.23ms/1.72s \\
          & ACMII-Snowball-3 & 0.05  & 1.00E-02 & 0.7   & 64    & 3     & -     & -     & -     & -     & 93.61 & 2.79  & 11.70ms/2.37s \\
          \midrule
    \multirow{10}[0]{*}{\textbf{Wisconsin}} & ACM-SGC-1 & 0.05  & 5.00E-03 & 0.7   & 64    & -     & -     & -     & -     & -     & {93.25} & 2.92  & 5.96ms/1.34s \\
          & ACM-SGC-2 & 0.1   & 5.00E-03 & 0.2   & 64    & -     & -     & -     & -     & -     & 94    & 2.61  & 4.60ms/0.95s \\
          & ACM-GCN & 0.1   & 5.00E-03 & 0     & 64    & 2     & -     & -     & -     & -     & 95.75 & 2.03  & {8.11ms/1.64s} \\
          & ACMII-GCN & 0.1   & 1.00E-02 & 0.2   & 64    & 2     & -     & -     & -     & -     & 96.62 & 2.44  & 8.28ms/1.68s \\
          & ACM-GCNII & 0.01  & 5.00E-03 & 0.5   & 64    & 1     & -     & -     & 1     & 0.1   & 94.63 & 2.96  & 9.31ms/2.19s \\
          & ACM-GCNII* & 0.01  & 1.00E-03 & 0.5   & 64    & 1     & -     & -     & 1.5   & 0.4   & 94.37 & 2.81  & 7.11ms/1.45s \\
          & ACM-Snowball-2 & 0.1   & 5.00E-03 & 0.1   & 64    & 2     & -     & -     & -     & -     & 96.38 & 2.59  &  8.63ms/1.74s \\
          & ACM-Snowball-3 & 0.05  & 1.00E-02 & 0.3   & 64    & 3     & -     & -     & -     & -     & 96.62 & 1.86  & 12.79ms/2.58s \\
          & ACMII-Snowball-2 & 0.1   & 1.00E-02 & 0.1   & 64    & 2     & -     & -     & -     & -     & {96.63} & 2.24  & 8.11ms/1.65s \\
          & ACMII-Snowball-3 & 0.1   & 5.00E-03 & 0.1   & 64    & 3     & -     & -     & -     & -     & 97    & 2.63  & 12.38ms/2.51s \\
          \midrule
    \multirow{10}[0]{*}{\textbf{Texas}} & ACM-SGC-1 & 0.01  & 5.00E-03 & 0.6   & 64    & -     & -     & -     & -     & -     & 93.61 & 1.55  & 5.43ms/2.18s \\
          & ACM-SGC-2 & 0.05  & 5.00E-03 & 0.4   & 64    & -     & -     & -     & -     & -     & 93.44 & 2.54  & 4.59ms/1.01s \\
          & ACM-GCN & 0.05  & 1.00E-02 & 0.6   & 64    & 2     & -     & -     & -     & -     & 94.92 & 2.88  & 8.33ms/1.70s \\
          & ACMII-GCN & 0.1   & 5.00E-03 & 0.4   & 64    & 2     & -     & -     & -     & -     & 95.08 & 2.54  & 8.49ms/1.72s \\
          & ACM-GCNII & 0.01  & 1.00E-03 & 0.5   & 64    & 1     & -     & -     & 0.5   & 0.4   & 92.46 & 1.97  & 6.47ms/1.36s \\
          & ACM-GCNII* & 0.01  & 1.00E-03 & 0.5   & 64    & 1     & -     & -     & 0.5   & 0.4   & 93.28 & 2.79  & 7.03ms/1.45s \\
          & ACM-Snowball-2 & 0.05  & 1.00E-02 & 0.1   & 64    & 2     & -     & -     & -     & -     & 95.74 & 2.22  & 8.35ms/1.71s \\
          & ACM-Snowball-3 & 0.01  & 5.00E-03 & 0.6   & 64    & 3     & -     & -     & -     & -     & 94.75 & 2.41  & 12.56ms/2.63s \\
          & ACMII-Snowball-2 & 0.1   & 1.00E-02 & 0.4   & 64    & 2     & -     & -     & -     & -     & 95.25 & 1.55  & 9.74ms/1.97s \\
          & ACMII-Snowball-3 & 0.05  & 1.00E-02 & 0.6   & 64    & 3     & -     & -     & -     & -     & 94.75 & 3.09  & 11.91ms/2.42s \\
          \midrule
    \multirow{10}[0]{*}{\textbf{Film}} & ACM-SGC-1 & 0.05  & 5.00E-05 & 0.7   & 64    & -     & -     & -     & -     & -     & 39.33 & 1.25  & 5.21ms/2.33s \\
          & ACM-SGC-2 & 0.1   & 5.00E-05 & 0.7   & 64    & -     & -     & -     & -     & -     & 40.13 & 1.21  & 12.41ms/4.87s \\
          & ACM-GCN & 0.1   & 5.00E-04 & 0.5   & 64    & 2     & -     & -     & -     & -     & 41.62 & 1.15  & 10.72ms/2.66s \\
          & ACMII-GCN & 0.1   & 5.00E-04 & 0.5   & 64    & 2     & -     & -     & -     & -     & 41.24 & 1.16  & 10.51ms/2.44s \\
          & ACM-GCNII & 0.01  & 0.00E+00 & 0.5   & 64    & 3     & -     & -     & 1.5   & 0.2   & 41.37 & 1.37  & 13.65ms/2.74s \\
          & ACM-GCNII* & 0.01  & 1.00E-05 & 0.5   & 64    & 3     & -     & -     & 1.5   & 0.1   & 41.27 & 1.24  & 14.98ms/3.01s \\
          & ACM-Snowball-2 & 0.1   & 5.00E-03 & 0     & 64    & 2     & -     & -     & -     & -     & 41.4  & 1.23  & 10.30ms/2.08s \\
          & ACM-Snowball-3 & 0.05  & 1.00E-02 & 0     & 64    & 3     & -     & -     & -     & -     & 41.27 & 0.8   & 16.43ms/3.52s \\
          & ACMII-Snowball-2 & 0.1   & 5.00E-03 & 0     & 64    & 2     & -     & -     & -     & -     & 41.1  & 0.75  & 10.74ms/2.19s \\
          & ACMII-Snowball-3 & 0.05  & 5.00E-03 & 0.2   & 64    & 3     & -     & -     & -     & -     & 40.31 & 1.6   & 16.31ms/3.29s \\
          \midrule
    \multirow{10}[0]{*}{\textbf{Chameleon}} & ACM-SGC-1 & 0.1   & 5.00E-06 & 0.9   & 64    & -     & -     & -     & -     & -     & 63.68 & 1.62  & 5.41ms/1.21s \\
          & ACM-SGC-2 & 0.1   & 5.00E-06 & 0.9   & 64    & -     & -     & -     & -     & -     & 60.48 & 1.55  & 7.86ms/1.81s \\
          & ACM-GCN & 0.01  & 5.00E-05 & 0.8   & 64    & 2     & -     & -     & -     & -     & 68.18 & 1.67  & 10.55ms/3.12s \\
          & ACMII-GCN & 0.05  & 5.00E-05 & 0.7   & 64    & 2     & -     & -     & -     & -     & 68.38 & 1.36  & 10.90ms/2.39s \\
          & ACM-GCNII & 0.01  & 5.00E-06 & 0.5   & 64    & 4     & -     & -     & 0.5   & 0.1   & 58.73 & 2.52  & 18.31ms/3.68s \\
          & ACM-GCNII* & 0.01  & 1.00E-03 & 0.5   & 64    & 1     & -     & -     & 1     & 0.1   & 61.66 & 2.29  & 6.68ms/1.40s \\
          & ACM-Snowball-2 & 0.05  & 5.00E-05 & 0.7   & 64    & 2     & -     & -     & -     & -     & 68.51 & 1.7   & 9.92ms/2.06s \\
          & ACM-Snowball-3 & 0.01  & 1.00E-04 & 0.7   & 64    & 3     & -     & -     & -     & -     & 68.4  & 2.05  & 14.49ms/3.15s \\
          & ACMII-Snowball-2 & 0.1   & 5.00E-05 & 0.6   & 64    & 2     & -     & -     & -     & -     & 67.83 & 2.63  & 9.99ms/2.10s \\
          & ACMII-Snowball-3 & 0.05  & 1.00E-04 & 0.7   & 64    & 3     & -     & -     & -     & -     & 67.53 & 2.83  & 15.03ms/3.29s \\
          \midrule
    \multirow{10}[0]{*}{\textbf{Squirrel}} & ACM-SGC-1 & 0.05  & 0.00E+00 & 0.9   & 64    & -     & -     & -     & -     & -     & 46.4  & 1.13  & 6.96ms/2.16s \\
          & ACM-SGC-2 & 0.05  & 0.00E+00 & 0.9   & 64    & -     & -     & -     & -     & -     & 40.91 & 1.39  & 35.20ms/10.66s \\
          & ACM-GCN & 0.05  & 5.00E-06 & 0.6   & 64    & 2     & -     & -     & -     & -     & 58.02 & 1.86  & 14.35ms/2.98s \\
          & ACMII-GCN & 0.05  & 0.00E+00 & 0.7   & 64    & 2     & -     & -     & -     & -     & 53.76 & 1.63  & 14.08ms/3.39s \\
          & ACM-GCNII & 0.01  & 1.00E-05 & 0.5   & 64    & 4     & -     & -     & 0.5   & 0.1   & 40.9  & 1.58  & 20.72ms/4.17s \\
          & ACM-GCNII* & 0.01  & 1.00E-03 & 0.5   & 64    & 4     & -     & -     & 0.5   & 0.3   & 38.32 & 1.5   & 21.78ms/4.38s \\
          & ACM-Snowball-2 & 0.05  & 5.00E-06 & 0.6   & 64    & 2     & -     & -     & -     & -     & 55.97 & 2.03  & 15.38ms/3.15s \\
          & ACM-Snowball-3 & 0.01  & 1.00E-04 & 0.6   & 64    & 3     & -     & -     & -     & -     & 55.73 & 2.39  & 26.15ms/5.94s \\
          & ACMII-Snowball-2 & 0.1   & 5.00E-06 & 0.6   & 64    & 2     & -     & -     & -     & -     & 53.48 & 0.6   & 15.54ms/3.19s \\
          & ACMII-Snowball-3 & 0.05  & 5.00E-05 & 0.5   & 64    & 3     & -     & -     & -     & -     & 52.31 & 1.57  & 26.24ms/5.30s \\
          \midrule
    \multirow{10}[0]{*}{\textbf{Cora}} & ACM-SGC-1 & 0.01  & 5.00E-06 & 0.9   & 64    & -     & -     & -     & -     & -     & 86.63 & 1.13  & 6.00ms/7.40s \\
          & ACM-SGC-2 & 0.1   & 5.00E-05 & 0.6   & 64    & -     & -     & -     & -     & -     & 87.64 & 0.99  & 4.85ms/1.17s \\
          & ACM-GCN & 0.1   & 5.00E-03 & 0.5   & 64    & 2     & -     & -     & -     & -     & 88.62 & 1.22  & 8.84ms/1.81s \\
          & ACMII-GCN & 0.1   & 5.00E-03 & 0.4   & 64    & 2     & -     & -     & -     & -     & 89    & 0.72  & 8.93ms/1.83s \\
          & ACM-GCNII & 0.01  & 1.00E-03 & 0.5   & 64    & 3     & -     & -     & 1     & 0.2   & 89.1  & 1.61  &  14.07ms/3.04s \\
          & ACM-GCNII* & 0.01  & 1.00E-02 & 0.5   & 64    & 4     & -     & -     & 1     & 0.2   & 89    & 1.35  & 11.36ms/2.48s \\
          & ACM-Snowball-2 & 0.05  & 1.00E-03 & 0.6   & 64    & 2     & -     & -     & -     & -     & 88.83 & 1.49  & 9.34ms/1.92s \\
          & ACM-Snowball-3 & 0.1   & 1.00E-02 & 0.3   & 64    & 3     & -     & -     & -     & -     & 89.59 & 1.58  & 13.33ms/2.75s \\
          & ACMII-Snowball-2 & 0.1   & 5.00E-03 & 0.5   & 64    & 2     & -     & -     & -     & -     & 88.95 & 1.04  & 9.29ms/1.90s \\
          & ACMII-Snowball-3 & 0.1   & 5.00E-03 & 0.5   & 64    & 3     & -     & -     & -     & -     & 89.36 & 1.26  & 14.18ms/2.89s \\
          \midrule
    \multirow{10}[0]{*}{\textbf{CiteSeer}} & ACM-SGC-1 & 0.01  & 5.00E-04 & 0.9   & 64    & -     & -     & -     & -     & -     & {80.96} & 0.93  & 5.90ms/4.31s \\
          & ACM-SGC-2 & 0.05  & 5.00E-04 & 0.9   & 64    & -     & -     & -     & -     & -     & 80.93 & 1.16  & 5.01ms/1.42s \\
          & ACM-GCN & 0.05  & 5.00E-03 & 0.7   & 64    & 2     & -     & -     & -     & -     & 81.68 & 0.97  & 11.35ms/2.57s \\
          & ACMII-GCN & 0.05  & 5.00E-05 & 0.7   & 64    & 2     & -     & -     & -     & -     & 81.58 & 1.77  & 9.55ms/1.94s \\
          & ACM-GCNII & 0.01  & 1.00E-02 & 0.5   & 64    & 3     & -     & -     & 0.5   & 0.3   & 82.28 & 1.12  & 15.61ms/3.56s \\
          & ACM-GCNII* & 0.01  & 1.00E-02 & 0.5   & 64    & 3     & -     & -     & 0.5   & 0.5   & 81.69 & 1.25  & 15.56ms/3.61s \\
          & ACM-Snowball-2 & 0.05  & 5.00E-03 & 0.7   & 64    & 2     & -     & -     & -     & -     & 81.58 & 1.23  & 11.14ms/2.50s \\
          & ACM-Snowball-3 & 0.01  & 5.00E-03 & 0.9   & 64    & 3     & -     & -     & -     & -     & 81.32 & 0.97  & 15.91ms/5.36s \\
          & ACMII-Snowball-2 & 0.05  & 5.00E-03 & 0.7   & 64    & 2     & -     & -     & -     & -     & 82.07 & 1.04  & 10.97ms/2.55s \\
          & ACMII-Snowball-3 & 0.05  & 1.00E-04 & 0.6   & 64    & 3     & -     & -     & -     & -     & 81.56 & 1.15  & 14.95ms/3.03s \\
          \midrule
    \multirow{10}[0]{*}{\textbf{PubMed}} & ACM-SGC-1 & 0.05  & 5.00E-06 & 0.3   & 64    & -     & -     & -     & -     & -     & 87.75 & 0.88  & 6.04ms/2.61s \\
          & ACM-SGC-2 & 0.05  & 5.00E-05 & 0.1   & 64    & -     & -     & -     & -     & -     & 88.79 & 0.5   & 8.62ms/3.18s \\
          & ACM-GCN & 0.1   & 5.00E-04 & 0.2   & 64    & 2     & -     & -     & -     & -     & 90.54 & 0.63  & 10.20ms/2.08s \\
          & ACMII-GCN & 0.1   & 5.00E-04 & 0.2   & 64    & 2     & -     & -     & -     & -     & 90.74 & 0.5   & 10.20ms/2.07s \\
          & ACM-GCNII & 0.01  & 1.00E-04 & 0.5   & 64    & 3     & -     & -     & 1.5   & 0.5   & 90.12 & 0.4   & 15.07ms/3.35s \\
          & ACM-GCNII* & 0.01  & 1.00E-04 & 0.5   & 64    & 3     & -     & -     & 1.5   & 0.5   & 90.18 & 0.51  & 16.62ms/3.72s \\
          & ACM-Snowball-2 & 0.1   & 1.00E-04 & 0.3   & 64    & 2     & -     & -     & -     & -     & 90.81 & 0.52  & 11.52ms/2.36s \\
          & ACM-Snowball-3 & 0.05  & 1.00E-03 & 0.2   & 64    & 3     & -     & -     & -     & -     & 91.44 & 0.59  & 18.06ms/3.69s \\
          & ACMII-Snowball-2 & 0.1   & 1.00E-04 & 0.3   & 64    & 2     & -     & -     & -     & -     & 90.56 & 0.39  & 11.74ms/2.39s \\
          & ACMII-Snowball-3 & 0.1   & 5.00E-04 & 0.2   & 64    & 3     & -     & -     & -     & -     & 91.31 & 0.6   & 18.61ms/3.88s \\
          \midrule
    \multirow{6}[1]{*}{\textbf{Deezer-Europe}} & ACM-SGC-1 & 0.05  & 0,5e-6,1e-5,5e-5 & 0.3   & 64    & -     & -     & -     & -     & -     & 66.67 & 0.56  & 146.41ms/73.06s \\
          & ACM-SGC-2 & 0.002 & {5e-5,1e-4} & 0.3   & 64    & -     & -     & -     & -     & -     & 66.53 & 0.57  & 195.21ms/97.41s \\
          & ACM-GCN & 0.002 & 5.00E-04 & 0.5   & 64    & 2     & -     & -     & -     & -     & 67.01 & 0.38  & 136.45ms/68.09s \\
          & ACMII-GCN & 0.01  & 5.00E-05 & 0.8   & 64    & 2     & -     & -     & -     & -     & 67.15 & 0.41  & 135.24ms/67.48s \\
          & ACM-GCNII & 0.01  & 0,5e-6 & 0.5   & 64    & 1     & -     & -     & 0.5   & 0.4   & 66.39 & 0.56  & 80.82ms/40.33s \\
          & ACM-GCNII* & 0.01  & 0.0001,1e-3 & 0.5   & 64    & 1     & -     & -     & 1.5   & 0.2   & 66.6  & 0.57  & 80.95ms/40.40s \\
    \bottomrule
    \bottomrule
    \end{tabular}%
    \label{tab:hyperparameters_acmgnns}%
    }
  
\end{table}%

\subsection{Details of the Implementation of ACM and ACMII Frameworks}
\label{appendix:details_implementation_acm_acmII}
In ACM(II) framework, we first use dropout operation over the input data. The implementation of ACM(II)-GCN and ACM(II)-snowball is straightforward, but SGC-1, SGC-2, GCNII and GCNII* are not able to be applied under ACM(II) framework and we will make an explanation as follows.
\begin{itemize}
    \item SGC-1 and SGC-2: SGC does not contain nonlinearity, so the option 1 and option 2 in step 1 is the same for ACM-SGC and ACMII-SGC. Thus, we only implement ACM-SGC.
    \item GCNII and GCNII*: 
    \begin{align*}
       & \text{GCCII: } \mathbf{H}^{(\ell+1)}=\sigma\left(\left(\left(1-\alpha_{\ell}\right) \hat{\mathbf{A}} \mathbf{H}^{(\ell)}+\alpha_{\ell} \mathbf{H}^{(0)}\right)\left(\left(1-\beta_{\ell}\right) \mathbf{I}_{n}+\beta_{\ell} \mathbf{W}^{(\ell)}\right)\right)\\
       & \text{GCCII*: }
       \mathbf{H}^{(\ell+1)}= \sigma\left(\left(1-\alpha_{\ell}\right) \hat{\mathbf{A}} \mathbf{H}^{(\ell)}\left(\left(1-\beta_{\ell}\right) \mathbf{I}_{n}+\beta_{\ell} \mathbf{W}_{1}^{(\ell)}\right)+\right. \left.+\alpha_{\ell} \mathbf{H}^{(0)}\left(\left(1-\beta_{\ell}\right) \mathbf{I}_{n}+\beta_{\ell} \mathbf{W}_{2}^{(\ell)}\right)\right)
    \end{align*}
\end{itemize}
Without major modification, GCNII and GCNII* are hard to be put into ACMII framework. In ACMII frameworks, before apply the operator $\hat{A}$, we first implement a nonlinear feature operation over $H^{\ell}$. But in GCNII and GCNII*, before multiplying $W^\ell (\text{or }W_1^\ell, W_2^\ell)$ to extract features, we need to add another term including $H^{(0)}$, which are not filtered by $\hat{A}$. This is incompatible with ACMII framework, thus, we did not implement GCNII and GCNII* in ACMII framework. 

The open source code will be released soon.

\subsection{Computing Resources}
For all experiments on synthetic datasets and real-world datasets, we use NVidia V100 GPUs with 16/32GB GPU memory, 8-core CPU, 16G Memory. The
software implementation is based on PyTorch and PyTorch Geometric \cite{fey2019fast}.
\clearpage
\section{Details of Gradient Calculation in \eqref{eq:gradient_descent_update}}
\label{appendix:details_of_nll_loss_explanation}
\subsection{Derivation in Matrix Form}
In output layer, we have
\begin{align*}
Y & = \text{softmax} (\hat{A} X W ) \equiv  \text{softmax} (Y') = \left(\exp(Y') \bm{1}_C \bm{1}_C^T \right)^{-1}  \odot \exp(Y') > 0 \\   
\mathcal{L} & = -\trace(Z^T \log Y)    
\end{align*}
where $\bm{1}_C \in \mathcal{R}^{C\times 1}$, $(\cdot)^{-1}$ is point-wise inverse function and each element of $Y$ is positive. Then
\begin{align*}
d \mathcal{L} = -\trace\left(Z^T ((Y)^{-1} \odot d Y) \right) = -\trace\left(Z^T \left( \left(\text{softmax} (Y') \right)^{-1} \odot d\ \text{softmax} ( Y') \right) \right) 
\end{align*}
Note that
\begin{equation}
\begin{aligned}
d\ \text{softmax} (Y') 
= & - \left(\exp(Y') \bm{1}_C \bm{1}_C^T \right)^{-2} \odot [(\exp(Y') \odot d Y') \bm{1}_C \bm{1}_C^T] \odot \exp(Y')  \\ \nonumber
& + \left(\exp(Y') \bm{1}_C \bm{1}_C^T \right)^{-1}  \odot (\exp(Y') \odot d Y')\\ \nonumber
 = & - \text{softmax} (Y') \odot \left(\exp(Y') \bm{1}_C \bm{1}_C^T \right)^{-1} \odot [(\exp(Y') \odot d Y') \bm{1}_C \bm{1}_C^T]   \\
& + \text{softmax} (Y') \odot d Y'\\ \nonumber
 =  & \ \text{softmax} (Y') \odot \left(  - \left(\exp(Y') \bm{1}_C \bm{1}_C^T \right)^{-1} \odot \left[(\exp(Y') \odot d Y') \bm{1}_C \bm{1}_C^T \right] + d Y' \right) \nonumber
\end{aligned}
\end{equation}
Then,
\begin{equation}
\begin{aligned}
d \mathcal{L} =\ & -\trace\Bigg(Z^T \bigg((\text{softmax} (Y'))^{-1} \odot \bigg[ \text{softmax} (Y') \odot \bigg(  - \left(\exp(Y') \bm{1}_C \bm{1}_C^T \right)^{-1} \\
& \odot \left[(\exp(Y') \odot d Y') \bm{1}_C \bm{1}_C^T \right] + d Y' \bigg) \bigg] \bigg) \Bigg)\\ \nonumber
=\ & -\trace\left(Z^T \left(  - \left(\exp(Y') \bm{1}_C \bm{1}_C^T \right)^{-1} \odot \left[(\exp(Y') \odot d Y') \bm{1}_C \bm{1}_C^T \right] + d Y' \right) \right) \\ \nonumber
 =\ & \trace\left( \left( \left(Z \odot \left(\exp(Y') \bm{1}_C \bm{1}_C^T \right)^{-1} \right) \bm{1}_C \bm{1}_C^T \right)^T  \left[\exp(Y') \odot d Y'  \right] - Z^T d Y' \right)\\ \nonumber
=\ &  \trace\left( \left(  \exp(Y') \odot \left( \left(Z \odot \left(\exp(Y') \bm{1}_C \bm{1}_C^T \right)^{-1} \right) \bm{1}_C \bm{1}_C^T \right) \right)^T  d Y' - Z^T d Y' \right)\\ \nonumber
=\ &  \trace\left( \left(  \exp(Y') \odot \left(\exp(Y') \bm{1}_C \bm{1}_C^T \right)^{-1} \right)^T   d Y' - Z^T d Y' \right)\\ \nonumber
=\ & \trace\left( (\text{softmax}(Y') - Z)^T dY' \right)  \nonumber
\end{aligned}
\end{equation}
where the  4-th equation holds due to $\left(Z \odot \left(\exp(Y') \bm{1}_C \bm{1}_C^T \right)^{-1} \right) \bm{1}_C \bm{1}_C^T = \left(\exp(Y') \bm{1}_C \bm{1}_C^T \right)^{-1}$. Thus, we have
\begin{equation*}
    \frac{d \mathcal{L} }{d Y'} = \text{softmax}(Y') - Z = Y-Z
\end{equation*}
For $Y'$ and $W$, we have
\begin{align*}
   d Y' & = \hat{A} X d W \text{  and  } d\mathcal{L}= \text{trace}\left(\frac{d \mathcal{L} }{d Y'}^T d Y' \right) = \text{trace}\left(\frac{d \mathcal{L} }{d Y'}^T \hat{A}  X \ d W \right) = \text{trace}\left(\frac{d \mathcal{L} }{d W}^T \ d W \right) 
\end{align*}
To get $\frac{d \mathcal{L} }{d W}$ we have, 
\begin{equation}\label{eq5}
\begin{aligned}
   \frac{d \mathcal{L}}{d W} = X^T \hat{A}^T \frac{d \mathcal{L} }{ d Y'} = X^T \hat{A}^T (Y-Z)
\end{aligned}
\end{equation}

\subsection{Component-wise Derivation}
Denote $\tilde{X}=XW$. 
We rewrite $\cal L$ as follows:
    \begin{align*}
    \label{eq:nll_loss_explanation_details}
    \mathcal{L} & = -\trace\left(Z^T \log \left((\exp({Y'}) \bm{1}_C \bm{1}_C^T )^{-1}  \odot \exp({Y'})\right) \right) \\
    & = -\trace\left(Z^T  \left(-\log(\exp({Y'}) \bm{1}_C \bm{1}_C^T ) + {Y'} \right) \right) \\
    & = -\trace\left(Z^T {Y'} \right) + \trace\left(Z^T  \log\left(\exp({Y'}) \bm{1}_C \bm{1}_C^T \right)  \right)\\
    & = -\trace\left(Z^T \hat{A} X W  \right) + \trace\left(Z^T  \log\left(\exp({Y'}) \bm{1}_C \bm{1}_C^T \right)  \right)\\
    & = -\trace\left(Z^T \hat{A} X W  \right) + \trace\left(\bm{1}_C^T  \log\left(\exp({Y'}) \bm{1}_C \right)  \right)\\
    & = - \sum\limits_{i=1}^N \sum\limits_{j\in \mathcal{N}_i}  \hat{A}_{i,j} Z_{i,:}\tilde{X}_{j:}^T + \sum\limits_{i=1}^N \log \left( \sum\limits_{c=1}^C \exp(\sum\limits_{j\in \mathcal{N}_i} \hat{A}_{i,j} \tilde{X}_{j,c}) \right)\\
    & = - \sum\limits_{i=1}^N \log \left( \exp\left( \sum\limits_{c=1}^C \sum\limits_{j\in \mathcal{N}_i}  \hat{A}_{i,j} Z_{i,c}\tilde{X}_{j,c} \right) \right) + \sum\limits_{i=1}^N \log \left( \sum\limits_{c=1}^C \exp\left(\sum\limits_{j\in \mathcal{N}_i} \hat{A}_{i,j} \tilde{X}_{j,c} \right) \right)\\
    &= - \sum\limits_{i=1}^N \log \frac{\exp \left(\sum\limits_{c=1}^C \sum\limits_{j\in \mathcal{N}_i} \hat{A}_{i,j} Z_{i,c}\tilde{X}_{j,c}\right)}{\left(\sum\limits_{c=1}^C \exp(\sum\limits_{j\in \mathcal{N}_i} \hat{A}_{i,j} \tilde{X}_{j,c}) \right)}
    \end{align*}
Note that $\sum\limits_{c=1}^C Z_{j,c} = 1$ for any $j$. 
Consider the derivation of $\mathcal{L}$ over $\tilde{X}_{j',c'}$:
    \begin{align*}
    &\frac{d \mathcal{L}}{d \tilde{X}_{j',c'}} \\
     = & - \sum\limits_{i=1}^N \frac{\sum\limits_{c=1}^C \exp(\sum\limits_{j\in \mathcal{N}_i} \hat{A}_{i,j} \tilde{X}_{j,c}) }{\exp \left(\sum\limits_{c=1}^C \sum\limits_{j\in \mathcal{N}_i} \hat{A}_{i,j} Z_{i,c}\tilde{X}_{j,c}\right) } \\ 
& \times \left( \frac{ \left( \hat{A}_{i,j'} Z_{i,c'}  \right) \exp \left(\sum\limits_{c=1}^C \sum\limits_{j\in \mathcal{N}_i} \hat{A}_{i,j} Z_{i,c}\tilde{X}_{j,c}\right) \left(\sum\limits_{c=1}^C \exp(\sum\limits_{j\in \mathcal{N}_i} \hat{A}_{i,j} \tilde{X}_{j,c}) \right) }{ \left(\sum\limits_{c=1}^C \exp(\sum\limits_{j\in \mathcal{N}_i} \hat{A}_{i,j} \tilde{X}_{j,c}) \right)^2} \right.\\
    &\left. - \frac{\left( \hat{A}_{i,j'} \right) \exp \left(\sum\limits_{c=1}^C \sum\limits_{j\in \mathcal{N}_i} \hat{A}_{i,j} Z_{i,c}\tilde{X}_{j,c}\right) \left( \exp(\sum\limits_{j\in \mathcal{N}_i} \hat{A}_{i,j} \tilde{X}_{j,c'}) \right) }{ \left(\sum\limits_{c=1}^C \exp(\sum\limits_{j\in \mathcal{N}_i} \hat{A}_{i,j} \tilde{X}_{j,c}) \right)^2} \right) \\
     =&  - \sum\limits_{i=1}^N \left( \frac{ \left( \hat{A}_{i,j'} Z_{i,c'}  \right) \left(\sum\limits_{c=1}^C \exp(\sum\limits_{j\in \mathcal{N}_i} \hat{A}_{i,j} \tilde{X}_{j,c}) \right) - \left( \hat{A}_{i,j'} \right) \left( \exp(\sum\limits_{j\in \mathcal{N}_i} \hat{A}_{i,j} \tilde{X}_{j,c'}) \right)}{ \left(\sum\limits_{c=1}^C \exp(\sum\limits_{j\in \mathcal{N}_i} \hat{A}_{i,j} \tilde{X}_{j,c}) \right)} \right)\\
     = & - \sum\limits_{i=1}^N \left( \hat{A}_{i,j'}  \frac{\left(\sum\limits_{c=1, c\neq c'}^C (Z_{i,c'}) \exp(\sum\limits_{j\in \mathcal{N}_i} \hat{A}_{i,j} \tilde{X}_{j,c}) \right) +  \left( Z_{i,c'}-1 \right) \left( \exp(\sum\limits_{j\in \mathcal{N}_i} \hat{A}_{i,j} \tilde{X}_{j,c'}) \right)}{ \left(\sum\limits_{c=1}^C \exp(\sum\limits_{j\in \mathcal{N}_i} \hat{A}_{i,j} \tilde{X}_{j,c}) \right)} \right)\\
     =& - \sum\limits_{i=1}^N \hat{A}_{i,j'} \left( Z_{i,c'} \hat{P}(Y_i \neq c') + (Z_{i,c'} - 1) \hat{P} (Y_i = c') \right)\\
     = & - \sum\limits_{i=1}^N \hat{A}_{i,j'} \left( Z_{i,c'} - \hat{P} (Y_i = c') \right)
    \end{align*}
Writing the above in matrix form, we have
\begin{equation}
    \frac{d \mathcal{L}}{d \tilde{X} } = \hat{A}(Z- Y), \  \frac{d \mathcal{L}}{d \tilde{W} } = X^T \hat{A}^T (Z-Y), \ \Delta Y' \propto \hat{A}XX^T\hat{A}^T(Z-Y)
\end{equation}

\section{Details of Synthetic Experiments}
\label{appendix:details_syn_exps}
In our synthetic experiments, we generate graphs with edge homophily levels $h \in 0.95:0.05:0.05$ and $h \in 0.05:0.005:0.005$. We explore the interval $[0.05, 0.005]$ with a more fine-grained scale 0.005 because we empirically find that the performance of GNNs is sensitive in more this area. For a given $h$, we generate intra-class edges from $\texttt{numpy.random.multinomial(2, numpy.ones(399)/399, size=1)[0]}$ (does not include self-loop) and inter-class edges from \texttt{numpy.random.multinomial(int(2/h -2), numpy.ones(1600)/1600, size=1)[0]}.

\section{Proof of Theorem 1}
\label{appendix:proof_theorem1}
\begin{proof}

According to the given assumptions, for node $v$, we have $\hat{A}_{v,k}=\frac{1}{d+1}$, the expected number of intra-class edges is $dh$ (here the self-loop edge introduced by $\hat{A}$ is not counted based on the definition of edge homophily and data generation process) and inter-class edges is $(1-h)d$. Suppose there are $C \geq 2$ classes. Consider matrix $\hat{A}Z$,

Then, we have $\mathbb{E}\left[(\hat{A}Z)_{v,c}\right] =  \mathbb{E}\left[\sum\limits_{k \in \mathcal{V}} \hat{A}_{v,k} \textbf{1}_{\{Z_{k,:}= e_c^T\}}\right] = \sum\limits_{k \in \mathcal{V}} \frac{\mathbb{E}\left[\textbf{1}_{\{Z_{k,:}= e_c^T\}}\right] }{d+1}$, where $\bm{1}$ is the indicator function.

When $v$ is in class $c$, we have $\sum\limits_{k \in \mathcal{V}} \frac{\mathbb{E}\left[ \textbf{1}_{\{Z_{k,:}= e_c^T\}}\right]}{d+1} = \frac{hd+1}{d+1}$ ($hd+1=hd$ intra-class edges $+$ 1 self-loop introduced by $\hat{A}$).

When $v$ is not in class $c$, we have $\sum\limits_{k \in \mathcal{V}} \frac{\mathbb{E}\left[ \textbf{1}_{\{Z_{k,:}= e_c^T\}}\right]}{d+1} = \frac{(1-h)d}{(C-1)(d+1)}$ ($(1-h)d$ inter-class edges uniformly distributed in the other $C-1$ classes).

For nodes $v,u$, we have $(\hat{A}Z)_{v,:}, (\hat{A}Z)_{u,:} \in \mathbb{R}^C$ and since elements in $\hat{A}_{v,k}$ and $\hat{A}_{u,k'}$ are independently generated for all $k,k' \in \mathcal{V}$, we have
\begin{align*}
    \mathbb{E}\left[(\hat{A}Z)_{v,c} (\hat{A}Z)_{u,c}\right] & =  \mathbb{E}\left[(\sum\limits_{k \in \mathcal{V}} \hat{A}_{v,k} \textbf{1}_{\{Z_{k,:}= e_c^T\}}) (\sum\limits_{k' \in \mathcal{V}} \hat{A}_{u,k'} \textbf{1}_{\{Z_{k',:}= e_c^T\}}) \right] \\
    & = \mathbb{E}\left[(\sum\limits_{k \in \mathcal{V}} \hat{A}_{v,k} \textbf{1}_{\{Z_{k,:}= e_c^T\}}) \right] \mathbb{E} \left[ (\sum\limits_{k' \in \mathcal{V}} \hat{A}_{u,k'} \textbf{1}_{\{Z_{k',:}= e_c^T\}}) \right]
\end{align*}
Thus,

\begin{align*}
  \mathbb{E}\left[ S(\hat{A},Z)_{v,u} \right] 
  & = \mathbb{E}\left[<(\hat{A}Z)_{v,:}, (\hat{A}Z)_{u,:}> \right] = \sum\limits_c \mathbb{E}\left[(\sum\limits_{k \in \mathcal{V}} \hat{A}_{v,k} \textbf{1}_{\{Z_{k,:}= e_c^T\}}) \right] \mathbb{E} \left[ (\sum\limits_{k' \in \mathcal{V}} \hat{A}_{u,k'} \textbf{1}_{\{Z_{k',:}= e_c^T\}}) \right]\\
  & = \left\{
             \begin{array}{ll}
              \left(\frac{hd+1}{d+1}\right)^2 + \frac{\left( (1-h)d \right)^2}{(C-1)(d+1)^2}  , & \text{$u,v$ are in the same class} \\
               \frac{2(hd+1)(1-h)d}{(C-1)(d+1)^2}+ \frac{(C-2)(1-h)^2 d^2}{(C-1)^2 (d+1)^2}  , & \text{$u,v$ are in different classes} 
             \end{array}\right.
\end{align*}
For nodes $u_1$, $u_2$, and $v$,
where $Z_{u_1,:}=Z_{v,:}$ and $Z_{u_2,:}\neq Z_{v,:}$, 
\begin{align}
\label{eq:expectation_of_similarity_element}
    g(h) & \equiv \mathbb{E}\left[ S(\hat{A},Z)_{v,u_1} \right] - \mathbb{E}\left[ S(\hat{A},Z)_{v,u_2} \right] \\ \nonumber
    & = \frac{  (C-1)^2(hd+1)^2 + (C-1)\left[(1-h)d\right]^2 - (C-1)\left(2(hd+1)(1-h)d \right) - (C-2)\left[(1-h)d\right]^2}{(C-1)^2(d+1)^2} \\ \nonumber
 & 
 = \left(\frac{(C-1)(hd+1)-(1-h)d}{(C-1)(d+1)} \right)^2
\end{align}
Setting $g(h)=0$, we obtain the optimal $h$:
\beq \label{eq:optimalh}
h=\frac{d+1-C}{Cd} 
\eeq
For the data generation process in the synthetic experiments, 
we fix $d_\text{intra}$, then $d=d_\text{intra}/h$, which is a function of $h$.
We change $d$ in \eqref{eq:optimalh} to $d_\text{intra}/h$, leading to
\beq \label{eq:newh}
h = \frac{d_\text{intra}/h+1-C}{Cd_\text{intra}/h}
\eeq
It is easy to observe that $h$ satisfying \eqref{eq:newh} still makes $g(h)=0$, when $d$ in $g(h)$ 
is replaced by $d_\text{intra}/h$.
From \eqref{eq:newh} we obtain the optimal $h$ in terms of $d_\text{intra}$:
$$
h=\frac{d_{\text{intra}}}{C d_{\text{intra}} +C-1}
$$

\end{proof}
\subsection{An extension of Theorem 1}
\begin{align*}
    S_\text{agg}\left(S(\hat{A},Z)\right) &= \frac{\left| \left\{v   \,\big| \,
    \mathrm{Mean}_u\big( \{S(\hat{A},Z)_{v,u} | Z_{u,:}=Z_{v,:} \}\big) 
    \geq \mathrm{Mean}_u\big(\{S(\hat{A},Z)_{v,u} | Z_{u,:} \neq Z_{v,:} \} \big) \right\} \right|}{\left| \mathcal{V} \right|}\\
    & = \frac{ \sum\limits_{v\in \mathcal{V}} \bm{1}_{ \left\{
    \mathrm{Mean}_u \big( \{S(\hat{A},Z)_{v,u} | Z_{u,:}=Z_{v,:} \}\big) 
    \geq \mathrm{Mean}_u\big(\{S(\hat{A},Z)_{v,u} | Z_{u,:} \neq Z_{v,:} \} \big) \right\}}}{\left| \mathcal{V} \right|}
\end{align*}
Then,
\begin{align*}
    \mathbb{E}\left(S_\text{agg}\left(S(\hat{A},Z)\right)\right) & = \mathbb{E} \left(\frac{ \sum\limits_{v\in \mathcal{V}} \bm{1}_{ \left\{
    \mathrm{Mean}_u \big( \{S(\hat{A},Z)_{v,u} | Z_{u,:}=Z_{v,:} \}\big) 
    \geq \mathrm{Mean}_u\big(\{S(\hat{A},Z)_{v,u} | Z_{u,:} \neq Z_{v,:} \} \big) \right\}}}{\left| \mathcal{V} \right|} \right)\\
    & =  \frac{ \sum\limits_{v\in \mathcal{V}} \mathbb{P}\left(  
    \mathrm{Mean}_u \big( \{S(\hat{A},Z)_{v,u} | Z_{u,:}=Z_{v,:} \}\big) 
    \geq \mathrm{Mean}_u\big(\{S(\hat{A},Z)_{v,u} | Z_{u,:} \neq Z_{v,:} \} \big)   \right)}{\left| \mathcal{V} \right|}  \\
    & = \mathbb{P}\left(\mathrm{Mean}_u \big( \{S(\hat{A},Z)_{v,u} | Z_{u,:}=Z_{v,:} \}\big) - \mathrm{Mean}_u\big(\{S(\hat{A},Z)_{v,u} | Z_{u,:} \neq Z_{v,:} \} \big) \geq 0 \right) \\
\end{align*}
Consider the random variable
\begin{equation*}
    RV = \mathrm{Mean}_u \big( \{S(\hat{A},Z)_{v,u} | Z_{u,:}=Z_{v,:} \}\big) - \mathrm{Mean}_u\big(\{S(\hat{A},Z)_{v,u} | Z_{u,:} \neq Z_{v,:} \} \big)
\end{equation*}
Since $RV$ is symmetrically distributed and under the conditions in theorem 1, its expectation is $\mathbb{E}[RV] = g(h)$ as showed in \eqref{eq:expectation_of_similarity_element}. Since the minimum of $g(h)$ is $0$ and $RV$ is symmetrically distributed, we have $\mathbb{P}(RV \geq 0) \geq 0.5$ and this can explain why $H_{\text{agg}}(\mathcal{G})$ is always greater than 0.5 in many real-world tasks.
\section{Proof of Theorem 2}
\label{appendix:proof_theorem2}
\begin{proof}
Define $ W_v^{c}=(\hat{A}Z)_{v,c}$. 
Then,
\begin{equation*}
    W_v^{c}=
    \sum\limits_{k\in \mathcal{V}} \hat{A}_{v,k} \bm{1}_{\{Z_{k,:} = e_c^T\}} \in [0,1], \ \ \sum\limits_{c=1}^C W_v^c = 1
\end{equation*}
Note that
\begin{equation} \label{eq:S}
S(I-\hat{A},Z) = (I-\hat{A})ZZ^T(I-\hat{A})^T = ZZ^T + \hat{A}ZZ^T\hat{A}^T - \hat{A}ZZ^T - ZZ^T\hat{A}^T 
\end{equation}
For any node $v$, let the class $v$ belongs to be denoted by $c_v$.
For two nodes $v,u$, if $Z_{v,:} \neq Z_{u,:}$, we have
\begin{align*}
    &(ZZ^T)_{v,u} = 0\\
    & (\hat{A}ZZ^T\hat{A}^T)_{v,u} = \sum\limits_{c=1}^C W_v^c W_u^c \\
    & (\hat{A}ZZ^T)_{v,u} = W_v^{c_u} \\
    & (ZZ^T\hat{A}^T)_{v,u} = (\hat{A}ZZ^T)_{u,v} =  W_u^{c_v}
\end{align*}
Then, from \eqref{eq:S} it follows that
\begin{align*}
    (S(I-\hat{A},Z))_{v,u} = \sum\limits_{c=1}^C W_v^c W_u^c  - W_v^{c_u} - W_u^{c_v}
\end{align*}
When $C=2$,  
\begin{align*}
    S(I-\hat{A},Z)_{v,u} = W_v^{c_u}( W_u^{c_u}-1) + W_u^{c_v} (W_v^{c_v}-1) \leq 0
\end{align*}
If $Z_{v,:} = Z_{u,:}$, \ie{} $c_v=c_u$, we have
\begin{align*}
    &(ZZ^T)_{v,u} = 1\\
    & (\hat{A}ZZ^T\hat{A}^T)_{v,u} = \sum\limits_{c=1}^C W_v^c W_u^c \\
    & (\hat{A}ZZ^T)_{v,u} = W_v^{c_v} \\
    & (ZZ^T\hat{A}^T)_{v,u} = (\hat{A}ZZ^T)_{u,v} =  W_u^{c_u} = W_u^{c_v}
\end{align*}
Then, from \eqref{eq:S} it follows that
\begin{align*}
    S(I-\hat{A},Z)_{v,u} &= 1+\sum\limits_{c=1}^C W_v^c W_u^c  - W_v^{c_v} - W_u^{c_v}\\
    & = \sum\limits_{c=1,c\neq c_v}^C W_v^c W_u^c + 1+ W_v^{c_v}W_u^{c_v} - W_v^{c_v} - W_u^{c_v}\\
    & = \sum\limits_{c=1,c\neq c_v}^C W_v^c W_u^c + (1- W_v^{c_v})(1-W_u^{c_v}) \geq 0
\end{align*}
Thus, if $C=2$, for any $v\in\mathcal{V}$, if $Z_{u,:} \neq Z_{v,:}$, we have $S(I-\hat{A},Z)_{v,u} \leq 0$; if $Z_{u,:} = Z_{v,:}$, we have $S(I-\hat{A},Z)_{v,u} \geq 0$. Apparently, the two conditions in \eqref{eq:diversification_distinguishability} are satisfied.
Thus $v$ is  diversification distinguishable and $\mathrm{DD}_{\hat{A},X}(\mathcal{G})=1$.
The theorem is proved.
\end{proof}

\section{Model Comparison on Synthetic Graphs}
\label{appendix:model_comparison_synthetic_datasets}

\begin{figure}[H]
    \centering
     {
    \subfloat[ \texttt{syn-Cora}]{
     \captionsetup{justification = centering}
     \includegraphics[width=0.32\textwidth]{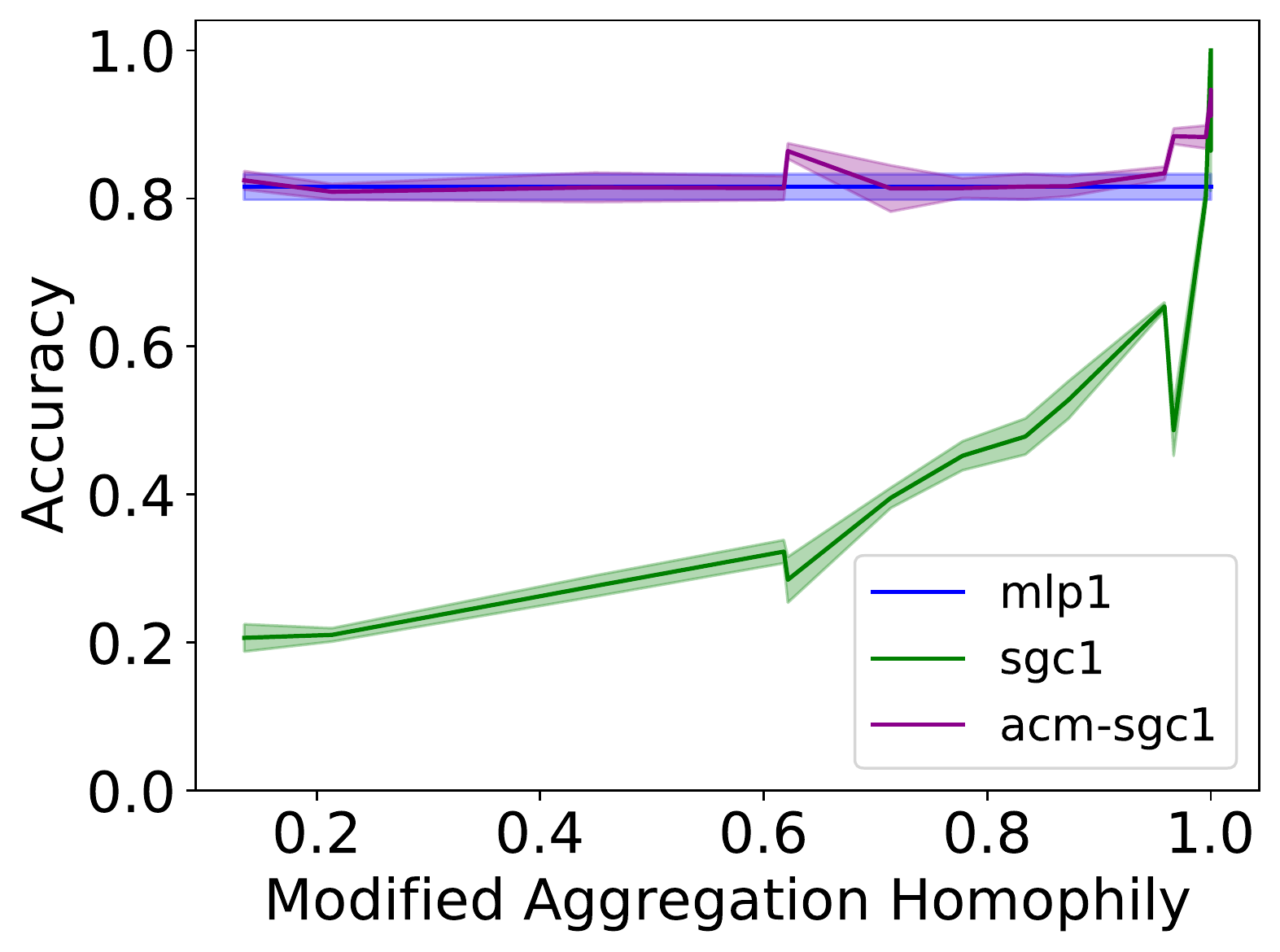}
     } 
     \subfloat[ \texttt{syn-CiteSeer}]{
     \captionsetup{justification = centering}
     \includegraphics[width=0.32\textwidth]{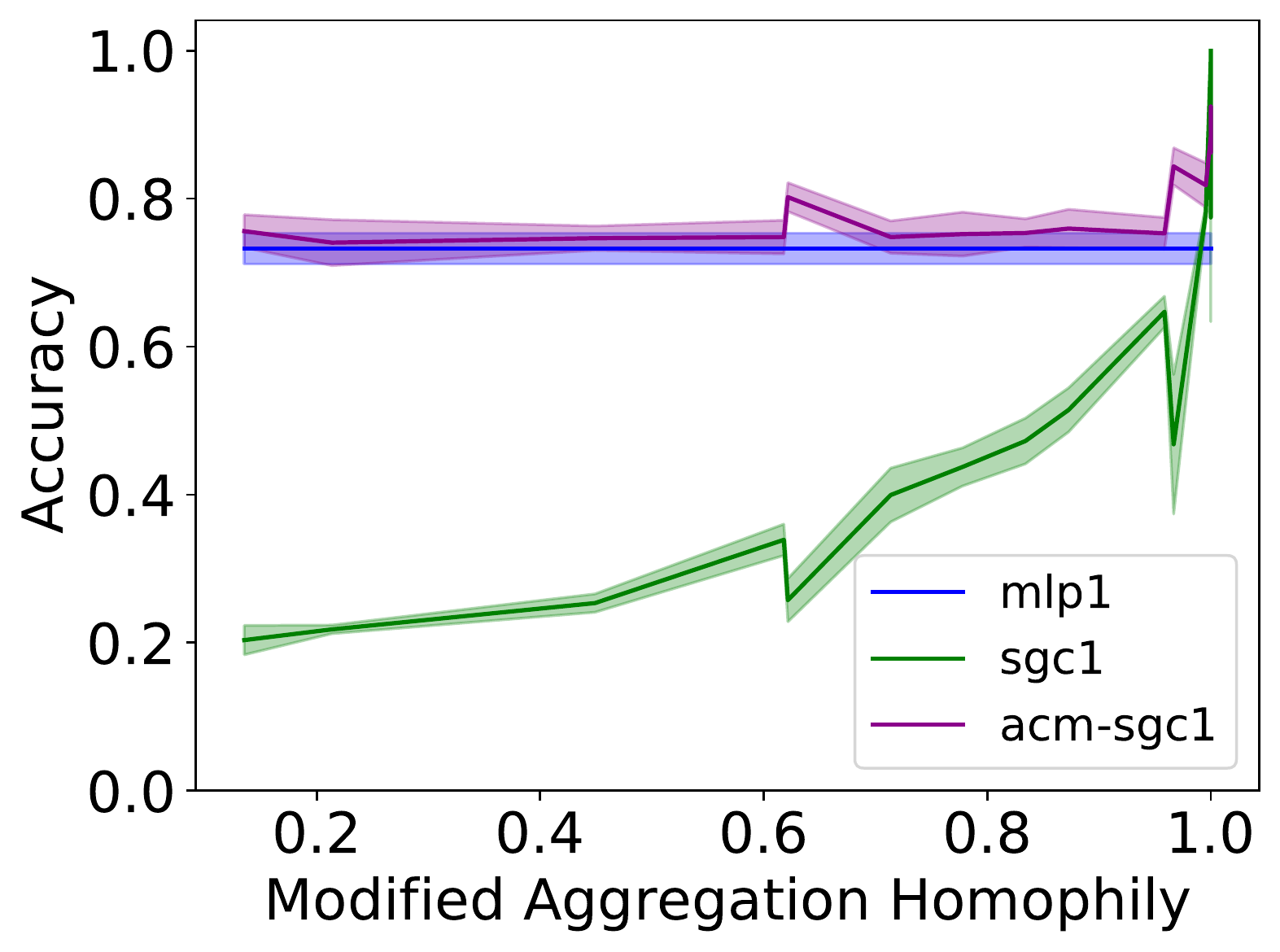}
     } 
     \subfloat[ \texttt{syn-PubMed}]{
     \captionsetup{justification = centering}
     \includegraphics[width=0.32\textwidth]{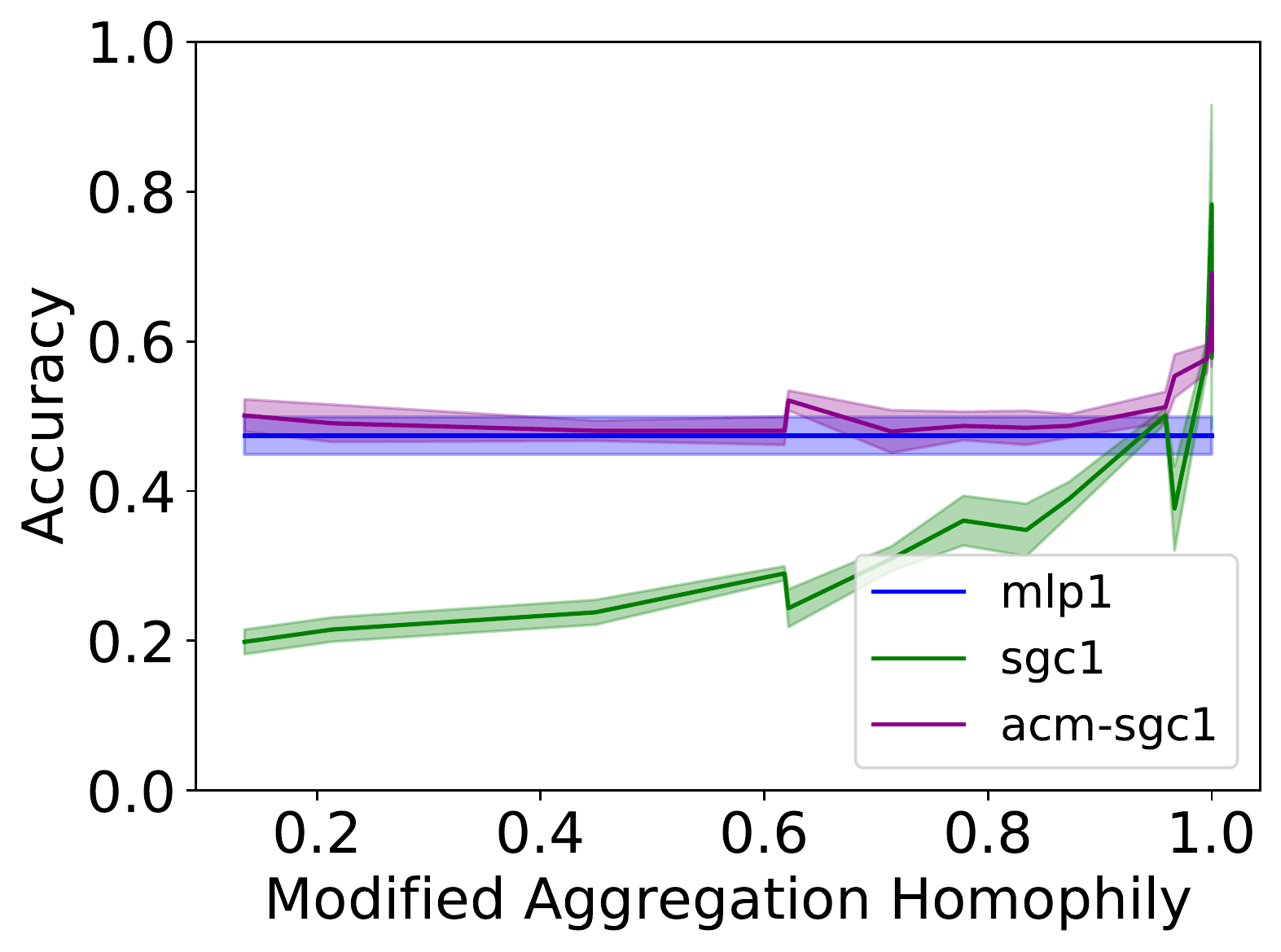}
     } \\
     \subfloat[ \texttt{syn-Chameleon}]{
     \captionsetup{justification = centering}
     \includegraphics[width=0.32\textwidth]{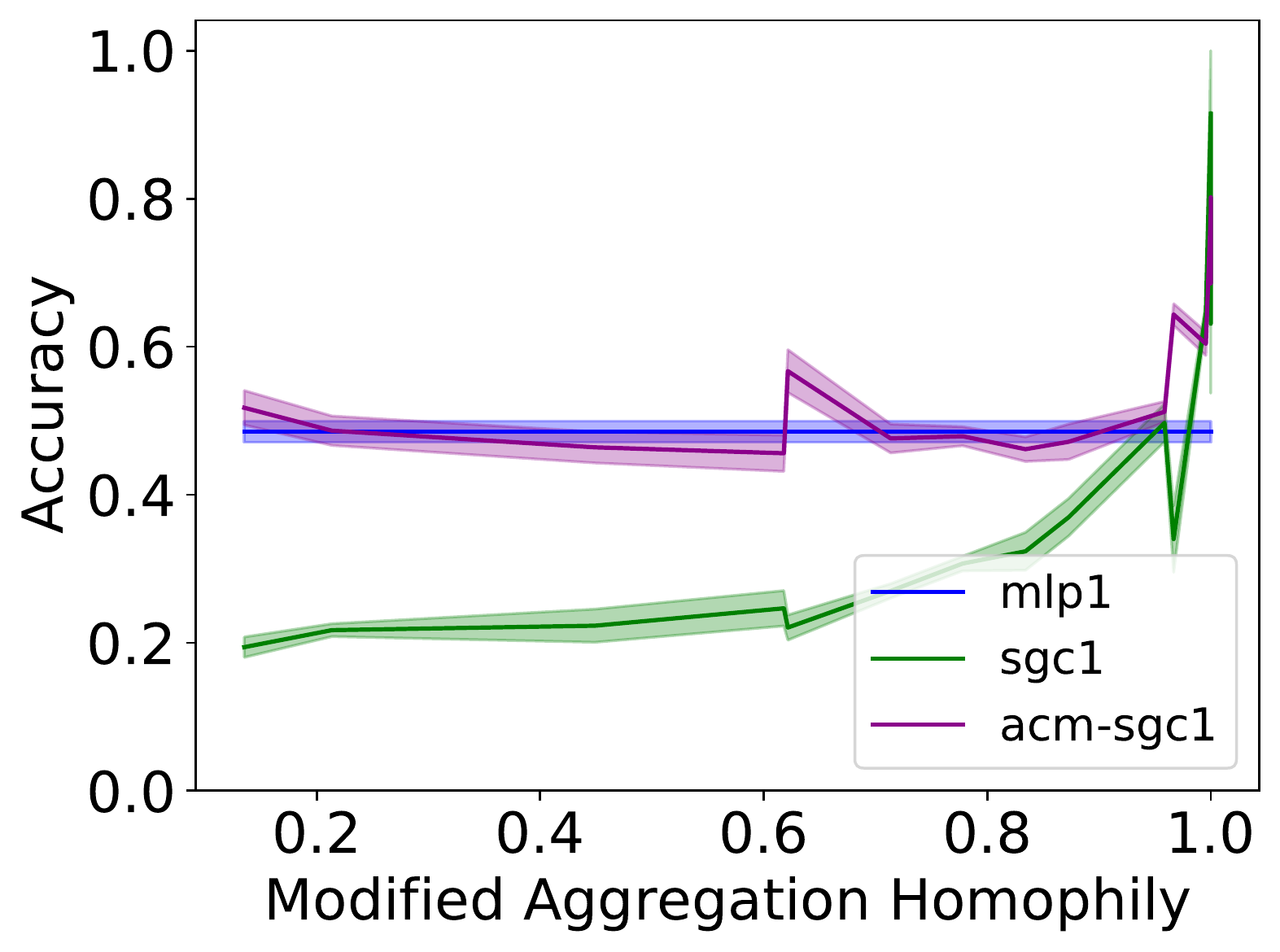}
     } 
     \subfloat[ \texttt{syn-Squirrel}]{
     \captionsetup{justification = centering}
     \includegraphics[width=0.32\textwidth]{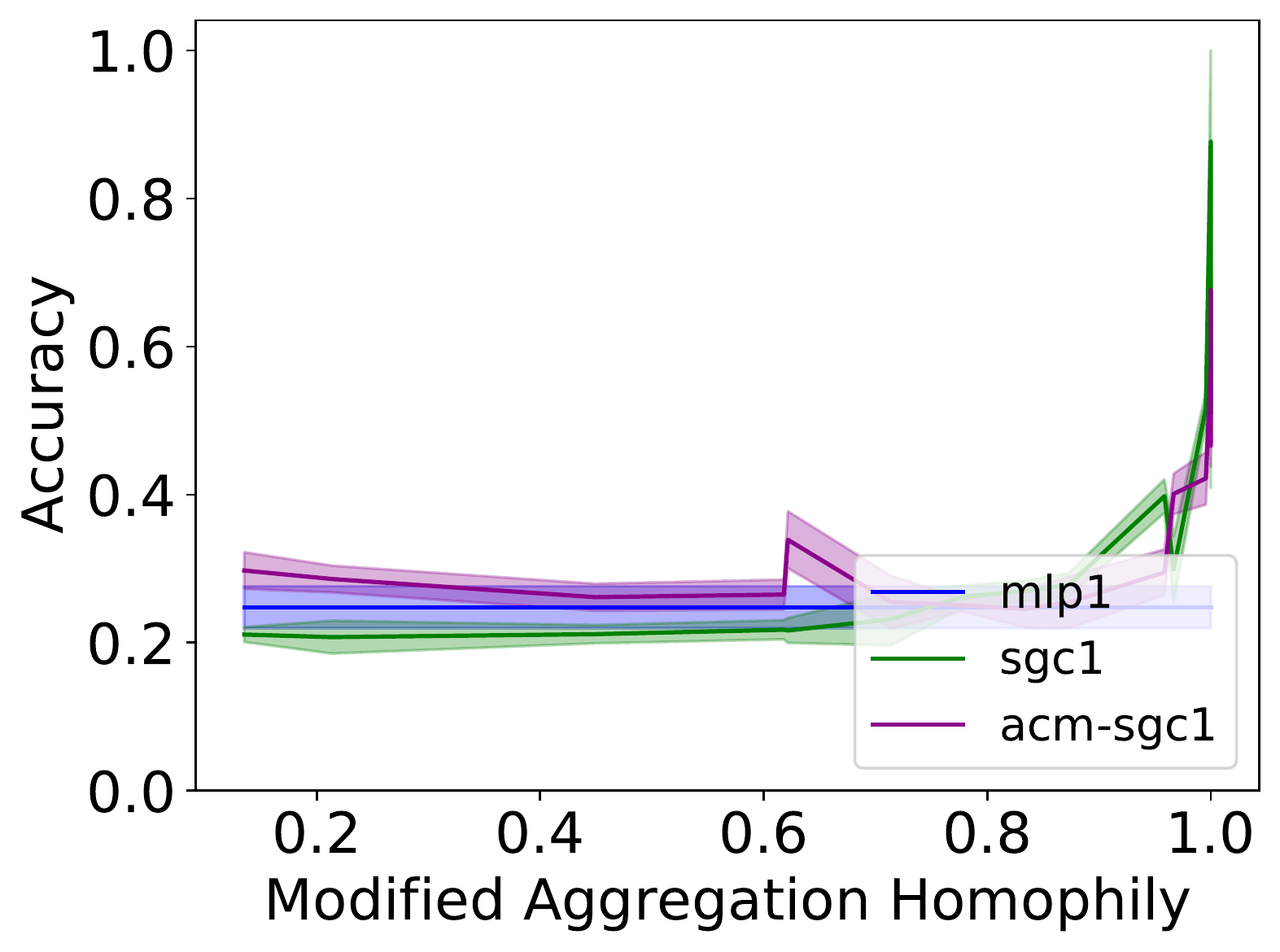}
     } 
     \subfloat[ \texttt{syn-Film}]{
     \captionsetup{justification = centering}
     \includegraphics[width=0.32\textwidth]{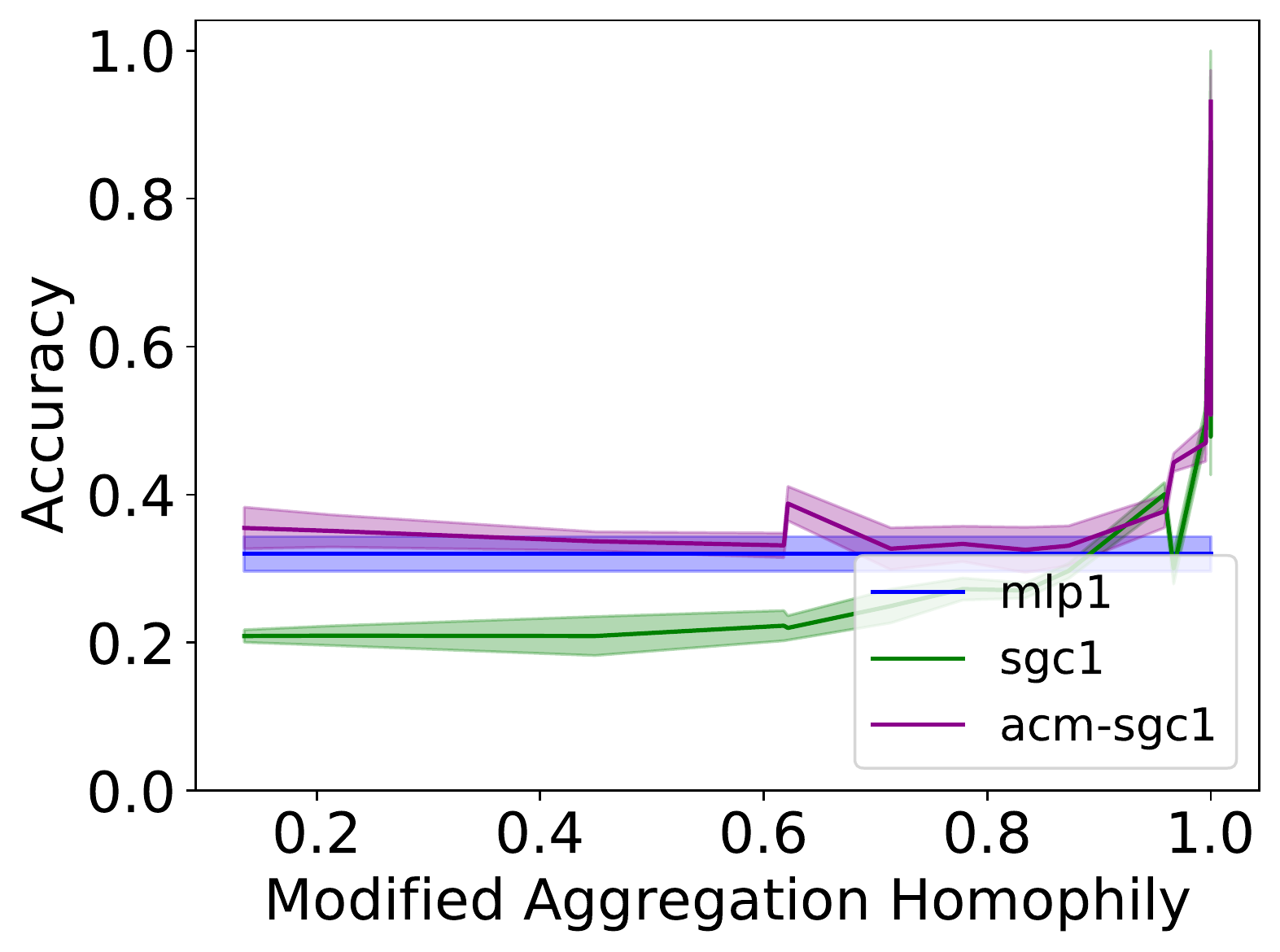}
     } 
     }
     \caption{Comparison of test accuracy (mean $\pm$ std) of MLP-1, SGC-1 and ACM-SGC-1 on synthetic datasets}
     \label{fig:sgc_acmsgc_synthetic_comparison}
\end{figure}

\begin{figure}[H]
    \centering
     { 
     \subfloat[\texttt{syn-Cora}]{
     \captionsetup{justification = centering}
     \includegraphics[width=0.32\textwidth]{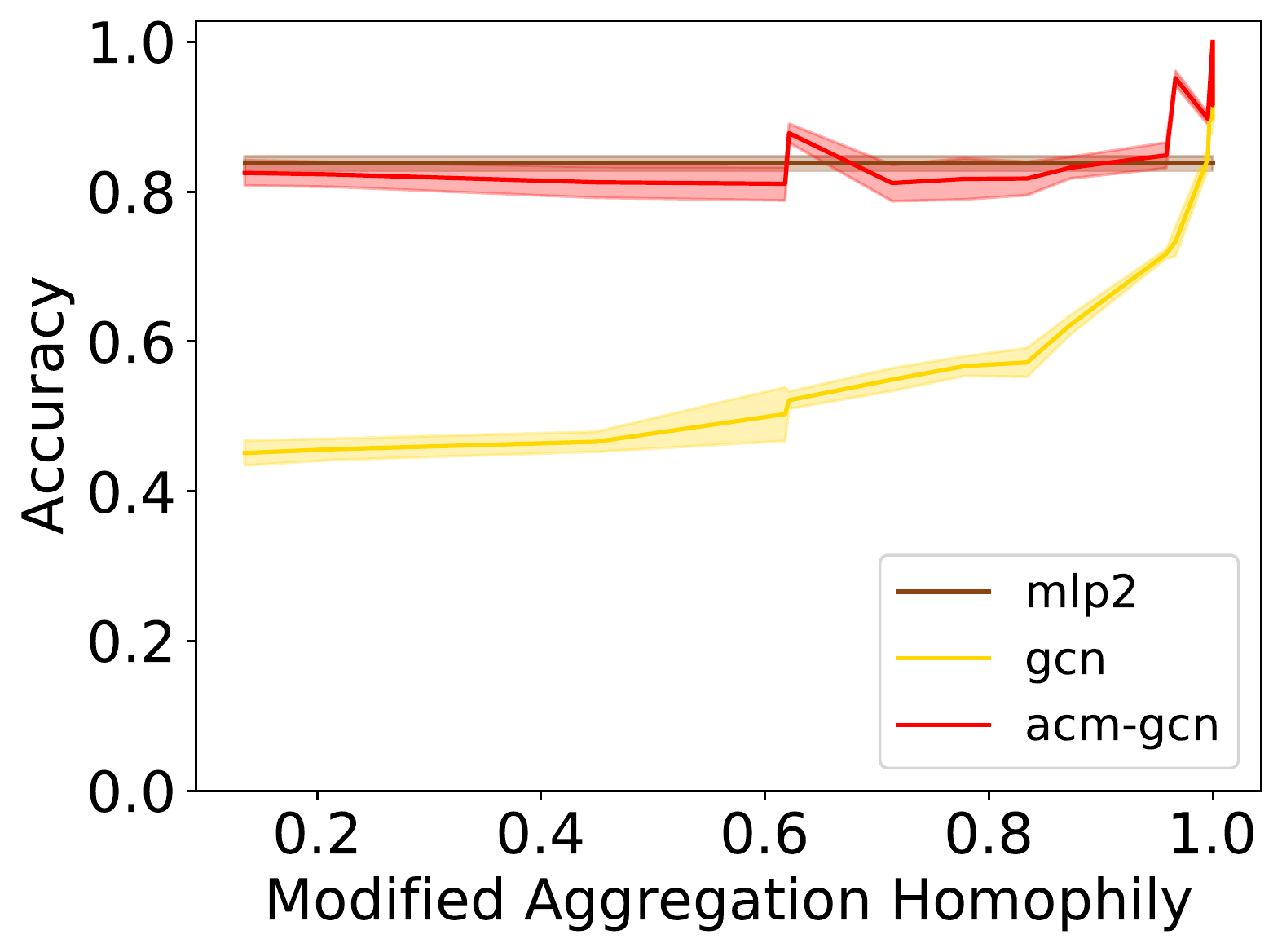}
     }
     \subfloat[\texttt{syn-CiteSeer}]{
     \captionsetup{justification = centering}
     \includegraphics[width=0.32\textwidth]{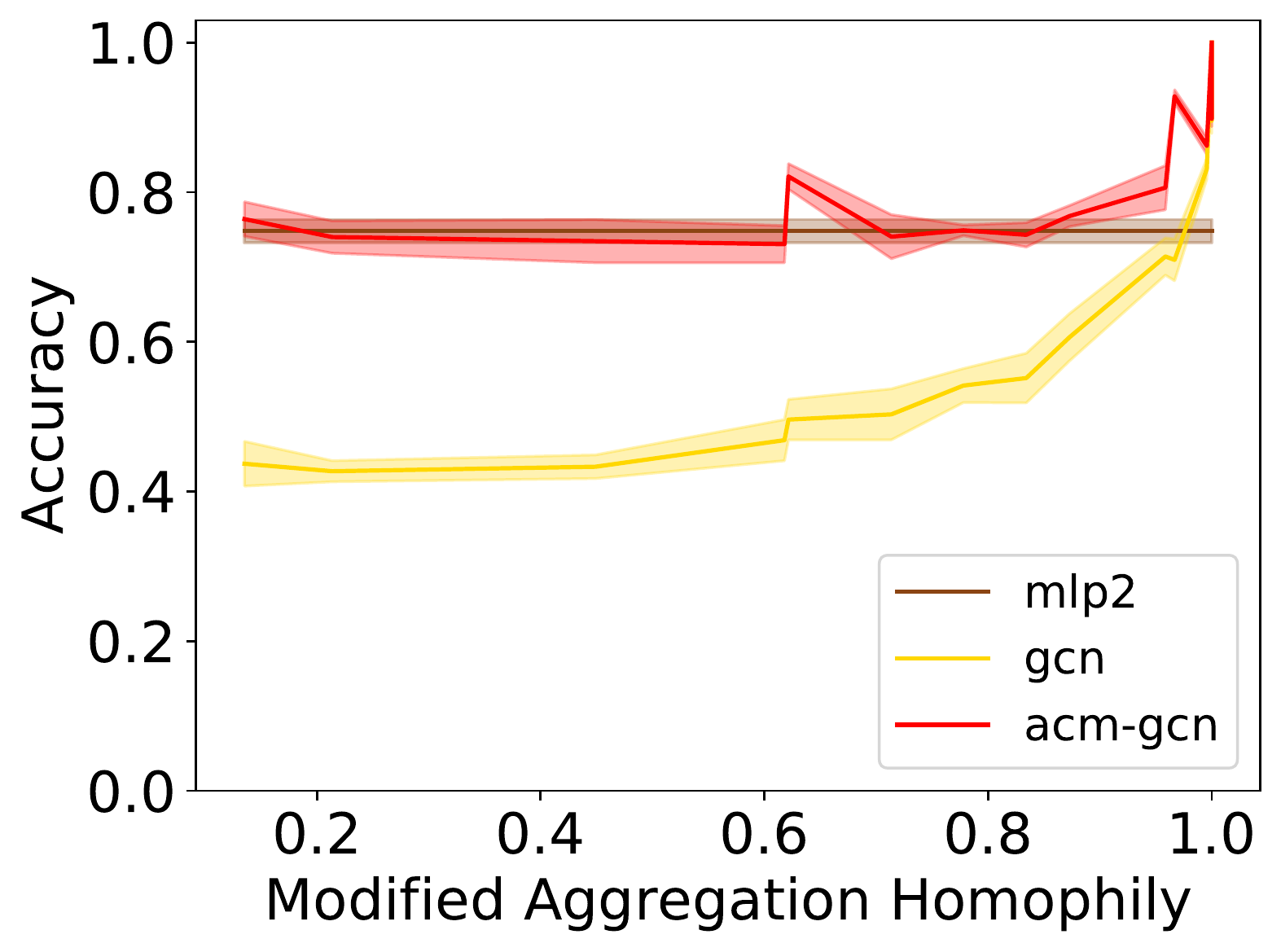}
     }
     \subfloat[\texttt{syn-PubMed}]{
     \captionsetup{justification = centering}
     \includegraphics[width=0.32\textwidth]{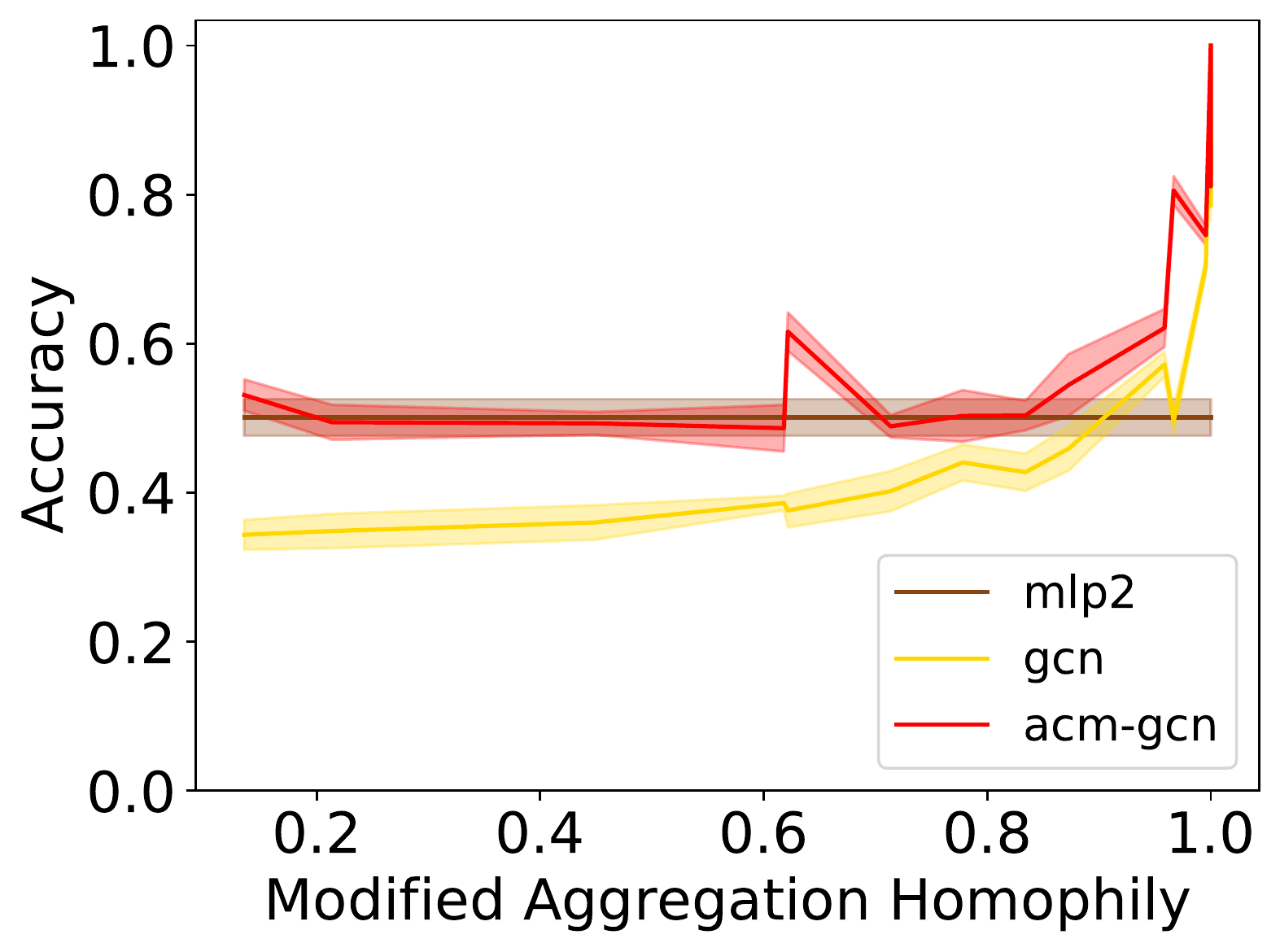}
     } \\
     \subfloat[\texttt{syn-Chameleon}]{
     \captionsetup{justification = centering}
     \includegraphics[width=0.32\textwidth]{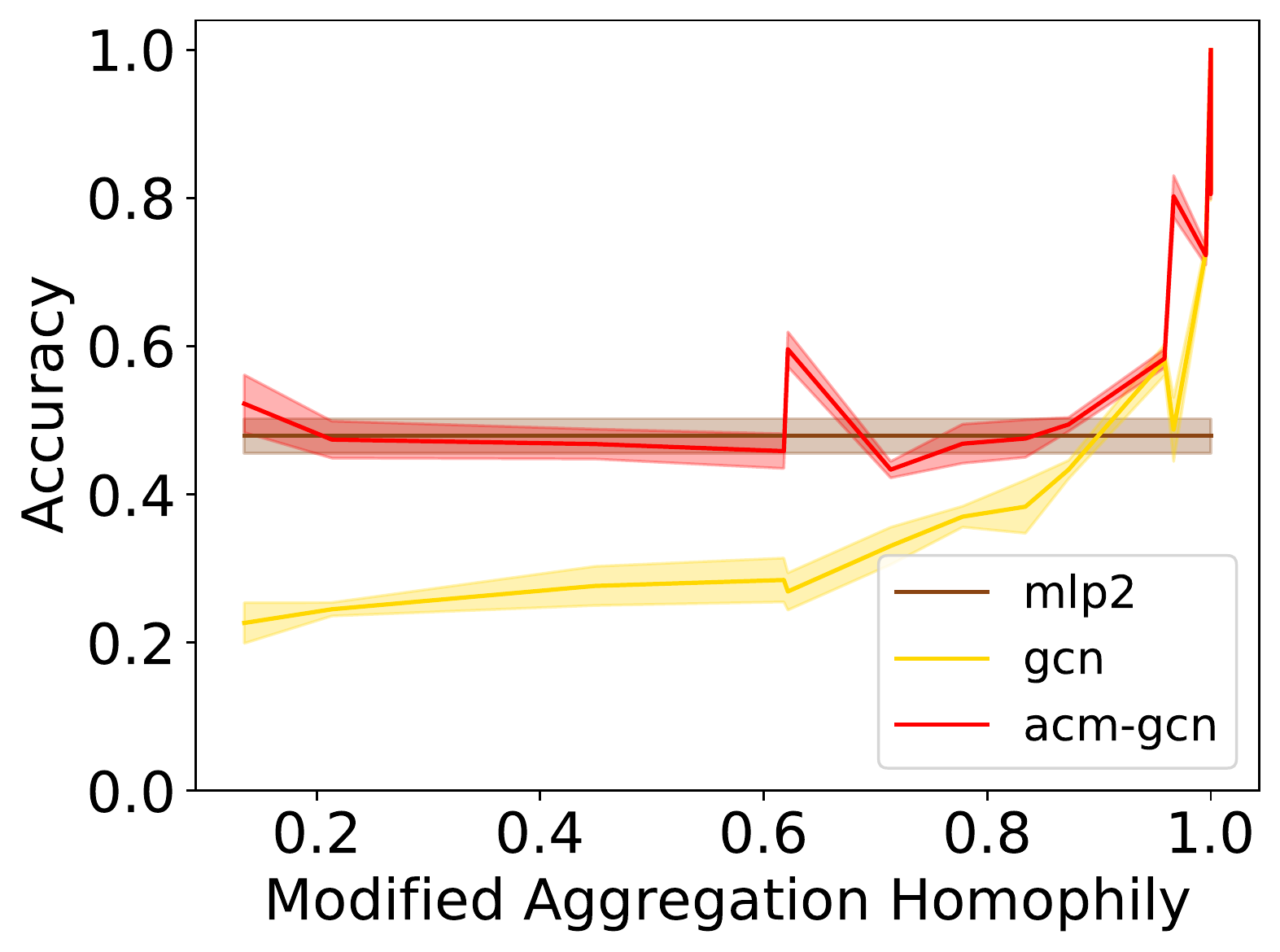}
     } 
     \subfloat[\texttt{syn-Squirrel}]{
     \captionsetup{justification = centering}
     \includegraphics[width=0.32\textwidth]{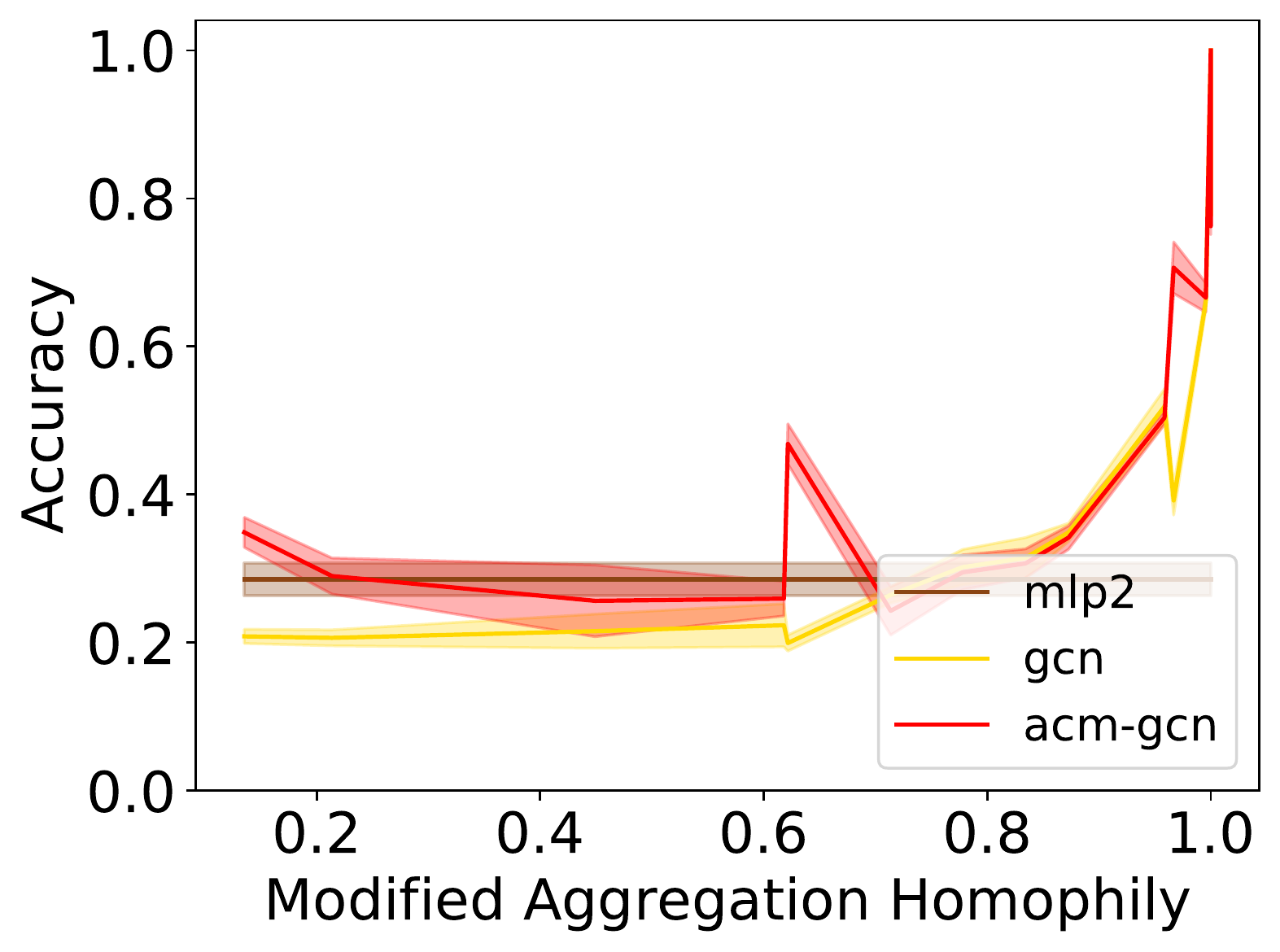}
     }
     \subfloat[\texttt{syn-Film}]{
     \captionsetup{justification = centering}
     \includegraphics[width=0.32\textwidth]{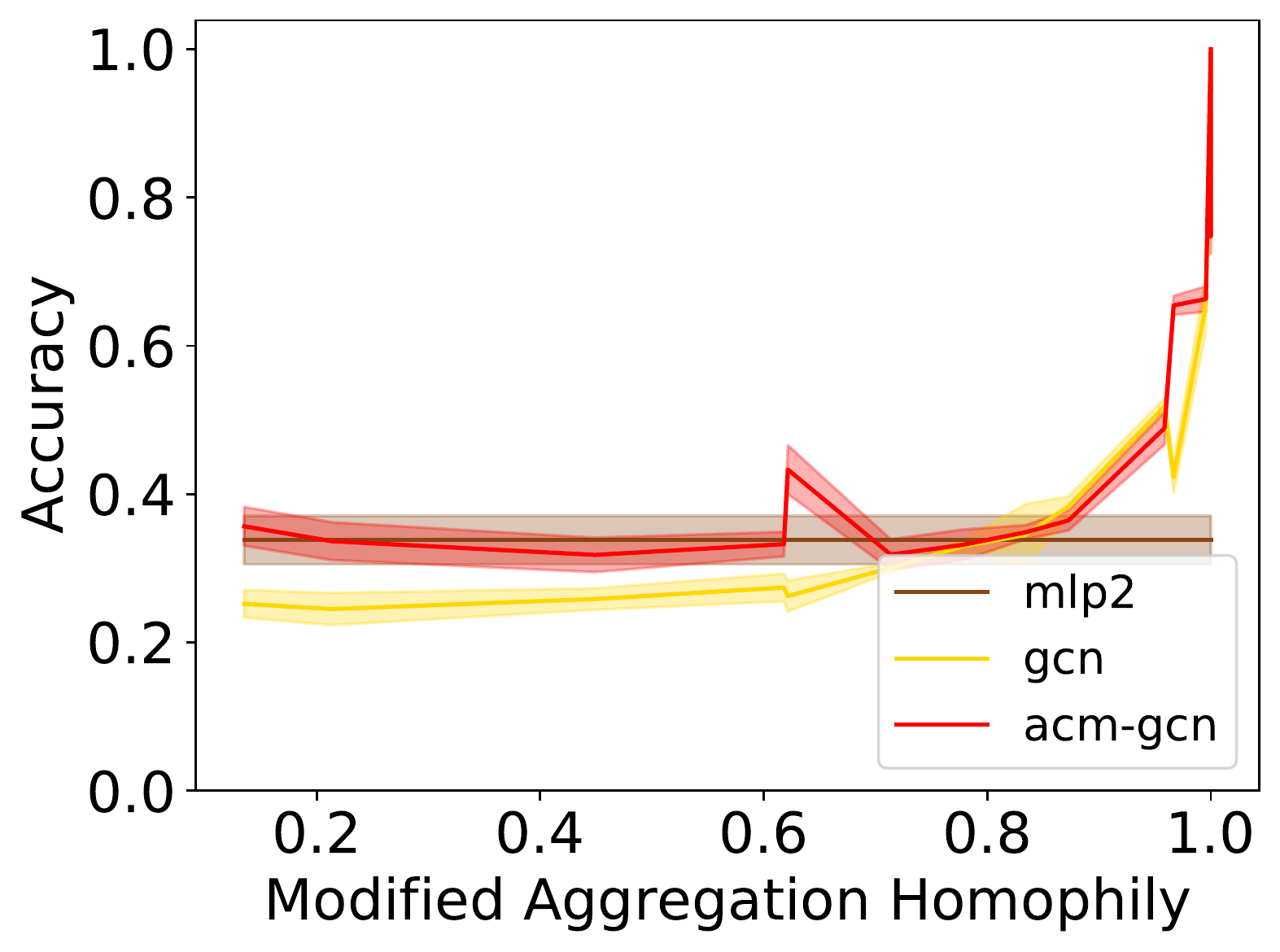}
     } 
     }
     \caption{Comparison of test accuracy (mean $\pm$ std) of MLP-2, GCN and ACM-GCN on synthetic datasets}
     \label{fig:gcn_acmgcn_synthetic_comparison}
\end{figure}

In order to separate the effects of nonlinearity and graph structure, we compare sgc with 1 hop (sgc-1) with MLP-1 (linear model). For GCN which includes nonlinearity, we use MLP-2 as the graph-agnostic baseline model. We train the above GNN models, graph-agnostic baseline models and ACM-GNN models on all synthetic datasets and plot the mean test accuracy with standard deviation on each dataset. From Figure \ref{fig:sgc_acmsgc_synthetic_comparison} and Figure \ref{fig:gcn_acmgcn_synthetic_comparison}, we can see that on each $H_\text{agg}^M(\mathcal{G})$ level, ACM-GNNs will not underperform GNNs and graph-agnostic models. But when $H_\text{agg}^M(\mathcal{G})$ is small, GNNs will be outperformed by graph-agnostic models by a large margin. This demonstrate the advantage of the ACM framework.

\section{Discussion of the Limitations of Diversification Operation}
\label{appendix:limitation_diversification}
\begin{figure}[htbp]
\centering
{
\captionsetup{justification = centering}
\includegraphics[width=1\textwidth]{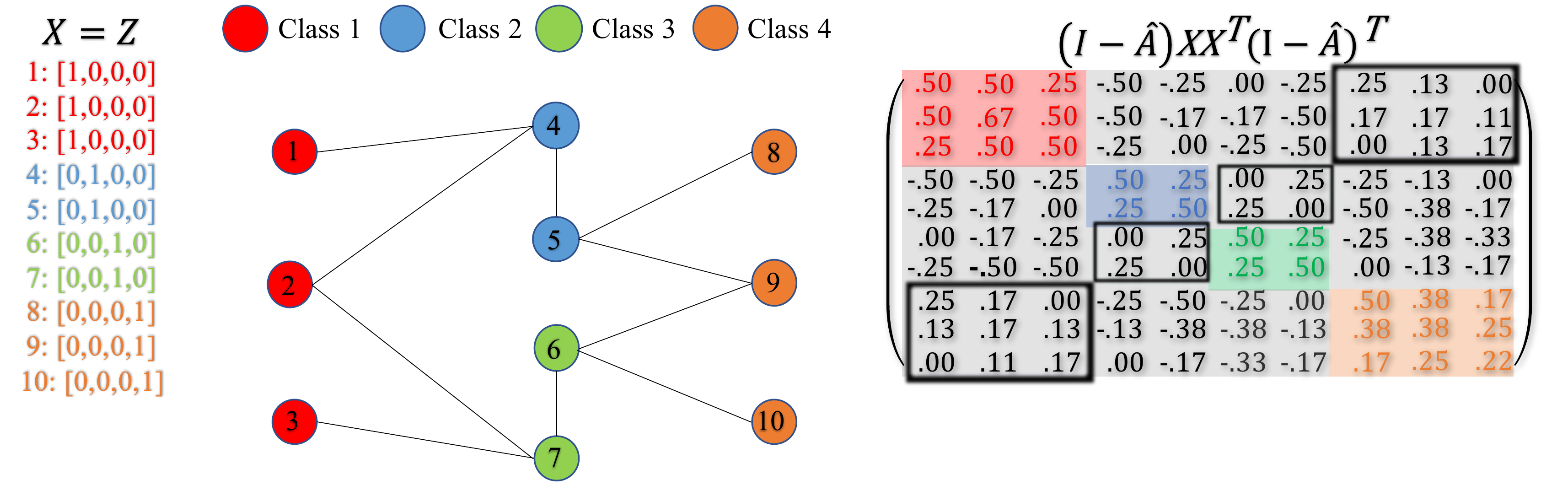}}
{%
  \caption{Example of the case (the area in black box) that HP filter does not work well for harmful heterophily}%
  \label{fig:unsuccessful_example_hp_filter}
}
\end{figure}
From the black box area of $S(I-\hat{A},X)$ in the example in Figure \ref{fig:unsuccessful_example_hp_filter} we can see that nodes in class 1 and 4 assign non-negative weights to each other; nodes in class 2 and 3 assign non-negative weights to each other as well. This is because the surrounding differences of class 1 are similar as class 4, so are class 2 and 3. In real-world applications, when nodes in several small clusters connect to a large cluster, the surrounding differences of the nodes in the small clusters will become similar. In such case, HP filter are not able to distinguish the nodes from different small clusters. 

\section{The Similarity, Homophily and $\mathrm{DD}_{\hat{A},X}(\mathcal{G})$ Metrics and Their Estimations}
\label{appendix:estimation_similarity_homophily_diversification_metrics}
\paragraph{Additional Metrics}
There are three key factors that influence the performance of GNNs in real-world tasks: labels, features and graph structure. The (modified) aggregation homophily tries to investigate how the graph structure will influence the performance with labels and features being fixed. And their correlation is verified through the synthetic experiments. 

Besides graph-label consistency, we need to consider feature-label consistency and aggregated-feature-label consistency as well. With aggregation similarity score of the features $S_\text{agg}\left(S(I,X)\right)$ and aggregated features $S_\text{agg}\left(S(\hat{A},X)\right)$ listed in appendix \ref{appendix:estimation_similarity_homophily_diversification_metrics}, our methods open up a new perspective on analyzing and comparing the performance of graph-agnostic models and graph-aware models in real-world tasks. Here are 2 examples.

Example 1: GCN (graph-aware model) underperforms MLP-2 (graph-agnostic model) on $\textit{Cornell, Wisconsin, Texas, Film}$. Based on the aggregation homophily, the graph structure is not the main cause of the performance degradation. And from Table 6 we can see that the $S_\text{agg}\left(S(\hat{A},X)\right)$ for the above 4 datasets are lower than their corresponding $S_\text{agg}\left(S(I,X)\right)$, which implies that it is the aggregated-feature-label inconsistency that causes the performance degradation.

Example 2: According to the widely used analysis based on node or edge homophily, the graph structure of $\textit{Chameleon}$, and $\textit{Squirrel}$ are heterophilous and bad for GNNs. But in practice, GCN outperforms MLP-2 on those 2 datasets which means the additional graph information is helpful for node classification instead of being harmful. Traditional homophily metrics fail to explain such phenomenon but our method can give an explanation from different angles: For Chameleon, its modified aggregation homophily is not low and its $S_\text{agg}\left(S(\hat{A},X)\right)$ is higher than its $S_\text{agg}\left(S(I,X)\right)$ which means its graph-label consistency and aggregated-feature-label consistency help the graph-aware model obtain the performance gain; for Squirrel, its modified aggregation homophily is low but its $S_\text{agg}\left(S(\hat{A},X)\right)$ is higher than its $S_\text{agg}\left(S(I,X)\right)$ which means although its graph-label consistency is bad but the aggregated-feature-label consistency is the key factor to help the graph-aware model perform better.

We also need to point out that (modified) aggregation similarity score, $S_\text{agg}\left(S(\hat{A},X)\right)$ and $S_\text{agg}\left(S(I,X)\right)$ are not deciding or threshold values because they do not consider the nonlinearity structure in the features. In practice, a low score does not tell us the GNN models will definitely perform bad.

\begin{table}[htbp]
  \centering
  \tiny
  \setlength{\tabcolsep}{2pt}
  \caption{Additional metrics and their estimations with only training labels (mean $\pm$ std)}
    \begin{tabular}{c|ccccccccc}
    \toprule
    \toprule
          & Cornell & Wisconsin & Texas & Film  & Chameleon & Squirrel & Cora  & CiteSeer & PubMed \\
    \midrule
    $H_\text{agg}(\mathcal{G})$ & 0.9016 & 0.8884 & 0.847 & 0.8411 & 0.805 & 0.6783 & 0.9952 & 0.9913 & 0.9716 \\
    $S_\text{agg}\left(S(\hat{A},X)\right)$ & 0.8251 & 0.7769 & 0.6557 & 0.5118 & 0.8292 & 0.7216 & 0.9439 & 0.9393 & 0.8623 \\
     $S_\text{agg}\left(S(I,X)\right)$ & 0.9672 & 0.8287 & 0.9672 & 0.5405 & 0.7931 & 0.701 & 0.9103 & 0.9315 & 0.8823 \\
    $DD_{\hat{A},X}(\mathcal{G})$ & 0.3497 & 0.6096 & 0.459 & 0.3279 & 0.3109 & 0.2711 & 0.2681 & 0.4124 & 0.1889 \\
    \midrule
    $\hat{H}_\text{agg}(\mathcal{G})$ & 0.9046 $\pm$ 0.0282 & 0.9147 $\pm$ 0.0260 & 0.8596 $\pm$ 0.0299 & 0.8451 $\pm$ 0.0041 & 0.8041 $\pm$ 0.0078 & 0.6788 $\pm$ 0.0077 & 0.9959 $\pm$ 0.0011 & 0.9907 $\pm$ 0.0015 & 0.9724 $\pm$ 0.0015 \\
    $\hat{S}_\text{agg}\left(S(\hat{A},X)\right)$ & 0.8266 $\pm$ 0.0526 & 0.8280 $\pm$ 0.0351 & 0.6835 $\pm$ 0.0498 & 0.5345 $\pm$ 0.0421 & 0.8433 $\pm$ 0.0070 & 0.7352 $\pm$ 0.0132 & 0.9487 $\pm$ 0.0023 & 0.9451 $\pm$ 0.0038 & 0.8626 $\pm$ 0.0021 \\
    $\hat{S}_\text{agg}\left(S(I,X)\right)$ & 0.9752 $\pm$ 0.0174 & 0.8680 $\pm$ 0.0270 & 0.9661 $\pm$ 0.0336 & 0.5438 $\pm$ 0.0184 & 0.8257 $\pm$ 0.0050 & 0.7472 $\pm$ 0.0089 & 0.9204 $\pm$ 0.0044 & 0.9441 $\pm$ 0.0036 & 0.8835 $\pm$ 0.0019 \\
    $\hat{DD}_{\hat{A},X}(\mathcal{G})$ & 0.3936 $\pm$ 0.0663 & 0.6073 $\pm$ 0.0436 & 0.4817 $\pm$ 0.0762 & 0.3300 $\pm$ 0.0136 & 0.3329 $\pm$ 0.0151 & 0.3021 $\pm$ 0.0101 & 0.3198 $\pm$ 0.0225 & 0.4424 $\pm$ 0.0136 & 0.1919 $\pm$ 0.0046 \\
    \bottomrule
    \bottomrule
    \end{tabular}%
  \label{tab:estimation_dataset_stats}%
\end{table}%

In most real-world applications, not all labels are available to calculate the dataset statistics. In this section, We randomly split the data into 60\%/20\%/20\% for training/validation/test, and only use the training labels for the estimation of the statistics. We repeat each estimation for 10 times and report the mean with standard deviation. The results are shown in table \ref{tab:estimation_dataset_stats}.

\paragraph{Estimation} The statistics we estimate are $H_\text{agg}(\mathcal{G})$, $S_\text{agg}\left(S(\hat{A},X)\right)$, $S_\text{agg}\left(S(I,X)\right)$ and $DD_{\hat{A},X}(\mathcal{G})$ and are denoted as $\hat{H}_\text{agg}(\mathcal{G})$, $\hat{S}_\text{agg}\left(S(\hat{A},X)\right)$, $\hat{S}_\text{agg}\left(S(I,X)\right)$ and $\hat{DD}_{\hat{A},X}(\mathcal{G})$. The two similarity scores $S_\text{agg}\left(S(\hat{A},X)\right)$ and $S_\text{agg}\left(S(I,X)\right)$ measures the proportion of nodes, according to aggregated features and nodes features respectively, that will put larger weights on nodes in the same class than in other classes. The higher values of $S_\text{agg}\left(S(\hat{A},X)\right)$ and $S_\text{agg}\left(S(I,X)\right)$ indicates the better quality of aggregated features and nodes features.
\paragraph{Analysis}
From the reported results we can see that the estimations are accurate and the errors are within the acceptable range, which means the proposed metrics and similarity scores can be accurately estimated with a subset of labels and this is important for real-world applications. Furthermore, we notice some interesting results, \eg{} the performance of GNNs and MLP are bad on \textit{Squirrel} and \textit{Film}, and according to the aggregation homophily values, the graph structure of \textit{Film} is not quite harmful compared to other datasets, but its features and aggregated features are much worse than others; the features and aggregated features of \textit{Squirrel} are not too bad, buts its graph topology is more harmful than others. Combining the metrics defined in this paper together can help us separate different factors in graph structure and features and identify what might cause the performance degradation of GNNs.

\section{Experiments on Fixed Splits Provided by \cite{pei2020geom}}
See table \ref{tab:performance_comparison_fixed_splits} for the results and table \ref{tab:optimal_hyperparameters_fixed_splits} the optimal searched hyperparameters.
\label{appendix:results_on_fixed_splits_as_geomgcn}
\begin{table}[htbp]
  \centering
  \caption{Experimental results on fixed splits provided by \cite{pei2020geom}: average test accuracy $\pm$ standard deviation on 9 real-world benchmark datasets. The best results are highlighted. Results of Geom-GCN, H$_2$GCN and GPRGNN are from \cite{pei2020geom,zhu2020beyond,lingam2021simple}; results on the rest models are run by ourselves and the hyperparameter searching range is the same as table \ref{tab:real_world_datasets_hyperparameter_searching_range}.}
  \setlength{\tabcolsep}{3pt}
   \scalebox{.65}{
    \begin{tabular}{c|ccccccccc|c}
    \toprule
    \toprule
          & Cornell & Wisconsin & Texas & Film  & Chameleon & Squirrel & Cora  & CiteSeer & PubMed & Rank \\
    \midrule
    GPRGNN & 78.11 $\pm$ 6.55 & 82.55 $\pm$ 6.23 & 81.35 $\pm$ 5.32 & 35.16 $\pm$ 0.9 & 62.59 $\pm$ 2.04 & 46.31 $\pm$ 2.46 & 87.95 $\pm$ 1.18 & 77.13 $\pm$ 1.67 & 87.54 $\pm$ 0.38 & 8.22 \\
    H2GCN & 82.70 $\pm$ 5.28 & 87.65 $\pm$ 4.98 & 84.86 $\pm$ 7.23 & 35.70 $\pm$ 1.00 & 60.11 $\pm$ 2.15 & 36.48 $\pm$ 1.86 & 87.87 $\pm$ 1.20 & 77.11 $\pm$ 1.57 & 89.49 $\pm$ 0.38 & 6.78 \\
    FAGCN & 76.76 $\pm$ 5.87 & 79.61 $\pm$ 1.58 & 76.49 $\pm$ 2.87 & 34.82 $\pm$ 1.35 & 46.07 $\pm$ 2.11 & 30.83 $\pm$ 0.69 & \cellcolor[rgb]{ .816,  .808,  .808}\textbf{88.05 $\pm$ 1.57} & 77.07 $\pm$ 2.05 & 88.09 $\pm$ 1.38 & 9.56 \\
    Geom-GCN* & 60.54 $\pm$ 3.67 & 64.51 $\pm$ 3.66 & 66.76 $\pm$ 2.72 & 31.59 $\pm$ 1.15 & 60.00 $\pm$ 2.81 & 38.15 $\pm$ 0.92 & 85.35 $\pm$ 1.57 & \cellcolor[rgb]{ .816,  .808,  .808}\textbf{78.02 $\pm$ 1.15} & 89.95 $\pm$ 0.47 & 9.22 \\
    \midrule
    ACM-SGC-1 & 82.43 $\pm$ 5.44 & 86.47 $\pm$ 3.77 & 81.89 $\pm$ 4.53 & 35.49 $\pm$ 1.06 & 63.99 $\pm$ 1.66 & 45.00 $\pm$ 1.4 & 86.9 $\pm$ 1.38 & 76.73 $\pm$ 1.59 & 88.49 $\pm$ 0.51 & 8.44 \\
    ACM-SGC-2 & 82.43 $\pm$ 5.44 & 86.47 $\pm$ 3.77 & 81.89 $\pm$ 4.53 & 36.04 $\pm$ 0.83 & 59.21 $\pm$ 2.22 & 40.02 $\pm$ 0.96 & 87.69 $\pm$ 1.07 & 76.59 $\pm$ 1.69 & 89.01 $\pm$ 0.6 & 8.22 \\
    ACM-GCN & 85.14 $\pm$ 6.07 & \cellcolor[rgb]{ .816,  .808,  .808}\textbf{88.43 $\pm$ 3.22} & \cellcolor[rgb]{ .816,  .808,  .808}\textbf{87.84 $\pm$ 4.4} & 36.28 $\pm$ 1.09 & 66.93 $\pm$ 1.85 & \cellcolor[rgb]{ .816,  .808,  .808}\textbf{54.4 $\pm$ 1.88} & 87.91 $\pm$ 0.95 & 77.32 $\pm$ 1.7 & \cellcolor[rgb]{ .816,  .808,  .808}\textbf{90.00 $\pm$ 0.52} & \cellcolor[rgb]{ .816,  .808,  .808}\textbf{2.33} \\
    ACM-Snowball-2 & 85.41 $\pm$ 5.43 & 87.06 $\pm$ 2 & 87.57 $\pm$ 4.86 & \cellcolor[rgb]{ .816,  .808,  .808}\textbf{36.89 $\pm$ 1.18} & \cellcolor[rgb]{ .816,  .808,  .808}\textbf{67.08 $\pm$ 2.04} & 52.5 $\pm$ 1.49 & 87.42 $\pm$ 1.09 & 76.41 $\pm$ 1.38 & 89.89 $\pm$ 0.57 & 4.11 \\
    ACM-Snowball-3 & 83.24 $\pm$ 5.38 & 86.67 $\pm$ 4.37 & \cellcolor[rgb]{ .816,  .808,  .808}\textbf{87.84 $\pm$ 3.87} & 36.82 $\pm$ 0.94 & 66.91 $\pm$ 1.73 & 53.31 $\pm$ 1.88 & 87.1 $\pm$ 0.93 & 75.91 $\pm$ 1.57 & 89.81 $\pm$ 0.43 & 5.22 \\
    ACMII-GCN & \cellcolor[rgb]{ .816,  .808,  .808}\textbf{85.95 $\pm$ 5.64} & 87.45 $\pm$ 3.74 & 86.76 $\pm$ 4.75 & 36.16 $\pm$ 1.11 & 66.91 $\pm$ 2.55 & 51.85 $\pm$ 1.38 & 88.01 $\pm$ 1.08 & 77.15 $\pm$ 1.45 & 89.89 $\pm$ 0.43 & 3.22 \\
    ACMII-Snowball-2 & 85.68 $\pm$ 5.93 & 87.45 $\pm$ 2.8 & 86.76 $\pm$ 4.43 & 36.55 $\pm$ 1.24 & 66.49 $\pm$ 1.75 & 50.15 $\pm$ 1.4 & 87.57 $\pm$ 0.86 & 76.92 $\pm$ 1.45 & 89.84 $\pm$ 0.48 & 4.67 \\
    ACMII-Snowball-3 & 82.7 $\pm$ 4.86 & 85.29 $\pm$ 4.23 & 85.41 $\pm$ 6.42 & 36.49 $\pm$ 1.41 & 66.86 $\pm$ 1.74 & 48.87 $\pm$ 1.23 & 87.16 $\pm$ 1.01 & 76.18 $\pm$ 1.55 & 89.73 $\pm$ 0.52 & 7.00 \\
    \bottomrule
    \bottomrule
    \end{tabular}%
    }
  \label{tab:performance_comparison_fixed_splits}%
\end{table}%

\begin{table}[htbp]
  \centering
  \caption{Hyperparameters for FAGCN and ACM-GNNs on fixed splits}
   \scalebox{.675}{
    \begin{tabular}{c|c|ccccccc}
    \toprule
    Datasets & Models\textbackslash{}Hyperparameters & lr    & weight\_decay & dropout & hidden & results & std   & average epoch time/average total time \\
     \midrule
    \multirow{9}[0]{*}{\textbf{Cornell}} & ACM-SGC-1 & 0.01  & 5.00E-06 & 0     & 64    & 82.43 & 5.44  & 5.37ms/23.05s \\
          & ACM-SGC-2 & 0.01  & 5.00E-06 & 0     & 64    & 82.43 & 5.44  & 5.93ms/25.66s \\
          & ACM-GCN & 0.05  & 5.00E-04 & 0.5   & 64    & 85.14 & 6.07  & 8.04ms/1.67s \\
          & ACMII-GCN & 0.1   & 1.00E-04 & 0     & 64    & 85.95 & 5.64  & 7.83ms/2.66s \\
          & FAGCN & 0.01  & 1.00E-04 & 0.6   & 64    & 76.76 & 5.87  & 8.80ms/7.67s \\
          & ACM-Snowball-2 & 0.05  & 5.00E-03 & 0.3   & 64    & 85.41 & 5.43  & 11.50ms/2.35s \\
          & ACM-Snowball-3 & 0.05  & 5.00E-03 & 0.2   & 64    & 83.24 & 5.38  & 15.06ms/3.12s \\
          & ACMII-Snowball-2 & 0.1   & 5.00E-03 & 0.2   & 64    & 85.68 & 5.93  & 12.63ms/2.58s \\
          & ACMII-Snowball-3 & 0.05  & 5.00E-03 & 0.2   & 64    & 82.7  & 4.86  & 14.59ms/3.06s \\
          \midrule
    \multirow{9}[0]{*}{\textbf{Wisconsin}} & ACM-SGC-1 & 0.1   & 5.00E-06 & 0     & 64    & 86.47 & 3.77  & 5.07ms/14.07s \\
          & ACM-SGC-2 & 0.1   & 5.00E-06 & 0     & 64    & 86.47 & 3.77  & 5.30ms/16.05s \\
          & ACM-GCN & 0.05  & 1.00E-03 & 0.4   & 64    & 88.43 & 3.22  & {8.04ms/1.66s} \\
          & ACMII-GCN & 0.01  & 5.00E-05 & 0.1   & 64    & 87.45 & 3.74  & 8.40ms/2.19s \\
          & FAGCN & 0.01  & 5.00E-05 & 0.5   & 64    & 79.61 & 1.59  & 8.61ms/5.84s \\
          & ACM-Snowball-2 & 0.01  & 1.00E-03 & 0.4   & 64    & 87.06 & 2     & 12.51ms/2.60s \\
          & ACM-Snowball-3 & 0.01  & 1.00E-02 & 0.1   & 64    & 86.67 & 4.37  & 14.92ms/3.15s \\
          & ACMII-Snowball-2 & 0.01  & 5.00E-04 & 0.1   & 64    & 87.45 & 2.8   & 11.96ms/2.63s \\
          & ACMII-Snowball-3 & 0.01  & 5.00E-03 & 0.5   & 64    & 85.29 & 4.23  & 14.87ms/3.10s \\
           \midrule
    \multirow{9}[0]{*}{\textbf{Texas}} & ACM-SGC-1 & 0.01  & 1.00E-05 & 0     & 64    & 81.89 & 4.53  & 5.34ms/19.00s \\
          & ACM-SGC-2 & 0.05  & 1.00E-05 & 0     & 64    & 81.89 & 4.53  & 5.50ms/9.26s \\
          & ACM-GCN & 0.05  & 5.00E-04 & 0.5   & 64    & 87.84 & 4.4   & 9.62ms/1.99s \\
          & ACMII-GCN & 0.01  & 1.00E-03 & 0.1   & 64    & 86.76 & 4.75  & 9.98ms/2.22s \\
          & FAGCN & 0.01  & 1.00E-05 & 0     & 64    & 76.49 & \textcolor[rgb]{ .267,  .267,  .267}{2.87} & 10.45ms/5.70s \\
          & ACM-Snowball-2 & 0.01  & 5.00E-03 & 0.2   & 64    & 87.57 & 4.86  & 11.56ms/2.45s \\
          & ACM-Snowball-3 & 0.01  & 5.00E-03 & 0.2   & 64    & 87.84 & 3.87  & 15.17ms/3.15s \\
          & ACMII-Snowball-2 & 0.01  & 1.00E-03 & 0.2   & 64    & 86.76 & 4.43  & 11.36ms/2.30 \\
          & ACMII-Snowball-3 & 0.01  & 5.00E-03 & 0.6   & 64    & 85.41 & 6.42  & 15.84ms/3.48s \\
           \midrule
    \multirow{9}[0]{*}{\textbf{Film}} & ACM-SGC-1 & 0.05  & 5.00E-04 & 0     & 64    & 35.49 & 1.06  & 5.39ms/1.17s \\
          & ACM-SGC-2 & 0.05  & 5.00E-04 & 0.1   & 64    & 36.04 & 0.83  & 13.22ms/3.31s \\
          & ACM-GCN & 0.01  & 5.00E-03 & 0     & 64    & 36.28 & 1.09  & 8.96ms/1.82s \\
          & ACMII-GCN & 0.01  & 5.00E-03 & 0     & 64    & 36.16 & 1.11  & 9.06ms/1.83s \\
          & FAGCN & 0.01  & 5.00E-05 & 0.4   & 64    & 34.82 & 1.35  & 15.60ms/2.51s \\
          & ACM-Snowball-2 & 0.01  & 1.00E-02 & 0     & 64    & 36.89 & 1.18  & 14.77ms/3.01s \\
          & ACM-Snowball-3 & 0.01  & 1.00E-02 & 0.2   & 64    & 36.82 & 0.94  & 16.57ms/3.36s \\
          & ACMII-Snowball-2 & 0.01  & 5.00E-03 & 0.1   & 64    & 36.55 & 1.24  & 12.76ms/2.57s \\
          & ACMII-Snowball-3 & 0.05  & 5.00E-03 & 0.3   & 64    & 36.49 & 1.41  & 16.51ms/3.49s \\
           \midrule
    \multirow{9}[0]{*}{\textbf{Chameleon}} & ACM-SGC-1 & 0.1   & 5.00E-06 & 0.9   & 64    & 63.99 & 1.66  & 5.92ms/1.74s \\
          & ACM-SGC-2 & 0.1   & 0.00E+00 & 0.9   & 64    & 59.21 & 2.22  & 8.84ms/1.78s \\
          & ACM-GCN & 0.05  & 5.00E-05 & 0.7   & 64    & 66.93 & 1.85  & 8.40ms/1.71s \\
          & ACMII-GCN & 0.05  & 5.00E-06 & 0.8   & 64    & 66.91 & 2.55  & 8.90ms/2.10s \\
          & FAGCN & 0.01  & 5.00E-05 & 0     & 64    & 46.07 & 2.11  & 16.90ms/7.94s \\
          & ACM-Snowball-2 & 0.01  & 1.00E-04 & 0.7   & 64    & 67.08 & 2.04  & 12.50ms/2.69s \\
          & ACM-Snowball-3 & 0.01  & 1.00E-05 & 0.8   & 64    & 66.91 & 1.73  & 16.12ms/4.91s \\
          & ACMII-Snowball-2 & 0.01  & 5.00E-05 & 0.8   & 64    & 66.49 & 1.75  & 12.65ms/3.42s \\
          & ACMII-Snowball-3 & 0.05  & 5.00E-05 & 0.7   & 64    & 66.86 & 1.74  & 17.60ms/4.06s \\
           \midrule
    \multirow{9}[0]{*}{\textbf{Squirrel}} & ACM-SGC-1 & 0.05  & 5.00E-06 & 0.9   & 64    & 45    & 1.4   & 6.10ms/2.18s \\
          & ACM-SGC-2 & 0.05  & 0.00E+00 & 0.9   & 64    & 40.02 & 0.96  & 35.75ms/9.62s \\
          & ACM-GCN & 0.05  & 5.00E-06 & 0.7   & 64    & 54.4  & 1.88  & 10.48ms/2.68s \\
          & ACMII-GCN & 0.05  & 5.00E-06 & 0.7   & 64    & 51.85 & 1.38  & 11.69ms/2.91s \\
          & FAGCN & 0     & 5.00E-03 & 0     & 64    & 30.86 & 0.69  & 10.90ms/13.91s \\
          & ACM-Snowball-2 & 0.01  & 1.00E-04 & 0.7   & 64    & 52.5  & 1.49  & 17.89ms/5.78s \\
          & ACM-Snowball-3 & 0.01  & 5.00E-05 & 0.7   & 64    & 53.31 & 1.88  & 22.60ms/7.53s \\
          & ACMII-Snowball-2 & 0.05  & 5.00E-05 & 0.6   & 64    & 50.15 & 1.4   & 16.95ms/3.45s \\
          & ACMII-Snowball-3 & 0.01  & 5.00E-04 & 0.6   & 64    & 48.87 & 1.23  & 23.52ms/4.94s \\
           \midrule
    \multirow{9}[0]{*}{\textbf{Cora}} & ACM-SGC-1 & 0.05  & 5.00E-05 & 0.7   & 64    & 86.9  & 1.38  & 4.99ms/2.40s \\
          & ACM-SGC-2 & 0.1   & 0     & 0.8   & 64    & 87.69 & 1.07  & 5.16ms/1.16s \\
          & ACM-GCN & 0.01  & 5.00E-05 & 0.6   & 64    & 87.91 & 0.95  & 8.41ms/1.84s \\
          & ACMII-GCN & 0.01  & 1.00E-04 & 0.6   & 64    & 88.01 & 1.08  & 8.59ms/1.96s \\
          & FAGCN & 0.02  & 1.00E-04 & 0.5   & 64    & 88.05 & 1.57  & 9.30ms/10.64s \\
          & ACM-Snowball-2 & 0.01  & 1.00E-03 & 0.5   & 64    & 87.42 & 1.09  & 12.54ms/2.72s \\
          & ACM-Snowball-3 & 0.01  & 5.00E-06 & 0.9   & 64    & 87.1  & 0.93  & 15.83ms/11.33s \\
          & ACMII-Snowball-2 & 0.01  & 1.00E-03 & 0.6   & 64    & 87.57 & 0.86  & 12.06ms/2.64s \\
          & ACMII-Snowball-3 & 0.01  & 5.00E-03 & 0.5   & 64    & 87.16 & 1.01  & 16.29ms/3.62s \\
           \midrule
    \multirow{9}[0]{*}{\textbf{CiteSeer}} & ACM-SGC-1 & 0.05  & 0.00E+00 & 0.7   & 64    & 76.73 & 1.59  & 5.24ms/1.14s \\
          & ACM-SGC-2 & 0.1   & 0.00E+00 & 0.8   & 64    & 76.59 & 1.69  & 5.14ms/1.03s \\
          & ACM-GCN & 0.01  & 5.00E-06 & 0.3   & 64    & 77.32 & 1.7   & 8.89ms/1.79s \\
          & ACMII-GCN & 0.01  & 5.00E-05 & 0.5   & 64    & 77.15 & 1.45  & 8.95ms/1.80s \\
          & FAGCN & 0.02  & 5.00E-05 & 0.4   & 64    & 77.07 & 2.05  & 10.05ms/5.69s \\
          & ACM-Snowball-2 & 0.01  & 5.00E-05 & 0     & 64    & 76.41 & 1.38  & 12.87ms/2.59s \\
          & ACM-Snowball-3 & 0.01  & 5.00E-06 & 0.9   & 64    & 75.91 & 1.57  & 17.40ms/11.92s \\
          & ACMII-Snowball-2 & 0.01  & 5.00E-03 & 0.5   & 64    & 76.92 & 1.45  & 13.10ms/2.94s \\
          & ACMII-Snowball-3 & 0.1   & 5.00E-05 & 0.9   & 64    & 76.18 & 1.55  & 17.47ms/5.88s \\
           \midrule
    \multirow{9}[1]{*}{\textbf{PubMed}} & ACM-SGC-1 & 0.05  & 5.00E-06 & 0.4   & 64    & 88.49 & 0.51  & 5.77ms/3.65s \\
          & ACM-SGC-2 & 0.05  & 5.00E-06 & 0.3   & 64    & 89.01 & 0.6   & 8.50ms/5.18s \\
          & ACM-GCN & 0.01  & 5.00E-05 & 0.4   & 64    & 90    & 0.52  & 8.99ms/2.51s \\
          & ACMII-GCN & 0.01  & 1.00E-04 & 0.3   & 64    & 89.89 & 0.43  & 9.70ms/2.57s \\
          & FAGCN & 0.01  & 1.00E-04 & 0     & 64    & 88.09 & 1.38  & 10.30ms/8.75s \\
          & ACM-Snowball-2 & 0.01  & 1.00E-03 & 0.3   & 64    & 89.89 & 0.57  & 15.05ms/3.11s \\
          & ACM-Snowball-3 & 0.01  & 5.00E-03 & 0.1   & 64    & 89.81 & 0.43  & 20.51ms/4.63s \\
          & ACMII-Snowball-2 & 0.01  & 5.00E-04 & 0.4   & 64    & 89.84 & 0.48  & 15.10ms/3.2s \\
          & ACMII-Snowball-3 & 0.01  & 1.00E-03 & 0.4   & 64    & 89.73 & 0.52  & 20.46ms/4.32s \\
    \bottomrule
    \bottomrule
    \end{tabular}%
    }
  \label{tab:optimal_hyperparameters_fixed_splits}%
\end{table}%

\section{A Detailed Explanation of the Differences Between ACM(II)-GNNs and Several SOTA Models}

\begin{itemize}
    \item Difference with GPRGNN \cite{chien2021adaptive}: To explain the difference between channel mixing mechanism and the learning mechanism in GPRGNN, we first rewrite GPRGNN as $\mathbf{Z} = \sum\limits_{k=0}^{K} \gamma_{k} \mathbf{H}^{(k)} = \sum\limits_{k=0}^{K} \gamma_{k} I \mathbf{H}^{(k)} = \sum\limits_{k=0}^{K} diag(\gamma_{k}, \gamma_{k},\dots,\gamma_{k}) \mathbf{H}^{(k)}$. The node-wise channel mixing mechanism in GPRGNN form is $\mathbf{Z} = \sum\limits_{k=0}^{K} diag(\gamma_{k}^1,\gamma_{k}^2,\dots,\gamma_{k}^N) \mathbf{H}^{(k)}$, where $N$ is the number of nodes and $\gamma_{k}^i, i=1,\dots,N$ are learnable parametric mixing weights.

    \item Difference with FAGCN \cite{bo2021beyond}: instead of using a fixed $\hat{A}$, FAGCN learns a new filter $\hat{A}'$ based on $\hat{A}$. And $\hat{A}'$ can be decomposed as $\hat{A}'=\hat{A}_1'-\hat{A}_2'$, where $\hat{A}_1'$ and $-\hat{A}_2'$ represents positive and negative edge (propagation) information respectively. In our paper, we are not discussing the advantages of using the learned filter $\hat{A}'$ over the fixed filter $\hat{A}$, we are comparing the models with and without channel mixing mechanism. We believe the empirical results on real-world tasks in table \ref{tab:sota} and table \ref{tab:performance_comparison_fixed_splits} is the best way to compare the models with fixed filter and node-wise channel mixing and the models with learned filter but without channel mixing
\end{itemize}

\end{document}